\documentclass[
reprint,
superscriptaddress,
nofootinbib,
amsmath,amssymb,
aps,
prx,
showkeys
]{revtex4-2}

\usepackage{graphicx}
\usepackage{bm}
\usepackage{xspace}
\usepackage[table]{xcolor}
\usepackage[
  colorlinks=true,
  linkcolor=blue,
  citecolor=blue,
  urlcolor=blue,
  breaklinks=true
]{hyperref}  
\newcommand{\figref}[2][]{Fig.\ \ref{#2}#1\xspace}  
\newcommand{\eqqref}[1]{Eq.\ \ref{#1}\xspace}
\newcommand{\sectref}[1]{Sect.\ \ref{#1}\xspace}

\newcommand{\figrefs}[2]{Figs.\ \ref{#1} - \ref{#2}\xspace}
\newcommand{\refref}[1]{Ref.~#1\xspace}
\newcommand{\tabref}[1]{Tab.\ \ref{#1}\xspace}

\newcommand{\probability}{p}
\newcommand{\matrixrank}{r}
\newcommand{\bvec}[1]{\mathbf{#1}}
\newcommand{\curly}[1]{\mathcal{#1}}
\newcommand{\mean}[1]{\langle #1 \rangle}
\newcommand{\mcmcsteps}{N_{MCMC}}
\newcommand{\numspins}{N}
\newcommand{\latticestate}{s}
\newcommand{\latticespinstate}{[s_i]_{i=0}^\numspins}
\newcommand{\sitespin}[1]{s_#1}
\newcommand{\siteenergy}[1]{\epsilon_#1}
\newcommand{\statespace}{\curly{S}}
\newcommand{\nearestneighbors}{\langle ij\rangle}
\newcommand{\siteneighborhood}[1]{\curly{N_#1}}
\newcommand{\coupling}{J}
\newcommand{\extfield}{h_i}
\newcommand{\hamiltonian}{H}
\newcommand{\temperature}{T}
\newcommand{\criticaltemperature}{T_c}
\newcommand{\lowtemperatures}{T_{low}}
\newcommand{\transitiontemperatures}{T_{trans}}
\newcommand{\hightemperatures}{T_{high}}
\newcommand{\boltzmannconstant}{k_B}
\newcommand{\boltzmannfactor}{b}
\newcommand{\inversetemperature}{\beta}
\newcommand{\observable}{O}
\newcommand{\magnetization}{M}
\newcommand{\energy}{E}
\newcommand{\partitionfunction}{Z}
\newcommand{\densityofstates}{g}
\newcommand{\modelseed}{\text{seed}_\text{model}}
\newcommand{\dataloaderseed}{\text{seed}_\text{batch}}
\newcommand{\model}{f}

\newcommand{\modelparams}{\theta}

\newcommand{\autoencoder}{\model_\modelparams}

\newcommand{\width}{w}
\newcommand{\depth}{d}
\newcommand{\bottleneck}{b}

\newcommand{\rawrepr}{\bvec{x}}
\newcommand{\rawpixel}[1]{\rawrepr_#1}
\newcommand{\rawspace}{\curly{X}}
\newcommand{\rawvalidspace}{\rawspace_\text{valid}}
\newcommand{\trainset}{\rawspace_\text{train}}
\newcommand{\valset}{\rawspace_\text{val}}
\newcommand{\trainstep}{t}
\newcommand{\learningrate}{\lambda}
\newcommand{\batch}{\curly{B}}
\newcommand{\batchsize}{|\batch|}
\newcommand{\trainbatch}{\batch^\text{train}_t}
\newcommand{\batchinput}{\bvec{x}_b}
\newcommand{\batchoutput}{\bvec{\hat{x}}_b}
\newcommand{\msdloss}{\curly{L}^\text{MSD}}

\newcommand{\inputs}[1]{#1^\text{in}}
\newcommand{\outputs}[1]{#1^\text{out}}

\newcommand{\rawinput}{\inputs{\rawrepr}}
\newcommand{\rawoutput}{\outputs{\rawrepr}}
\newcommand{\rawloss}[1]{\msdloss_{\rawrepr\in #1}}
\newcommand{\rawlosslong}[1]{\mean{\msdloss(\rawinput,\rawoutput)}_{\rawrepr\in #1}}
\newcommand{\obsinput}{\bvec{\observable}(\rawinput)}
\newcommand{\obsoutput}{\bvec{\observable}(\rawoutput)}
\newcommand{\observableloss}[1]{\msdloss_{\observable, #1}}
\newcommand{\observablelosslong}[1]{\mean{\msdloss(\obsinput,\obsoutput)}_{\rawrepr\in #1}}

\newcommand{\kernelsize}{k}
\newcommand{\kernelneighborhood}[1]{\curly{K}^\kernelsize_#1}
\newcommand{\numkernelelements}[1]{|\kernelneighborhood{#1}|}
\newcommand{\kernelloss}[1]{\msdloss_{\kernelsize=#1}}

\newcommand{\localmagnetization}[1]{\magnetization^\text{loc}_#1}
\newcommand{\scalemagnetization}{\bvec{\magnetization}(\kernelsize)}
\newcommand{\meanmagnetization}{\overline{\magnetization}}
\newcommand{\magscaleloss}{\msdloss_{\scalemagnetization}}
\newcommand{\magscalelosslong}[1]{\kernelloss{#1}(\inputs{\scalemagnetization},\outputs{\scalemagnetization})}

\newcommand{\localenergy}[1]{\energy^\text{loc}_#1}

\newcommand{\meanenergy}{\overline{\energy}}

\begin{document}

\title{Interpreting learning dynamics of autoencoders: Transient scaling and emerging concepts of the Ising model}

\newcommand{\affilsimtech}{Stuttgart Center for Simulation Science,
Cluster of Excellence EXC 2075,\\ University of Stuttgart,
Universit\"atsstra{\ss}e 32, 70569 Stuttgart, Germany}
\newcommand{\affilwin}{WIN-Kolleg of the Young Academy $\vert$ Heidelberg Academy of Sciences and Humanities, Karlstraße 4, 69117 Heidelberg, Germany}

\author{Max Weinmann}
\affiliation{\affilwin}
\affiliation{\affilsimtech}

\author{Miriam Klopotek}
\email{miriam.klopotek@simtech.uni-stuttgart.de}
\affiliation{\affilsimtech}
\affiliation{\affilwin}

\date{\today}

\begin{abstract}
We study how unsupervised autoencoders trained on microscopic spin configurations from the Ising model learn macroscopic, theory-relevant variables underlying the data-generating process. 
We quantify learning across multiple spatial (coarse-graining) scales and reveal two distinct dynamical regimes that appear sequentially, controlled by the main hyperparameters (model depth, width, and learning rate): one in which magnetization and another in which energy is learned across scales. 
The first exhibits error fluctuations ordered to scale and learns global averages only; 
The second gradually resolves smaller scales relevant for the energy representation. 
Deep models trained at moderate and fast rates become arrested before reaching these regimes. 
We connect reconstruction errors with the latent representations using a novel analysis of self-recursive trajectories. 
These intrinsic dynamics are induced by prediction errors, exposing how training drives representation changes for macroscopic concepts. 
We utilize the intuition that learning operates as a process driven far from equilibrium by fluctuations from the training data to provide an interpretive basis grounded in both the physical world and the machine models that represent it. 
\end{abstract}

\maketitle

\section{Introduction}\label{introduction}
While we have a good understanding of linear methods like linear regression and related applications \cite{pearsonLIIILinesPlanes1901,cohenAppliedMultipleRegression2013,golubSingularValueDecomposition1970}, we lack a comparably deep understanding of the nonlinear (stochastic) methods used to train neural networks (NNs). 
Encouragingly, many nonlinear systems in physics arise from many interacting constituents forming a collective, whose qualitative behavior we can describe well at the macroscopic scale. 
This motivates our attempt to use well-established methods and theoretical principles from physics to explain the nonlinear training dynamics of NNs. 

We focus on studying learning dynamics of deterministic nonlinear autoencoders \cite{tishbyInformationBottleneckMethod2000}, which can be seen as nonlinear analogs of principal component analysis \cite{kramerNonlinearPrincipalComponent1991}. 
The most informative (non-linear) principal components are selected by optimizing the reconstruction loss — comparing input and output — while requiring all information to pass through a bottleneck with fewer dimensions than the input representation. 
This bottleneck limits the number of independent features, driving the discovery of collective properties that reveal strong correlations in the training data. 
From a physical perspective, this suggests that learning behavior in autoencoders will involve coarse-graining operations to reveal theoretically relevant concepts within the compressed latent space.
The discovery of more complex compressed representations of data leads to generalizing abstractions being created, which are indispensable for automating cognitive labor and advancing more capable artificial intelligence. 
This explains why understanding this process has become a prominent field of research 
\cite{bishopPatternRecognitionMachine2006, alpaydinIntroductionMachineLearning2010, mohriFoundationsMachineLearning2012, valiantTheoryLearnable1984, vapnikUniformConvergenceRelative1971, solomonoffFormalTheoryInductive1964, goldLanguageIdentificationLimit1967, vapnikNatureStatisticalLearning1995}. 

A variety of post hoc methods have gained popularity for explaining the final function of trained models \cite{wetzelInterpretableMachineLearning2025} without explaining the acquisition of general abilities, which is our primary interest. 
We aim to understand the influence of commonly tuned hyperparameters on learning dynamics and the underlying concept formation. 

In practice, a range of techniques is employed to improve generalization, including careful parameter initialization \cite{heDelvingDeepRectifiers2015}, training data normalization, stochastic batches \cite{robbinsStochasticApproximationMethod1951, kieferStochasticEstimationMaximum1952, rosenblattPerceptronProbabilisticModel1958}, and strategies based on adjusting learning rates over time \cite{loshchilovSGDRStochasticGradient2017} or depending on gradient fluctuations \cite{kingmaAdamMethodStochastic2014}. 
The interplay of all these components creates a complex system (\sectref{sect:learning-components}) whose dynamics are, in general, neither normal nor regular without hyperparameter tuning \cite{robertsPrinciplesDeepLearning2022a}. 
Understanding the abstract process of machine learning requires us to identify patterns within the complex dynamical details of the learning algorithm.

To explore this, we conduct a large-scale ablation study of unsupervised training for autoencoder models using a synthetically generated dataset (\sectref{sect:data}) that is physically motivated (based on the Ising model described in \sectref{sect:ising-model}). 
We vary the most commonly tuned hyperparameters and assess sensitivity to stochastic elements by repeatedly training the same model with different initializations or training data orders (\sectref{sect:model}, \ref{sect:optimization-algorithm}, \ref{sect:statistics}). 
While analyzing individual models guides our intuition about fine-tuned details, examining ensembles of models with similar hyperparameters reveals broader trends. 

The Ising model is a useful testbed for studying how learning systems capture correlated structure. 
It is defined through high-dimensional, temperature-controlled distributions with nontrivial couplings across arbitrary dimensions and topologies. 
More importantly, it is a physical model system grounded in a substantial body of theoretical work. 
Previous research demonstrated that both unsupervised and supervised methods can recover phase transitions or critical temperatures \cite{wangDiscoveringPhaseTransitions2016,wetzelUnsupervisedLearningPhase2017a,alexandrouCriticalTemperature2DIsing2020,carrasquillaMachineLearningPhases2017,wetzelMachineLearningExplicit2017a}. 
However, these methodologies have typically focused on global order parameters such as magnetization. 
Theories surrounding linear and shallow autoencoders have linked representation learning to principal components and specifics of dynamics caused by bottlenecks \cite{baldiNeuralNetworksPrincipal1989,kuninLossLandscapesRegularized2019,gidelImplicitRegularizationDiscrete2019,saxeMathematicalTheorySemantic2019,refinettiDynamicsRepresentationLearning2023,goldtModelingInfluenceData2020,robertsPrinciplesDeepLearning2022a}. 
This background motivates us to study whether sample reconstruction losses can be decomposed into physical features across training -- especially near criticality, where the Ising data spans broad ranges of magnetization and energy. 
The associated nonlinear dynamics that arise can reveal how physical concepts are learned through an abstraction process that, by rudimentary hypothesis, should bridge scales \cite{dangeloLearningIsingModel2020,yevickVariationalAutoencoderAnalysis2022,wangDiscoveringPhaseTransitions2016,huDiscoveringPhasesPhase2017,yueIncrementalLearningPhase2022}.
Physical theory fundamental to statistical descriptions of matter supports the conception that collective quantities and dynamics are central to critical behavior, and, indeed, in how to coarse-grain or abstract it \cite{Hohenberg1977, Zinn-Justin-book-2007, Kardar-book-2007, Balian1986, Dhont-book-1996, Das2004}. 
We employ this as a conceptual basis to study the learning dynamics across different spatial scales.  
In \sectref{sect:learning-of-observables}, we examine the formation of physical concept representations by monitoring their development, as these are emergent properties not explicitly optimized during training. 
We find that learning dynamics can entail two distinct dynamical regimes when analyzed by spatial scale. 
In each regime, one of the two most important concepts governing the data distribution is learned across scales:  magnetization and energy. 

To complement our observation of the autoencoder's outputs, we study its internal behavior in the most compressed (low-dimensional) latent space, referred to as the bottleneck. 
We leverage the compatibility between an autoencoder's input and output spaces to apply the model recursively to its own outputs (\sectref{sect:representation-dynamics}). 
This recursive process generates trajectories in the model's representation spaces. 
By analyzing their relation to concepts in input space throughout the training process, we gain insights into the formation of internal concepts and how they relate to the physical concepts defined externally. 

Together, these investigations aim to reduce the opacity of machine learning by connecting the collective learning dynamics, as indicated by loss functions, to the development of representations of recognizable concepts. 
Our approach illustrates that a physics-based effective description can encompass the full breadth of phenomenology, including highly non-linear and critical behavior. 
Arguably, these viewpoints establish a foundation for grounding scientific discovery processes in artificial intelligence in physics (\sectref{sect:discussion-conclusion}).

\section{Background}

\subsection{Relevant Components for NN Learning}\label{sect:learning-components}

The learning process for artificial neural networks (NNs) can be decomposed into three main components, each entailing a distinct physical interpretation:

\begin{enumerate}
\item
  The \textbf{data-loader} provides the structure of the function to be learned and how its inputs are represented to the model for processing. 
  It defines the dynamics of the external environment (fluctuations in the data distribution) driving the learning algorithm. 
\item
  The \textbf{model class} specifies the available processing operations and how they must relate to each other. The complexity of learning dynamics is shaped by the relation between node activations and their dependence on inputs and parameters. 
  Parameterized transformations  (i.e., linear layers) specify node connectivity that depends on how node activations depend on each other (i.e., on the previous layer). This is determined by the evolving model parameters, which enable an adaptation through parameter changes and thus form the basis for the self-organization process. 
  Nonlinear activation functions make the interaction strength between activations both input- and parameter-dependent. 
\item
  The \textbf{optimizer} determines how models of the model class are updated in response to their interactions with the dataset. This creates a nonreciprocal feedback coupling between parameters and the `environment' (specified by the data, data-loader, and optimizer). 
  Parameter dynamics can (locally) incorporate past interactions (so-called momentum), introducing memory that can ultimately extend interactions between activations over time, dampen fluctuations, accelerate parameter drift, and affect the model's stability under external driving (i.e., during training on datasets, which have intrinsic fluctuations). 
\end{enumerate}

These components are combined to produce a self-organizing system that automatically finds relationships to represent a system of interest (e.g., the Ising model configurations and their macroscopic and microscopic details).

\subsection{The Ising Model}\label{sect:ising-model}

The Ising model \cite{isingBeitragZurTheorie1925} is a simple model for magnetization that is composed of a grid with local interaction $\coupling_{i,j}$ between binary states $\sitespin{i}\in\{-1,1\}$ and interactions between the states and an external field $\extfield$, which we omit for simplicity.
The probability $\probability(\latticestate)$ of a lattice spin state $\latticestate=\latticespinstate\in\statespace$ is given by the Boltzmann distribution, which depends only on the temperature $\temperature$ of the system and the energy $\energy$ of the particular state.

The energy of a lattice state simplifies to
$$
\energy = \hamiltonian(\latticestate)= -\coupling \sum_{\nearestneighbors} \sitespin{i} \cdot \sitespin{j}
$$
with the spin $\sitespin{i}$ at site $i$ on the lattice and the coupling coefficient $\coupling$ between nearest neighbors, denoted $\nearestneighbors$.
If the coupling coefficient is positive, spins preferably align (ferromagnetic order), while misaligned spin states (paramagnetic order) are energetically preferred when $\coupling <0$.
We assume $\coupling=1$ without loss of generality, as we can swap the configurations with opposite sign energies and thus exchange negative with positive correlations.

The probability of a spin state $\latticestate$ is given by
$$
\probability(\latticestate) =\frac{1}{\partitionfunction(\inversetemperature)} \exp\left(-\inversetemperature \hamiltonian(\latticestate) \right) = \probability(\energy)/\densityofstates(\energy)
$$
with inverse temperature $\inversetemperature=\frac{1}{\boltzmannconstant \temperature}$ and partition function $\partitionfunction(\inversetemperature)=\sum_{\energy} \densityofstates(\energy) e^{-\inversetemperature \energy}$ to normalize the probability distribution according to the density of states over energy $\densityofstates(\energy)$.
Increasing the temperature is effectively reducing the coupling constant $\coupling$ between neighboring sites, while the energy-dependent Boltzmann factor $\boltzmannfactor(\energy|\inversetemperature)=\exp\left(-\inversetemperature \energy \right)$ converges to a uniform distribution, increasing the Shannon entropy over the state space $\statespace$.

\subsubsection{Phase Transition}\label{sect:phase-transition}

The product between the Boltzmann factor $\boltzmannfactor(\energy|\inversetemperature)$ and the density of states over energy $\densityofstates(\energy)$, determining the partition function $\partitionfunction(\beta)$, has a non-linear dependence on the control parameter $\temperature$.
Temperature changes shift the state distribution $\probability(\latticestate|\inversetemperature)=\probability(\hamiltonian(\latticestate))$, which depends only on the state's energy, most rapidly near a critical temperature $\criticaltemperature$, which depends on the system size.
For infinite systems, the density over energies becomes concentrated close to the origin.
This localizes most changes in the state distribution around $\criticaltemperature \approx 2.269$, where the peak in $\densityofstates(\energy)$ competes with the suppression by $\boltzmannfactor(\energy)$ most evenly.
This can result in a divergence of observables or their derivatives if they change on average across configurations that quickly become more probable near $\criticaltemperature$.
In finite systems, these sharp transitions broaden and shift to higher or lower temperatures depending on the observable.

More generally, the shift of the state distribution $\probability(\latticestate)$, resulting from temperature changes, translates to distributions $\probability(\observable)$ of related observable $\observable(\latticestate)$. 
How this observable distribution changes as a function of temperature depends on the joint distribution $\probability(\observable,\energy)$ between the observable and the energy. 
This is theoretically true for any observable, as the probability of the respective states $\latticestate$ having an observable property only depends on their energy. 
The expectation values of observables such as the absolute total magnetization $|\magnetization(\latticestate)|=|\sum_i s_i|$ change rapidly around the critical temperature, as $\probability(\magnetization,\energy)\cdot\probability(\energy)$ converges to a symmetric distribution with respect to $\magnetization$.
The absolute magnetization $|\magnetization|$ is one of many properties that can be used to define a phase transition from mainly homogeneous/ordered states to inhomogeneous/unordered states in the Ising model.
In general, the distribution of observables depends in complex ways on the state distribution of the Ising model that can't be reduced to a single phase transition.
Here, we focus on magnetization and energy, for which a single phase transition is sufficient to describe the changes in $\probability(\energy), \probability(\magnetization)$.

\section{Methods}\label{sect:methods}

\begin{figure}
    \centering
    \includegraphics[width=1\linewidth]{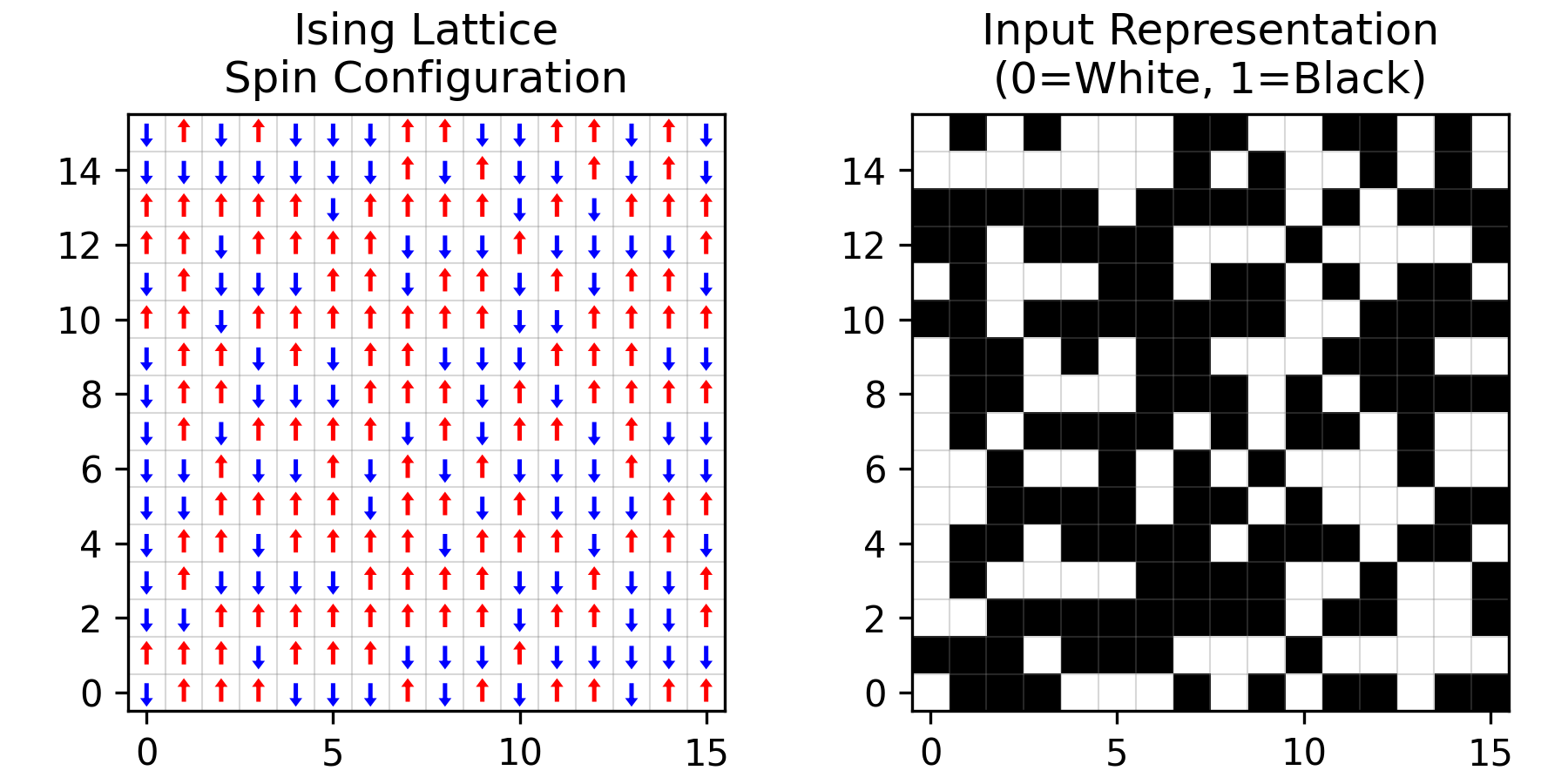}
    \caption{
    Ising model spin configurations on a square lattice (left) are mapped to a binary representation (right). 
    Spins pointing down (blue) are mapped to zero (white) and spins pointing up (red) to one (black). 
    The two-dimensional binary representation is converted to floating-point numbers and flattened into a one-dimensional vector. 
    This one-dimensional pixel vector serves as input to an autoencoder. 
    }
    \label{fig:ising-lattice}
\end{figure}

We study the learning dynamics of autoencoders on compression tasks.
The goal of such tasks is to derive a lower-dimensional representation from given input data that is sufficient to approximately reconstruct the inputs.
The input data is a one-dimensional vector corresponding to a two-dimensional grid, as illustrated in \figref{fig:ising-lattice}.
The output data has the same shape and can be interpreted as a two-dimensional grid of spins.

The quality of the reconstruction is measured by the mean-squared difference (MSD) between the input and output pixels.
The pixel-wise MSD is used for the optimization and is referred to as (raw data) loss.

The analysis was made feasible by limiting it to the following classes:

\begin{enumerate}
\item
  Training on 256-dimensional binary encoded data with a two-dimensional, globally uniform, correlation structure of variable strength
\item
  Deterministic autoencoders with ReLU activations up to 1M parameters per layer and 33M parameters in total
\item
  Self-supervised Stochastic Gradient Descent using the \emph{Adam} optimizer, combined with learning rate scheduling
\end{enumerate}

While every hyperparameter can affect learning dynamics, only a few are routinely changed to improve model quality.
To make our insights more transferable and the analysis more feasible, we use commonly used default values whenever possible.

\subsection{Data}\label{sect:data}

We chose a physical model system — the Ising model in two dimensions — to generate states according to a nontrivial probability distribution.
This allows generating datasets with predefined correlation patterns and a correlation strength conditioned by a control parameter.
The resulting datasets have different effective sample sizes and Shannon entropies, resulting in variable difficulty for data compression tasks.

\subsubsection{Data Generation}\label{sect:data-generation}

The training and evaluation data are generated using Markov Chain Monte Carlo with Glauber Dynamics \cite{glauberTimeDependentStatisticsIsing1963}, according to the Ising model (see \sectref{sect:ising-model}).
This generates sample configurations of a square lattice (see \figref{fig:ising-lattice}) at different temperatures.
The sampling process starts with a random initial configuration and then attempts to flip one randomly selected spin in each step.
The transition probability to accept this flip is given by
$$
\probability(\latticestate^{(1)} \rightarrow \latticestate^{(2)}) \propto \frac{\probability(\latticestate^{(2)})}{\probability(\latticestate^{(1)})}
= \exp\left(-\beta [\hamiltonian(\latticestate^{(2)})-\hamiltonian(\latticestate^{(1)})] \right).
$$
This produces samples distributed according to the Boltzmann Distribution in the limit.
However, the initial samples are biased by the initialization, and samples are generally highly correlated when close together in the sequence, reducing the effective sample size.
To ensure the training data samples originate from sufficiently equilibrated and uncorrelated events for all chosen temperatures, only every $\mcmcsteps=10^6$-th sample is kept for the following analysis. 
In addition, multiple initialization seeds are used to avoid biases in the total magnetization $\magnetization$ at low temperatures, where the sign-flip probability of $\sitespin{i} \in \latticestate$ diminishes drastically.

\subsubsection{Datasets}\label{sect:datasets}

A 16x16 square lattice with $2^{256} \approx 10^{77}$ possible states is used to mimic the complexity of real-world problems while keeping good approximations of the probability distributions for observables tractable, for which no closed-form solution is known.
Each dataset contains 2000 samples and is defined by an initialization seed and a temperature.

We sampled datasets for 15 different temperatures, of which 5 are below, 5 around, and 5 above the phase transition of the infinite Ising model:

\begin{itemize}
    \item Low temperatures \\$\lowtemperatures=\{1.0, 1.4, 1.8, 1.9, 2.0\}$
    \item Transition temperatures \\$\transitiontemperatures=\{2.1, 2.2, 2.27, 2.3, 2.4\}$
    \item High temperatures \\$\hightemperatures=\{2.5, 2.6, 3.0, 5.0, 10.0\}$
\end{itemize}

\begin{figure*}
    \centering
    \includegraphics[width=0.75\linewidth]{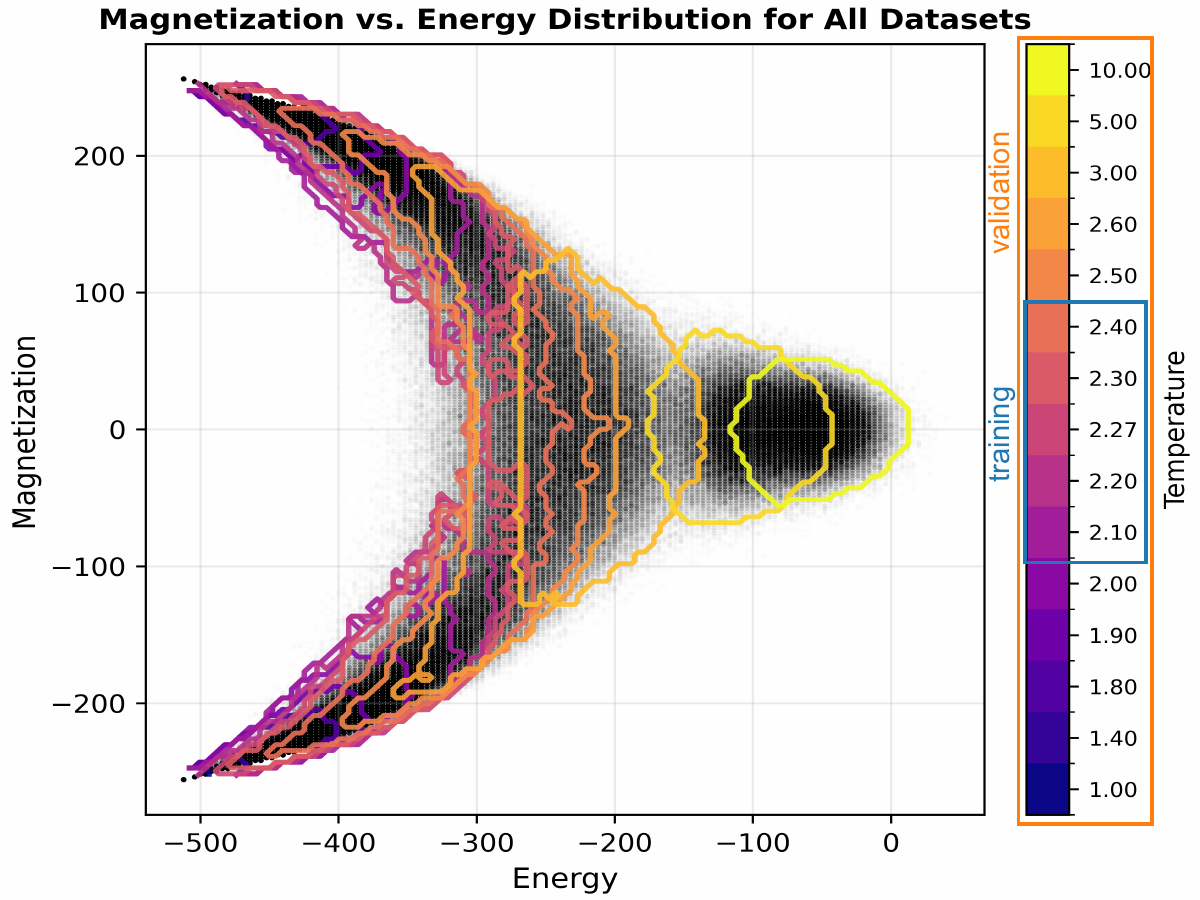}
    \caption{
    Joint distribution of the entire dataset (black dots) over sample energy and magnetization. 
    The area containing almost all samples from a given temperature is indicated by colored contours. 
    At low temperatures, there are two distinct clusters with low energy and high absolute magnetization (upper and lower left). 
    At high temperatures, a single cluster with a small absolute magnetization and higher energy becomes dominant (center-right).
    }
    \label{fig:dataset-contours}
\end{figure*}

The joint magnetization-energy distribution for all datasets (all initialization seeds and temperatures) is visualized in \figref{fig:dataset-contours}.

Only the transition temperatures are used for training.
In total, 34 different initialization seeds are used: 30 for training, 2 for validation, and 2 for hypothesis testing.
The combined datasets contain 300,000 training samples (5 temperatures, 30 seeds, 2000 samples) and 60,000 validation and test samples (15 temperatures, 2 seeds, 2000 samples).

\subsection{Model}\label{sect:model}

The autoencoder model is composed of two symmetric components, an
\textbf{encoder} and \textbf{decoder} \cite{NIPS1993_9e3cfc48}. 
The input is connected to the encoder, which produces a latent representation, which is then fed to the decoder, which generates the final output used during optimization. 

\begin{figure}
    \centering
    \includegraphics[width=\linewidth]{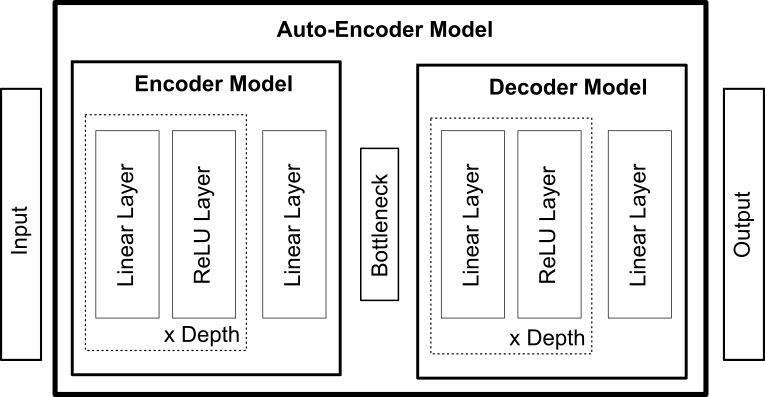}
    \caption{
    Autoencoder model architecture composed of an encoder (left) and decoder model (right). 
    The encoder transforms the input with a sequence of linear and non-linear ReLU layers into a lower-dimensional latent representation, referred to as a bottleneck. 
    The decoder transforms this latent representation with the same sequence of layers (but different parameters) into the output. 
    The parameters of each linear layer are optimized during training to reduce the mean squared deviation (MSD) between inputs and outputs.
    }
    \label{fig:auto-encoder}
\end{figure}

The encoder and decoder used in this study are composed of sequential layers alternating between affine transformations (linear layers) and the element-wise application of nonlinearities multiple times (across the model depth), followed by a final affine transformation.
The architecture is visualized in \figref{fig:auto-encoder}.
All layers in the encoder and decoder have the same number of hidden dimensions (i.e., the model width) and use ReLU (rectified linear unit) nonlinearities. 
ReLU is a popular choice for deep non-linear autoencoders because it is simple, fast, piecewise linear, and well studied \cite{householderTheorySteadystateActivity1941}. 
The final layer of the encoder and decoder has no nonlinearity.

The model's size is adjusted in three ways that affect its representational capacity and learning behavior:
\begin{itemize}
    \item The \textbf{bottleneck size} $\bottleneck$ limits the amount of information that can be transferred from input to output. 
    We only consider dimensions smaller than the input dimensions, which makes a perfect linear reconstruction impossible.
    \item The \textbf{model width} $\width$ defines the number of parallel nodes in each layer, resulting in a block structure. 
    We consider widths that are 1x, 2x, and 4x the input dimension, allowing the model to produce increasingly diverse projections.
    \item The \textbf{model depth} $\depth$ defines how often non-linear transformations are applied before and also after the low-dimensional bottleneck.
    We consider depths of 1, 4, and 16 that allow the representation of increasingly complex encoding and decoding functions.
\end{itemize}

Details on the motivation and constraints that guided this selection are given in \sectref{sect:appendix-training-configuration}

\subsection{Optimization Algorithm}\label{sect:optimization-algorithm}

We use Stochastic Gradient Descent (SGD) with the \emph{Adam} optimizer \cite{kingmaAdamMethodStochastic2014} and a learning rate scheduler to train the autoencoder models on Ising model configurations sampled at transition temperatures.
The optimization objective is the mean-squared error loss, denoted $\msdloss \left( \batchinput,\batchoutput \right)$, which is a feature defined externally to the neural network state itself. 
The function measures the difference between the training batch inputs, $\batchinput$, and the outputs, $\batchoutput$.
We explore various hyperparameters, primarily focusing on variations in batch size ($\batchsize=30$, $300$, $3000$ data points) and learning rate ($\learningrate=10^{-5}$, $10^{-4}$, $10^{-3}$). 
Additionally, we use a learning rate scheduler with a short linear warm-up followed by cosine annealing to stabilize the learning dynamics. 
Details on the motivation for this selection and the relevance of these hyperparameters can be found in \sectref{sect:appendix-training-configuration}
Unless otherwise specified, the default \emph{PyTorch} implementation, version 2.10, with its standard configurations, was used \refref{\cite{anselPyTorch2Faster2024}}.

\subsection{Statistics}\label{sect:statistics}

The optimization algorithm has two sources of variation that can influence the learning dynamics, even when all hyperparameters are unchanged. 
Both sources are determined by seeding a pseudo-random number generator.
Firstly, the model parameters are initialized by sampling from a uniform distribution, as outlined in \cite{kingmaAdamMethodStochastic2014}. 
This approach addresses depth-scaling issues associated with ReLU activations. 
Altering the model seed changes the initial model parameters, thereby affecting how they evolve under identical external influences.
Secondly, training batches are created by permuting the training dataset and then splitting it. 
Changing the data loader seed changes the composition and order of data points in training batches throughout the learning process, thereby altering the external influences on training.
By independently varying both the model seed and the data seed, we can estimate the effects of the initial model parameters and the data ordering on the quasi-critical learning dynamics. These effects contribute to the variance of the metrics calculated across runs (i.e., their loss trajectories), which are analyzed in \sectref{sect:learning-of-observables}.

\section{Results}\label{sect:results}

In the following, we show how microscopic optimization behavior can be understood through a small set of macroscopic observables, allowing for a simplified description of learning dynamics. 
We compare the evolution of the optimization loss at the microscopic (pixel-wise) level with macroscopic reconstruction errors computed from spatially coarse-grained averages across multiple scales (3x3, 5x5, ... blocks). 
These block-averaged observables are related to natural order parameters of the Ising model data (see \sectref{sect:ising-model}) and provide a compact, physically grounded description of the fields \cite{Kardar-book-2007} that the autoencoder is trying to reproduce.
Physically, coarse-grained averages suppress fluctuations, so comparing scale-resolved errors pinpoints the spatial scale at which fluctuations of the microscopic error are largest.
We emphasize how the learning rate $\learningrate$ and model size (defined by width $\width$, depth $\depth$, and bottleneck $\bottleneck$) correlate with these scale-dependent learning trajectories, and how representation dynamics relate to macroscopic observables of the Ising model.

The learning dynamics are analyzed by using three different coarse-graining steps.
These summarize the representation evolution during 972 runs across hyperparameter variations of all 256 elements (pixels) making up each of the 60/300k generated samples inside the training/validation dataset:

\begin{itemize}
    \item Samples are described through physical observables (e.g., magnetization, energy) that are real-valued functions depending on the elements $\rawpixel{i}$ (pixels).
    \item Sample descriptions are averaged over the training or validation dataset. 
    \item Dataset averages of sample descriptions are computed across multiple runs that share one or more hyperparameters (e.g., learning rate, model depth) to reveal trends.
\end{itemize}
Training evolution plots show averages across all runs that match the explicitly given variables, unless otherwise stated.

First, the evolution of the mean squared deviation between raw inputs and outputs is analyzed in \sectref{sect:learning-raw-data}, which is directly connected to the optimization process. 
Then, \sectref{sect:learning-of-observables} introduces the relationship between learning this `raw data' and learning physical observables. 
The latter is analyzed at different spatially averaged scales of the Ising lattice, to connect microscopic optimization details to learning of macroscopic concepts. 
Additionally, the evolution of the joint output distribution of magnetization and energy is measured in \sectref{sect:output-distribution-dynamics} to reveal how the models' concept representations relate and where they fall short. 
Finally, the learning dynamics of physical observables are connected to the evolution of compressed internal model representations in \sectref{sect:representation-dynamics}, to extend the analysis of concept formation from the outside towards the inside of the model.

\subsection{Learning Viewed from Raw Pixel Data (Spins)}\label{sect:learning-raw-data}

Concept formation is driven by the microscopic optimization process governing the learning dynamics. 
The optimization process aims to minimize the mean squared deviation (MSD) loss
$$
\rawloss{\rawvalidspace}=\rawlosslong{\rawvalidspace}
$$ 
between raw inputs $\rawinput \in \rawvalidspace \subset \rawspace$ and corresponding outputs $\rawoutput=\autoencoder(\rawinput)$ over all training batch samples $\rawinput \in \trainbatch$.
The pixels $\rawpixel{i}=(\sitespin{i}+1)/2$ of the raw representation $\rawrepr \in \rawspace$ correspond to (a normalized form of) the physical spins $\sitespin{i}$ and therefore can be related to other theoretical concepts of the Ising model. 
We consider a learning algorithm to be successful at learning an observable $\observable(\rawrepr)$ over a domain $\rawvalidspace$ if 
$$
\observableloss{\rawvalidspace} = \observablelosslong{\rawvalidspace}
$$ 
is consistently reduced over $\rawvalidspace$ and not only temporarily reduced for a specific batch $\trainbatch \subset \trainset$.
Practically, we estimate the loss over $\rawvalidspace$ by using an independently seeded validation set $\valset \subset \rawvalidspace$ spanning different temperatures, as it is infeasible to compute the loss over all $2^{256}$ valid states.

The evolution of the raw validation loss $\rawloss{\valset}$ across training steps $\trainstep$ reveals different dynamic regimes that depend on the model size $S(\width,\depth,\bottleneck)$ and optimization hyper-parameters ($\learningrate, \batchsize$). 
These learning regimes are characterized by transient states that maintain the same loss for varying durations, and transitions with diverse shapes and ranges typically involving smooth decays to lower losses (resembling sigmoid functions).
Important for interpretation is how the learning dynamics of the optimized raw loss $\rawloss{\trainset}$ are closely related to the loss of other observables $\observableloss{\trainset}$ like the magnetization $\magnetization$ and energy $\energy$ of the Ising model.
The study of this micro-macro connection is the focus of this work. 

All input pixels $\rawinput_i$ are either 0 or 1, while the outputs $\rawoutput_i$ are continuous (floating-point numbers) and not explicitly restricted to any value range.
When the loss $\rawloss{\rawvalidspace}$ is monotonically decreasing, the outputs become, on average, more similar to the discrete inputs with each training batch $\trainbatch$.
Learning thus involves confining outputs to an interpolation range (0-1), to approximate the two discrete values (0,1).
If the input pixels \emph{can't} be resolved sharply, the learned outputs $\rawoutput_i$ can represent interpolated values, like 0.5, to coarse-grain the input-dependent pixel value distribution (peaks at 0 and 1) and minimize $\rawloss{\rawvalidspace}$ under this constraint.
When the output space is sensitive to the input pixels, the latter \emph{can} be resolved.
This is because the relevant model parameters become susceptible to gradient descent. 
Therefore, outputs increasingly mimic the discrete input values rather than their distribution averages. 
How quickly the model $\autoencoder$ adapts depends on the local gradient of $\rawloss{\trainset}$ with respect to the model parameters $\modelparams$.
Therefore, the raw loss evolutions provide valuable insights into the self-organization process that adapts the models over time, resulting in different observables being learned. 

Generally, the loss for observables $\observableloss{\rawvalidspace}$ that are not explicitly optimized, such as physical observables, can fluctuate even when the optimized loss $\rawloss{\trainset}$ remains constant. 
Consequently, a reduction in $\rawloss{\valset}$ does not necessarily indicate a decrease in the mean squared discrepancy/deviation (MSD) $\observableloss{\valset}$ of other observables.
However, in theory, lower MSD optimization losses often reduce the variance of other convex loss functions, with the variance approaching zero when the optimization loss is zero. 
This correspondence is useful for understanding the dynamics of learning for other observables. 
The exact relationship depends on the joint distribution of pixel values and the respective observable. 
However, this is not tractable in practice, which is why we rely on simpler proxies, as discussed in \sectref{sect:learning-of-observables}. 

In the following, we examine the different dynamical regimes in the raw data loss $\rawloss{\valset}$. 
Our focus will be on the model's behavior on validation data, sampled independently of the training data used during optimization, to estimate generalization. 
The loss evolution for the training data is included in the supplementary data \cite{DARUS-6128_2026} for completeness. 

\begin{figure*}
    \centering
    \includegraphics[width=0.9\linewidth]{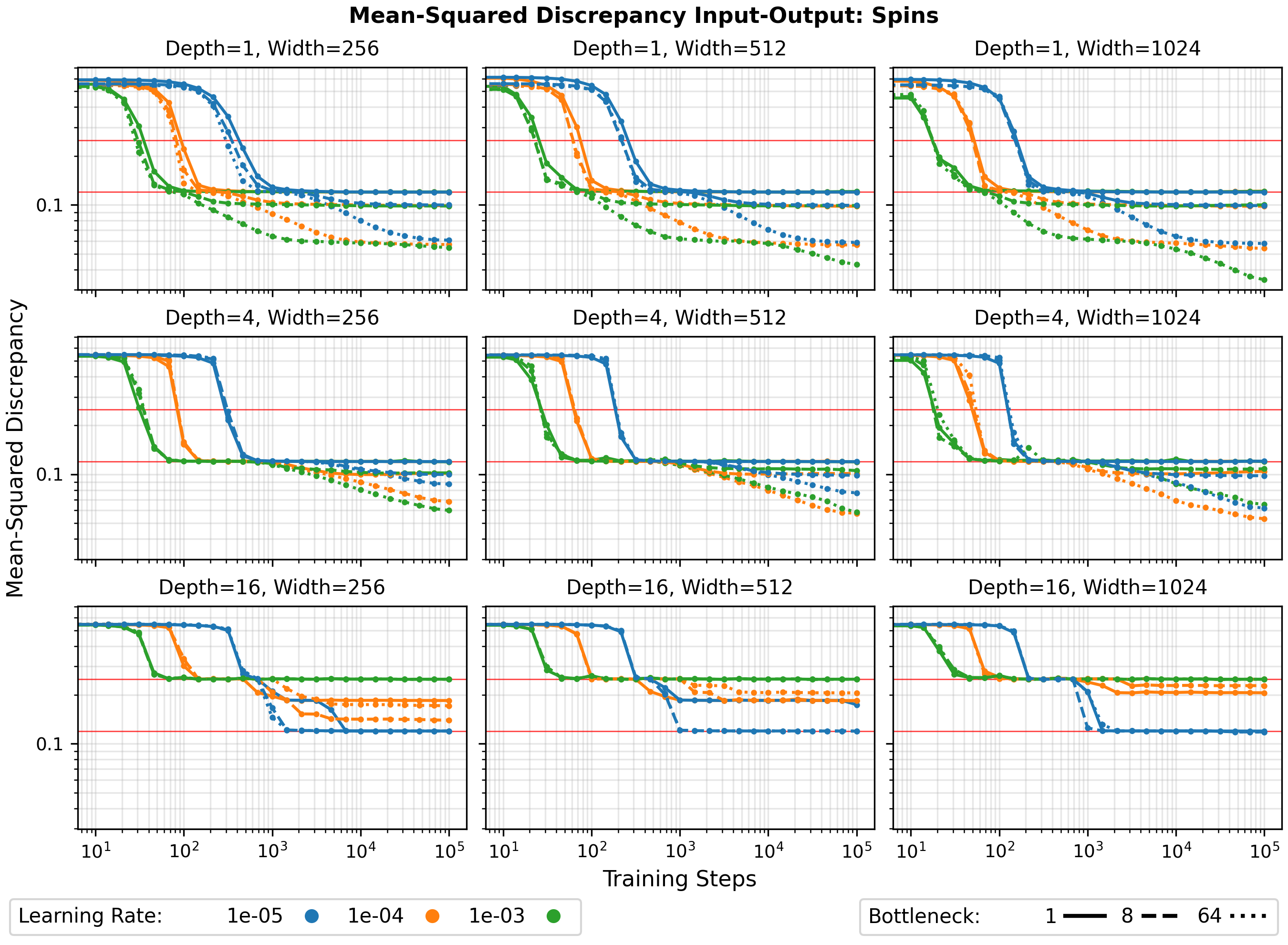}
    \caption{
    Validation loss evolution of raw data over training checkpoints averaged over all runs with the same model (depth, width, bottleneck) and learning rate. 
    Each subplot shows the average across different seeds ($\modelseed \in \{0,1\}, \dataloaderseed \in \{0,1\}$) and batch sizes $\batchsize \in \{30,300,3000\}$ for a given model width and depth. 
    The line styles correspond to different bottleneck sizes, while the line colors indicate the base learning rate used for training. 
    All loss curves show a rapid initial decrease, followed by none or another rapid or slow decrease depending on the hyperparameters. 
    The MSD plateau values are equivalent to the loss obtained by predicting the training data average ($\msdloss \approx 0.25$) and corresponding sample averages ($\msdloss \approx 0.1$), respectively, indicated by the red horizontal lines. 
    The individual loss curves used to compute the averages are shown in the appendix (\figref{fig:raw-msd-per-run}).
    }
    \label{fig:raw-msd}
\end{figure*}

The average evolution of validation losses, shown in \figref{fig:raw-msd}, indicates that the different dynamical regimes depend on the optimization hyperparameters. 

Models with the smallest bottleneck ($\bottleneck=1$, represented by the solid line in \figref{fig:raw-msd}) exhibit similar behavior across different depths and widths, except for an intermediate plateau noted in the deepest models. 
This suggests that within the tested parameter range, the encoder and decoder widths do not significantly affect the learned observables. 
Essentially, the single degree of freedom is primarily utilized to reconstruct averages, as will be explored in \sectref{sect:learning-of-observables}, which explains why increasing width provides no benefits. 

Depth begins to negatively influence learning dynamics only in the deepest models. 
There, convergence to the same loss value $\msdloss \approx 0.1$ takes longer or becomes arrested at a higher level for faster learning rates. 
The latter plateau is worse than predicting the average of individual samples. 
It is instead close to reconstructing only a single input-independent output (i.e., the same dataset input average at each pixel of each sample). 
Predicting the average is not particularly difficult, as it only requires multiplying each of the $n$ inputs by $1/n$ and summing the contributions, which is achievable with the simplest conceivable linear model. 
As this loss threshold was surpassed by all tested shallow models, we conclude that these deep models fail to learn any meaningful function of the input. 
This suggests that adding excessive depth reduces a model's capacity to efficiently learn input-dependent representations across all bottleneck sizes. 
The theoretical derivation in \cite{robertsPrinciplesDeepLearning2022a} describes a similar regime for overly deep models, where input signals decay or explode depending on the initialization. 

Potentially, depth introduces a stronger coupling of outputs, resulting in highly non-linear relaxation pathways (gradient flow) due to a rugged loss landscape (over parameter space). These require small learning rates for stable convergence. 
Conversely, a high learning rate results in strong interaction with the data, which could push the model past narrow minima, preventing relaxation.
Such an effect is observable for the deepest models (lower row) in \figref{fig:raw-msd}, \ref{fig:spin-pooling}, and \ref{fig:energy-pooling}.
The (critical) effective learning regime \cite{robertsPrinciplesDeepLearning2022a}, outside of which the model becomes insensitive or parameters explode, might shrink with depth, making successful learning depend on fine-tuned hyperparameters. 
If very slow base learning rates are required to maintain stability, this creates a threshold beyond which models are too deep to learn efficiently within a fixed computational budget. 

In the following sections, we will see how this state observed in deep models trained at nonzero learning rates manifests as a failure to form representations of macroscopic physical concepts. 
We point out that our study doesn't make use of techniques to improve learning stability, which are common practice when training very deep models (e.g., layer and batch normalization, gradient clipping, skip connections).

For the shallower models (specifically, $\depth=1$ and $\depth=4$), larger bottleneck sizes ($\bottleneck=8$ and $\bottleneck=64$) lead to lower validation loss, consistent with the expectation that increased representation capacity improves performance. 
The loss curves for the shallow models ($\depth=1$) display two additional consistent plateaus beneath the first plateau. 
One plateau is located just below the best final loss for the smallest bottleneck size ($\bottleneck=1$, $\msdloss \approx 0.1$), while the other is around the best final loss for the thinnest models ($\width=256$, $\msdloss \approx 0.06$). 
Importantly, the same plateaus appear across all model widths, suggesting that the key dependencies lie in the data and model depth. 
We note that extended training may reveal additional loss plateaus and lead to a more substantial decline in loss. 

This complex behavior may be linked to trade-offs between the number and resolution of properties that can be represented in the latent space. 
As we will see, there are trade-offs between representing different physical concepts in \sectref{sect:spin-pooling}. 
This is particularly relevant given the intermediate confinements imposed by the lower-dimensional bottlenecks on the full representation space. 
We expect a decrease in loss as more fluctuations are successfully encoded and decoded by the autoencoder and thus pass through the bottleneck. 
The extended plateaus, followed by rapid loss drops, signal sharp regime shifts in the effect that dominates the learning dynamics. 
One phase in which the external driving mainly results in chaotic fluctuations exploring the parameter space, and another where collective fluctuations result in a consistent loss reduction. 
The onset of the transition can be seen as the system suddenly uncovering a new feature dimension, followed by an alignment phase (scaling/translation) that precisely aligns this new dimension with the target output. 
We will connect these regimes to internal latent representation dynamics, which determine the changes in output, in \sectref{sect:representation-dynamics}. 

Interestingly, the deepest models ($\depth=16$) with the largest bottleneck ($\bottleneck=64$) exhibit a higher average validation loss than those with a medium-sized bottleneck ($\bottleneck=8$) when trained with a moderate learning rate ($\learningrate=10^{-4}$). 
This contrasts with the expectation that greater representation capacity (larger bottleneck) should facilitate learning. 
This behavior is also evident in the average training loss (\figref{fig:raw-msd-training}), suggesting that overfitting to the training data is not a contributing factor.
Instead, this phenomenon seems to result from the interplay between the data interaction strength, mediated by the learning rate, and activation averaging, increasing with width (and depth).
However, the average evolution over runs provides only limited insight into underlying dynamics.
The evolution of individual runs in \figref{fig:raw-msd-per-run} indicates a rapid transition from the intermediate to a lower loss plateau after step $\trainstep=200$, although the timing of this transition varies between runs. 
Moreover, this transition occurs only for a fraction of the deepest model runs and is notably rare in cases with larger bottlenecks, often delayed for wider models.
This suggests that the underlying process is chaotic, as expected in overly deep models \cite{robertsPrinciplesDeepLearning2022a}. 
The different timing of transitions will manifest in temporarily increased variance between runs for the macroscopic properties studied in \sectref{sect:learning-of-observables}. 
In contrast, we will see that the variance remains high in the chaotic regime. 

Learning from the raw data becomes increasingly challenging for deeper models, particularly as the width and bottleneck size grow. 
Possible explanations are discussed in \sectref{sect:discussion-conclusion}. 
It remains uncertain whether more training data, such as higher energies to allow for greater input diversity, different initialization strategies, or extending training time, could enhance learning. 

Before discussing the macroscopic description, we zoom in on the microscopic evolution details of individual samples that comprise the previously investigated MSD.
Since we can't visualize all samples, we focus on the highest-loss samples to highlight those patterns that dominate reconstruction errors. 

\begin{figure*}
    \centering
    \includegraphics[width=0.75\linewidth]{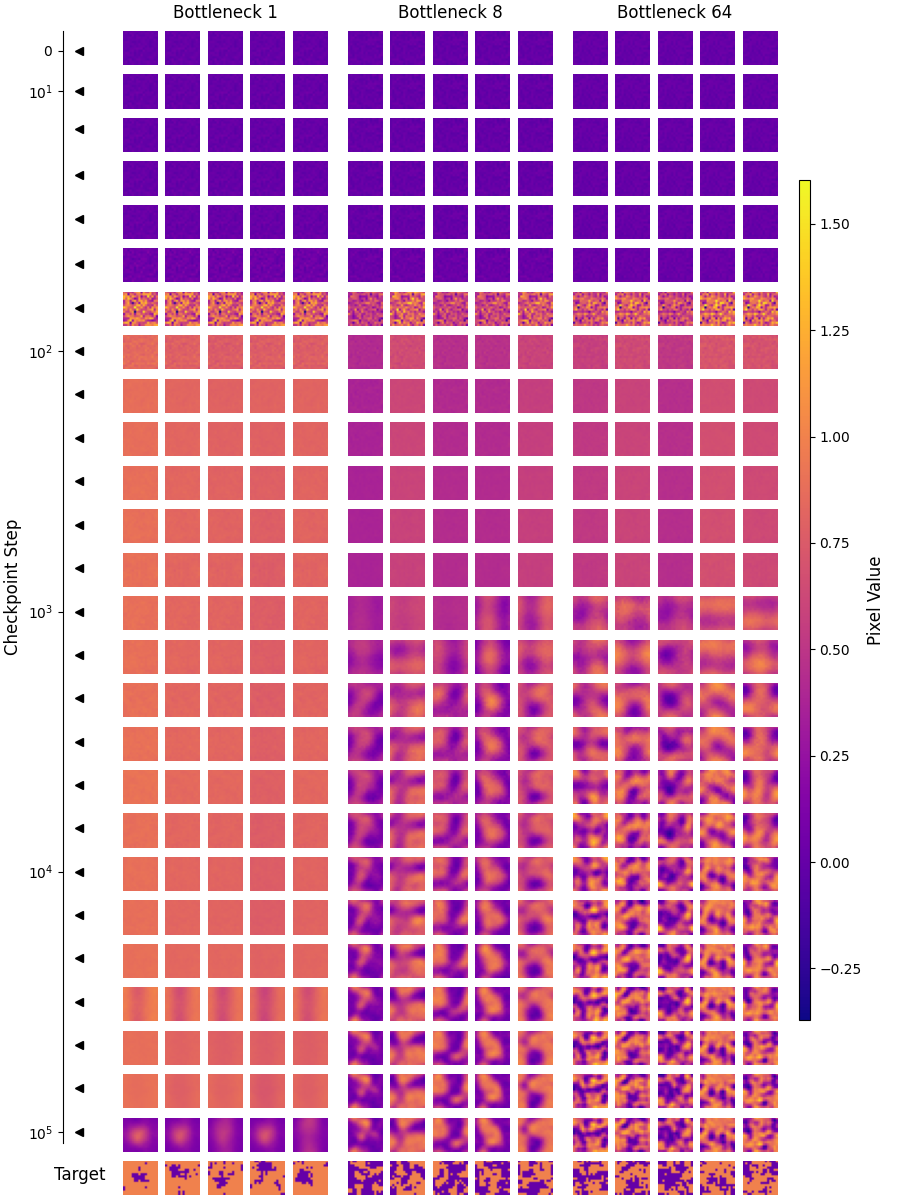}
    \caption{
    Model output evolution for samples from the critical temperature with the highest final loss for different bottlenecks. 
    Samples with the lowest loss from the same run are shown in \figref{fig:low-loss-outputs}.
    Each column of 5 samples shows the temporal evolution of one model, from bottom to top. 
    The last row shows the input, which is the target of the optimization. 
    The model checkpoints are taken from runs \textbf{MB1, MB8, MB64} (see \tabref{tab:run-names}), all trained with the most typical model configuration: depth $\depth=4$, width $\width=512$, and a medium learning rate $\learningrate=10^{-4}$. 
    The output for all samples starts around zero, where it remains until a rapid transition with fluctuations between pixels, after which it converges to uniform values between 0 and 1. 
    These fluctuations resemble an amplified version of the minimal initial fluctuations, indicating the dominance of collective dynamics. 
    Outputs from a small bottleneck converge to the dominant value (rather than the average) and change little, except at the end, when a non-uniform pattern emerges that is very different from the input, suggesting poor generalization. 
    Larger bottlenecks converge towards the mean and then reveal progressively finer details that match an increasingly less-blurred version of the input. 
    }
    \label{fig:high-loss-outputs}
\end{figure*}

\begin{figure*}[ht]
    \centering
    \includegraphics[width=0.75\linewidth]{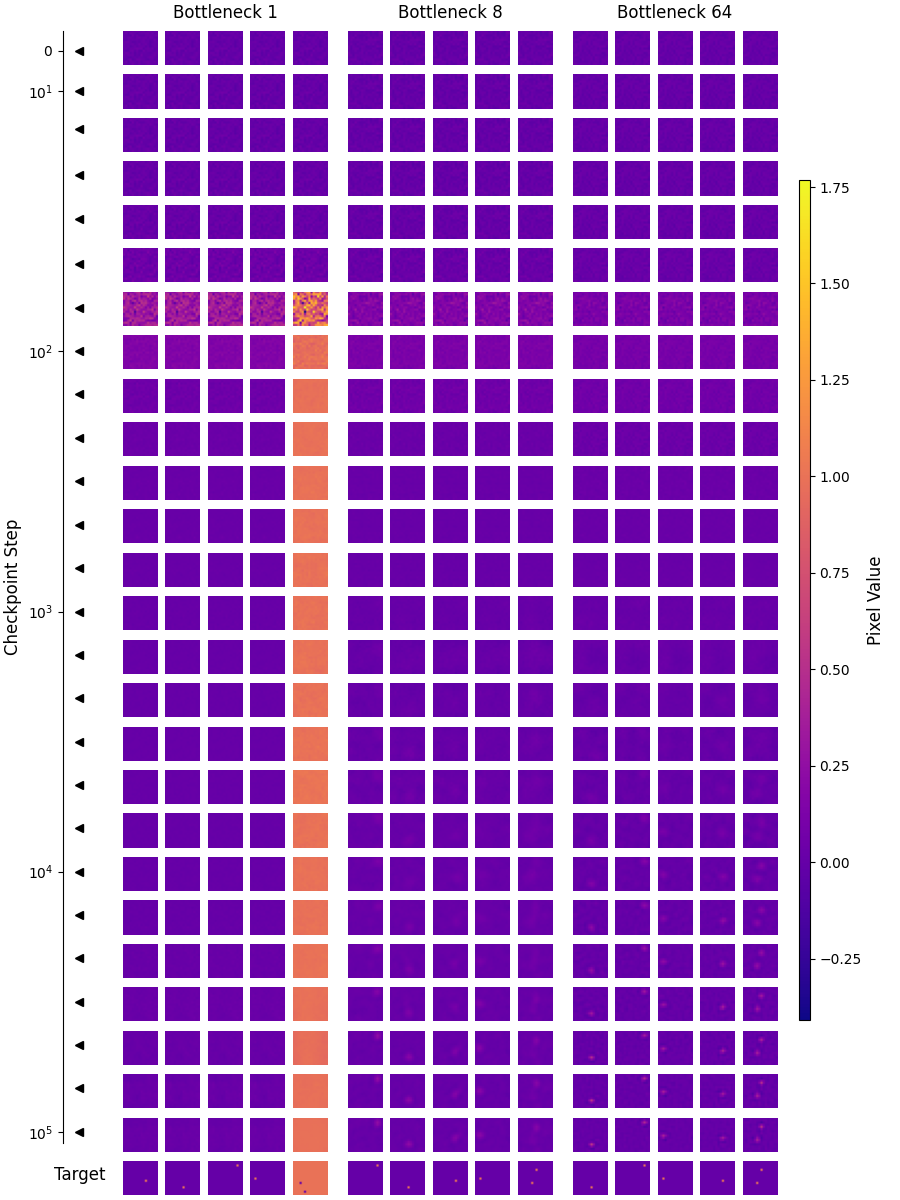}
    \caption{
    Model output evolution for samples from the critical temperature with the lowest final loss for different bottlenecks. 
    Samples with the highest loss from the same run are shown in \figref{fig:high-loss-outputs}.
    Each column of 5 samples shows the temporal evolution of one model, from bottom to top. 
    The last row shows the input, which is the target of the optimization. 
    The model checkpoints are taken from runs \textbf{MB1, MB8, MB64} (see \tabref{tab:run-names}), all training the most typical model configuration of depth $\depth=4$, width $\width=512$ with a medium learning rate $\learningrate=10^{-4}$. 
    The output for all samples starts around zero, where it remains until a rapid transition, with fluctuations between pixels, converges to uniform values between 0 and 1. 
    These fluctuations resemble an amplified version of the minimal initial fluctuations.
    Outputs from a small bottleneck converge to the mode or mean uniformly. 
    Larger bottlenecks converge towards the mean and then show progressively finer details that match an increasingly less blurred version of the input. 
    The final output is sensitive to single pixels but doesn't resolve them sharply. 
    }
    \label{fig:low-loss-outputs}
\end{figure*}

\figref{fig:high-loss-outputs} illustrates the pixel-wise evolution of reconstructed outputs for several samples with the highest loss for three runs with different bottleneck sizes, all with a medium learning rate $\learningrate=10^{-4}$ and the most typical model configuration of depth $\depth=4$, width $\width=512$. 
Other evolutions are shown in \sectref{sect:appendix-reconstruction-evolution} and included in the supplementary data \cite{DARUS-6128_2026}. 
Across runs, the sample pixels $\rawpixel{i}$ remain close to zero for several initial time steps. 
Then they undergo a rapid transition, showing small-scale spatial fluctuations, to a distinct, globally uniform value distribution. 
The small-scale fluctuations are similar to the minimal initial fluctuations, indicating the dominance of contrast-enhancing collective dynamics. 
For the smallest bottleneck $\bottleneck=1$, the values converge to the dominant pixel value (orange $\approx 1$) until local differences on large scales appear, followed by an almost inverted reconstruction in the very final stage.
However, this inversion effect was observed only for high-loss samples, which are particularly informative of overfitting. 
Low-loss samples of the same run (\figref{fig:low-loss-outputs}) remain spatially uniform.
For the larger bottlenecks $\bottleneck=8,64$, the first transition appears to end with pixel values all close to the global sample mean before becoming correlated with the mean at increasingly smaller scales.
This suggests a transition in the learning dynamics from global (macro) to local (micro) features.
The convergence with the smallest bottleneck to the dominant sample value (mode) rather than the more optimal sample mean suggests a bias towards a spatially uniform bimodal distribution. 
This bias could result from approximately uniform samples dominating the training data due to their lower energy, combined with a resolution-limiting factor, such as the mixing of independent input fluctuations. 
The bimodal distributions are not observed in narrower networks of the same bottleneck size (see supplementary data \cite{DARUS-6128_2026}), confirming that the model width plays a relevant role.
To systematically compare behavior, we introduce a set of spatially coarse-grained observables in the following. 
These are sensitive to fluctuations at different scales occurring across samples and individual runs, allowing for a macroscopic description.

\subsection{Learning Viewed from Physical Observables (Concepts)}\label{sect:learning-of-observables}

The mean squared deviation (MSD) of the raw data alone is not sufficient to capture the learning dynamics of other observables. 
Therefore, we study in detail the learning dynamics of physical observables that are theoretically important for the data-generating process. 

We analyze how learning depends on the spatial scale of coarse-grained physical observables, such as the mean magnetization and energy, and connect these to the learning dynamics observed in raw pixel data, as previously discussed in \sectref{sect:learning-raw-data}. 
Furthermore, we show that these observables represent a local first-order spatially isotropic Taylor expansion of the spins in \sectref{sect:energy-pooling}. 

In \figref{fig:high-loss-outputs}, we observed that the major variations between the reconstructed outputs are well represented by globally homogeneous fields for a one-dimensional bottleneck. 
For larger bottlenecks, this is complemented by smooth fields with increasingly high-frequency components (representing collective variables in the Ising configurations) in the later stages of training. 
This suggests that the theoretically motivated observables (physical concepts), including their coarse-grained averages across different spatial scales, are ideal candidates for a simplified, consistent, and physically meaningful explanation of the learning dynamics.

For better visualization, the following analysis is based on all runs with a medium learning rate of $\learningrate=10^{-4}$ and the most typical model configuration of depth $\depth=4$, width $\width=512$, and bottleneck size $\bottleneck=8$ where applicable. 
This analysis will focus on behaviors that generalize to other hyperparameter choices and, when necessary, highlight differences.
In general, behavior at larger scales varies little across runs, except in the deepest models (chaotic regime), while strong variations persist for small-scale observables across all runs and models. 
The interested reader can find the plots for other exemplary configurations in \sectref{sect:appendix-obs-losses-evo} and all other configurations in the supplementary data \cite{DARUS-6128_2026}.

\subsubsection{Coarse-Grained (Local) Averages of Microscopic Quantities}\label{sect:pooled-averages-of-microscopic-properties}

The mean sample magnetization and the mean sample energy can be expressed as global averages of the local properties at each lattice site.
The mean magnetization $\meanmagnetization$ is an average of all site spins $\sitespin{i}=\localmagnetization{i}$ (interpretable as local magnetizations), and the mean energy $\meanenergy$ is an average of site energies $\siteenergy{i}=\localenergy{i}$, depending on the spin difference of one site to all its neighboring sites (interpretable as local energies).
\begin{equation}
    \meanmagnetization = \frac{1}{\numspins}\sum_i^\numspins \underbrace{s_i}_{\localmagnetization{i}}
    \label{eq:global-mag}
\end{equation}
\begin{equation}
    \meanenergy = \frac{1}{\numspins}\sum_i^\numspins \underbrace{\sum_{j \in \siteneighborhood{i}} s_i s_j}_{\localenergy{i}}
    \label{eq:global-energy}
\end{equation}
This allows one to derive local mean magnetizations and energies on intermediate scales by averaging over differently sized square patches $\kernelneighborhood{i}$ centered around each spin site.
$$
\scalemagnetization_{i} = \frac{1}{\numkernelelements{i}}\sum_{j \in \kernelneighborhood{i}}^\numspins s_j
$$
Each kernel size $\kernelsize$ is particularly sensitive to collective modes at different scales.
The MSD of averages thus provides a scale-dependent probe for fluctuations in the error field, revealing spatial correlations.

Theoretically, the MSD between averages is guaranteed to be lower than or equal to the MSD of the individual elements due to the Cauchy-Schwarz inequality \cite{cauchyFormulesQuiResultent2009}. 
However, the loss of averaging at intermediate scales (i.e., over 3x3 patches) is not guaranteed to be bounded by averaging at larger scales (i.e., over 5x5 patches) as long as they can't be described as an average over the smaller patches (9x9 as an average over 3x3). 
This can be understood with the one-dimensional counterexample given in \sectref{sect:appendix-msd-scaling-example}.

The ordering of MSE loss magnitudes between $\kernelloss{a}, \kernelloss{b}$ depends on the fluctuations of errors across the averaged areas. 
If errors fluctuate across an area, these fluctuations are replaced with the area average, resulting in a reduced MSD (due to outlier sensitivity). 
With fewer fluctuations in the error field and hence more correlation, the coarse-grained macroscopic MSD converges to the microscopic MSD. 
The differences in losses across averaging scales can thus inform us about the spatial patterns of error fluctuations, characterizing the micro-macro connection.  
If error fluctuations bridge multiple scales, the scale-dependence is more likely to remain monotonic. 
More generally, the variance between spatial averages at different positions (sites) within a single sample decreases with larger kernel sizes. 
This is because the overlap in elements used for averaging increases until they are all identical at the global scale, and the variance vanishes. 
The scale dependence of learning the concepts of spin and energy is illuminated in the following section.

\subsubsection{Spins}\label{sect:spin-pooling}

Starting from the raw inputs $\rawinput$ and outputs $\rawoutput$, we can compute the physical spins $\sitespin{i}=2\rawpixel{i}-1$, mapping normalized values of 0,1 back to -1,1.
The averaging of spins over a growing square neighborhood around each site represents a coarse-graining operation that connects local (micro) values with global averages over the entire lattice (macro values). 
The local values are optimized explicitly during training, while the global averages represent the magnetization and thus an important order parameter of the Ising model. 
The loss for these average quantities is computed using the same MSE loss function as for the optimization algorithm, but with coarse-grained averages around each local site as inputs. 
$$
\magscaleloss=\magscalelosslong{a \times a}
$$

\begin{figure*}
    \centering
    \includegraphics[width=1\linewidth]{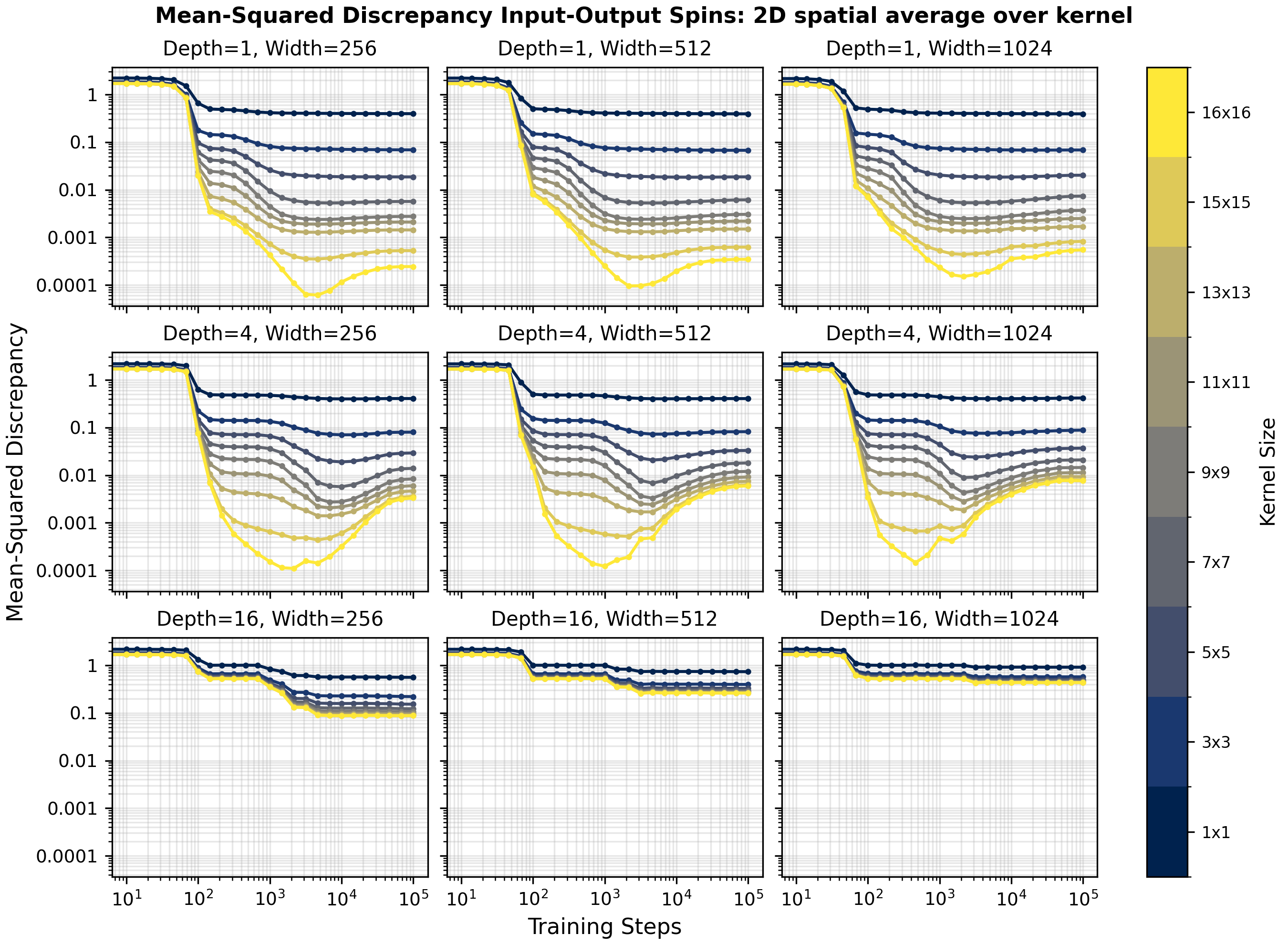}
    \caption{
    Validation loss evolution of coarse-grained averages of local lattice site spins using different kernel sizes, plotted across training checkpoints. 
    The averages include all \textbf{MB8DXWX runs} (see \tabref{tab:run-names}) with a learning rate of $\learningrate=10^{-4}$ and a bottleneck size of $\bottleneck=8$. 
    Each subplot shows averages across runs with a specific model width and depth, with line color indicating the kernel size used for averaging. 
    The validation loss decreases as the kernel size increases, despite the lack of a guarantee that the different averages have a particular order. 
    Large-scale losses have a temporary minimum for all shallower models. 
    Beyond this point, the average reconstruction on large scales becomes worse, while it continues to improve on small scales. 
    Deep models exhibit learning arrest following several transitory states (drops).
    }
    \label{fig:spin-pooling}
\end{figure*}

In \figref{fig:spin-pooling}, the average validation loss decreases with increasing kernel size for all runs.
Increasing the kernel $\kernelsize=b\times b \rightarrow a \times a$ adds the errors in the annular region between the two squares (the `square ring' of width $a-b$) to the averaged value.
A reduction in loss when enlarging the averaging area implies that the MSE over the b×b region and the MSE over the added square ring differ (so averaging reduces the square-error magnitude).
This behavior can occur if the reconstruction predicts a nearly constant (global mean) value as observed in \figref{fig:high-loss-outputs}, so that local deviations have opposite signs where spin orientations are anti-correlated (spin cluster boundaries). 
The losses, therefore, probe the model's ability to reconstruct clusters at the corresponding size scale. 
The observed separation of MSE losses across scales requires spatial error fluctuations between the corresponding length scales. 
The scale-dependent ordering of losses in \figref{fig:spin-pooling}, though not theoretically guaranteed (see \sectref{sect:appendix-msd-scaling-example}), suggests that error-field correlations are weak or lack a single dominant scale. 
An exception to this rule was found for runs with the largest bottleneck $\bottleneck=64$ and the slowest learning rate $\learningrate=10^{-5}$ shown in \figref{fig:spin-pooling-lr5-bn64}. 
Here, the $\kernelloss{13}$ kernels cross with the neighboring scales while all other scales remain ordered.
We conclude that, for almost all tested models, learning global properties precedes learning local properties. 
This underlines the importance of macroscopic observables in describing the initial learning dynamics. 

As the relative spacing between validation losses across scales changes during training, the distribution of error fluctuations at those scales must evolve. 
This evolving scale dependence provides insight into learning dynamics, connecting macroscopic and microscopic observables.
Notably, the loss of all large scales shows a global minimum for shallower models ($\depth=1,4$).
As the loss on these scales increases again, information about the global average must be lost, while local averages are better resolved.
This reveals a representational trade-off between local and global resolution, which may be caused by the model class and the training dynamics.
For wider bottlenecks, the trade-off seems to be weaker with decreasing width and vanishes completely, except for the deepest or widest model (see \figrefs{fig:spin-pooling-lr3-bn64}{fig:spin-pooling-lr5-bn64}).
This suggests that the representational trade-off is influenced by both the model size (width, depth, bottleneck) and the learning rate.
Why resolving smaller scales tends to increase the loss at larger scales remains an open question, which we attempt to answer with the bigger picture developed throughout this study.

In general, we observe that the loss across scales decreases consistently before reaching its minimum, albeit with stronger fluctuations at faster learning rates. 
In \figref{fig:high-loss-outputs}, the reconstruction matches the global mean during the initial phase, resulting in a spatially uniform output without the initialization noise. 
Therefore, the initial improvement in loss across scales appears to be mainly due to this mean matching. 
Only after the local minimum at the global scale is reached, smaller-scale features start to develop, seen in \figref{fig:high-loss-outputs}. 
The stagnation of optimization at large scales at the local minimum occurs when the loss is comparable to an optimal uniform prediction (i.e., the lower red reference line in \figref{fig:raw-msd}). 
Progressively smaller features develop in \figref{fig:high-loss-outputs} after this boundary is passed.
In contrast, the reconstructions remain uniform for deeper models that don't pass the loss threshold corresponding to an optimal uniform prediction (lower red reference line in \figref{fig:raw-msd}). 
For shallow models with small bottleneck sizes, we observe a trade-off between scales: large scales become slightly worse while smaller scales continue to improve. 
Thereafter, the loss contains local minima at smaller scales, which appear at progressively later stages of training with decreasing scale. 
All in all, the predominance of particular scales simplifies the description of learning dynamics. 
We built upon this picture in subsequent sections. 

The value of the loss minimum at the largest scale is similar to the respective base learning rate (see supplementary data \cite{DARUS-6128_2026}). 
With slower learning rates, we observed that the largest scales are initially reconstructed better than smaller scales before all scales converge to a similar loss (where the bottleneck is smaller: $\bottleneck=1, 8$) \cite{DARUS-6128_2026}. 
Therefore, a slower learning rate seems to favor learning coarse-grained observables via collective representation. 
This applies to small bottleneck sizes, for which we observe a trade-off between scales. 

\begin{figure*}
    \centering
    \includegraphics[width=1\linewidth]{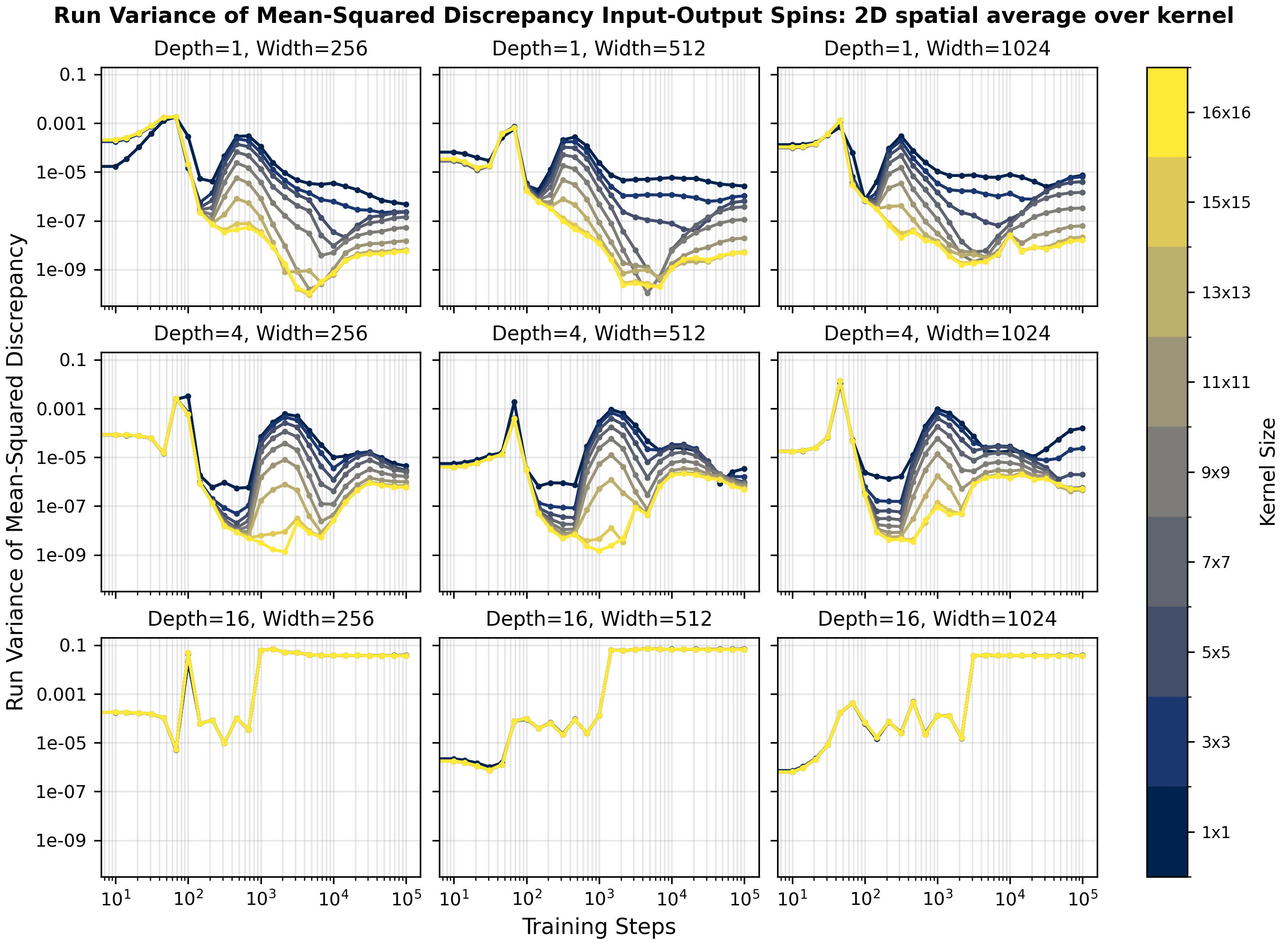}
    \caption{
    Run variance evolution of the raw spin loss, calculated over different trajectories (runs), as a function of the spatial-coarse-graining kernel size, corresponding to \figref{fig:spin-pooling}. 
    The variance is computed over all \textbf{MB8DXWX runs} (see \tabref{tab:run-names}) with $\learningrate=10^{-4}$ and $\bottleneck=8$. 
    This path variance shows a consistent hierarchical split according to the coarse-graining (averaging) scale. 
    The deep-learning regime purports no such splitting. 
    In the upper panels, the first `dips' in the evolution after the initial increase coincide with transitory states that exhibit scaling (plateaus) in the MSD in \figref{fig:spin-pooling}, characterizing a magnetization-dominant learning regime. 
    The peaks coincide with transitions in the MSD, suggesting the sensitivity to initial conditions is most pronounced during transition periods. 
    }
    \label{fig:variance-run-spins}
\end{figure*}

By measuring the variance over different trajectories (i.e., between all 12 runs with the same depth, width, bottleneck, and learning rate but variable initialization, training data order, and batch size), we explore how sensitive autoencoder learning is to initial conditions and the driving dynamics (\figref{fig:variance-run-spins}). 
As in \figref{fig:spin-pooling}, we evaluate the raw spin loss as a function of spatial coarse-graining size (kernel size). 
Typically, the variance between runs is high on scales having large loss. 
Additionally, the variance between runs differs across scales when the losses of the respective scales also differ. 
The run variance tends to reduce when the losses plateau. 
It rises again during loss transitions (i.e., at step 100 and between steps 1 000 - 10 000). 
This suggests that runs with different initialization and driving protocols converge to similar plateaus and differ mainly in their timing of these transitions.

\paragraph*{Deep models:} Interestingly, deep models (lower panel of \figref{fig:variance-run-spins}) exhibit different behavior in terms of run variance across spatial scales. 
Here, the variance remains almost identical across different scales, as in the initial phases of shallow models. 
Generally, the late-stage behavior depends on the learning rate and hyperparameters (contained in the supplementary materials \cite{DARUS-6128_2026}). 
For fast learning rates (not shown here), all variances remain similar across scales and vanish over time, suggesting that each run makes the same predictions (the input-independent dataset average). 
For medium learning rates (see \figref{fig:variance-run-spins}), the variance remains high, with exceptions for the largest-bottleneck and largest-width cases. 
This suggests more chaotic training dynamics. 
For slow learning rates, the variance differs slightly across scales at later stages (similar to shallow models). 
This reveals a strong dependence of the learning dynamics on the learning rate. 
Approximations of the lattice spins are worse with faster learning rates (see \figref{fig:raw-msd}). 
Therefore, deep models might remain in, or sink deeper into, a local minimum or a high-dimensional saddle point. Shallower models appear to escape from these (alternatively, training with slow learning rates helps). 

These statements also appear true for learning the energy concept, as described in \sectref{sect:energy-pooling}. 
There, learning the energy concept is not directly correlated with the optimization loss, at least above the lower reference line in \figref{fig:raw-msd} for the sample-average prediction. 
When the optimization loss is above this line, the global-scale errors dominate. 
Energy is learned across scales only after crossing the local minimum for large scales (the aforementioned trade-off between scales). 
Small-scale losses must then improve as the large scales are almost perfectly predicted. 
This necessitates the autoencoder to resolve spin-domain boundaries at progressively finer scales.

In summary, we note that averages of absolute spin values (i.e., the magnetization) provide a scale-consistent explanation for the initial learning dynamics. 
The averages of relative differences (i.e., the energies) offer a similarly effective explanation for later stages when trade-offs between largest and smaller scales occur (not for the arrested deep models). 
Potentially, the idea of scale-consistency can be extended to explain higher-order correlations. 
An alternative effective explanation is that training dynamics are first captured by larger scales. 
Over time, increasingly smaller scales play a stronger role, which benefits the energy loss and other refined concepts that correlate multiple spins. 
The slight shift of the local trade-off minima with shrinking scales towards later stages of training also supports this explanation of large-to-small-scale training (observable in \figref{fig:spin-pooling}).

\subsubsection{Joint Output Distribution Dynamics}\label{sect:output-distribution-dynamics}

To better explore the relationship between spin and energy averages, we observe their joint distribution evolving. We treat the macroscopic level only (global averages). 
Ideally, we would observe convergence towards the dataset distribution depicted in \figref{fig:dataset-contours}. 

\begin{figure*}
    \centering
    \includegraphics[width=1\linewidth]{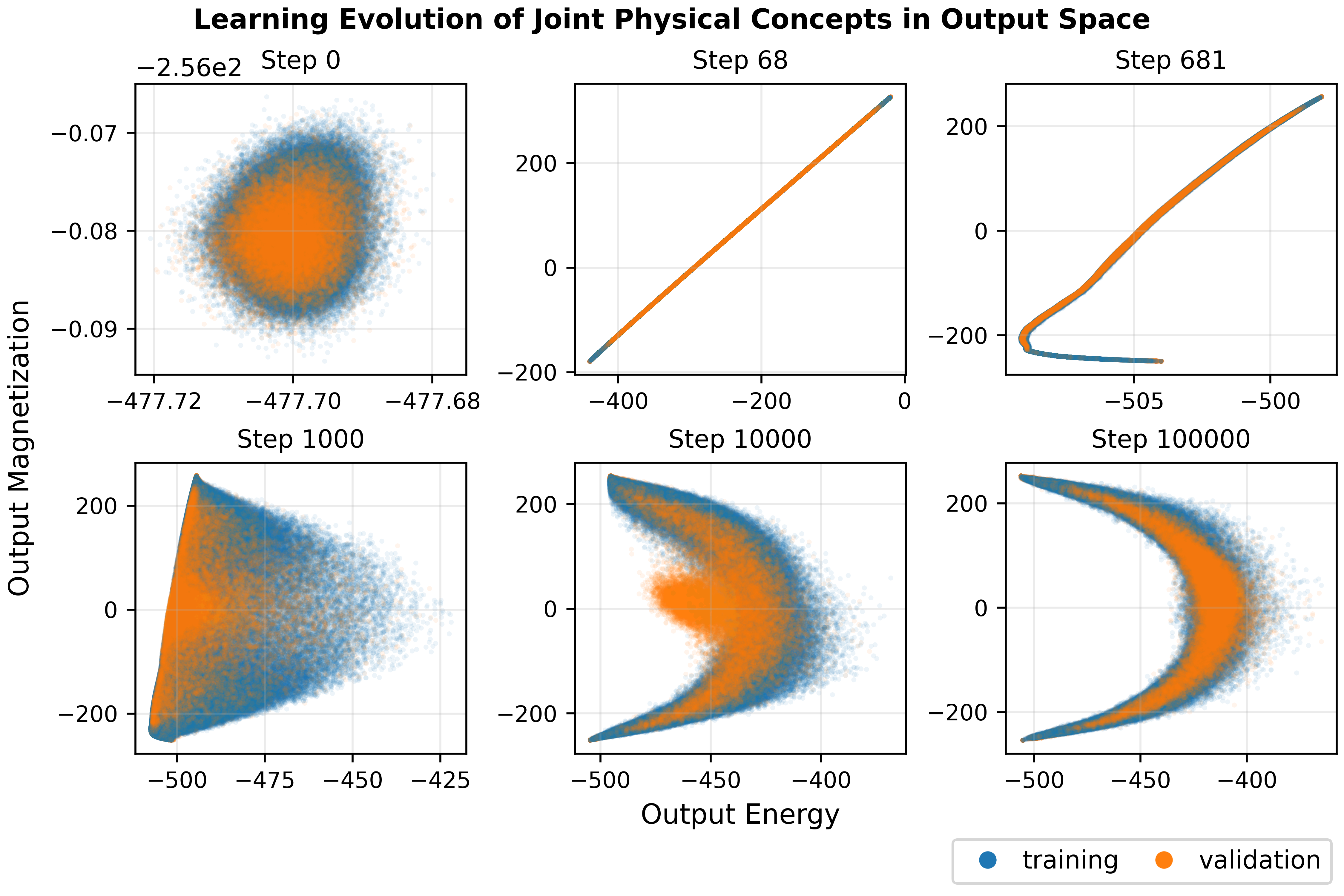}
    \caption{
    Joint distribution of magnetization and energy in the output space at six different checkpoints of \textbf{run MB8} (see \tabref{tab:run-names}) during training (validation: orange, training: blue). 
    The distribution initially resembles a Gaussian and then deforms into a one-dimensional line. 
    This line becomes a two-dimensional curve before expanding into a two-dimensional surface. 
    Finally, the distribution becomes similar in shape to the true input distribution for the training data. 
    At the same time, it remains slightly deformed for the validation data compared to the input distribution in \figref{fig:dataset-contours} (missing blob). 
    The output energies of both the training and validation data are shifted to much lower energies ($\energy\in[-512, -350]$) compared to the input data ($\energy\in[-512, 0]$) visible in \figref{fig:dataset-contours}. 
    The shift towards lower energies is less pronounced when the bottleneck size is increased (see \figref{fig:mag-vs-energy-64}).
    }
    \label{fig:mag-vs-energy}
\end{figure*}

\figref{fig:mag-vs-energy} displays the evolving joint output distribution of magnetization and energy calculated over training data. 
Qualitatively, this converges towards the joint input distribution while retaining a systematic shift towards lower energies, as observed in \figref{fig:io-energy}. 
In the initial stage of training, when the state distribution is point-like (Step 0), output magnetization and energy are uncorrelated. 
However, during the initial decrease in raw data loss (see \figref{fig:raw-msd}), a line-shaped joint distribution emerges between the two concepts (Step 68); hence, these become linearly correlated to first order.
A non-linear curve starts to develop (Step 681) during which the raw data loss remains constant. 
Different magnetizations can now correspond to the same energy. 
At this stage, the output energies are still uniquely determined by the output magnetization: The sample outputs represent only a single globally uniform mode (see \figref{fig:high-loss-outputs}), and no spin-domain boundaries are visible. 
The lack of spatial fluctuations in the output (present at step 68 in the form of initialization noise) means that the input energies are approximately invariant with respect to magnetization. 
Learning magnetization across scales at this early stage largely suffices to describe the initial learning dynamics. 

As the training progresses, collective modes in the configurations get resolved at smaller length scales: Boundaries begin to appear in the output space, i.e., spin domains (see \figref{fig:high-loss-outputs}). 
Such boundaries resolve energy differences without altering the global average magnetization. 
This breaking of energy-invariance is marked by a rapid expansion of the joint distribution across the energy dimension (Step 1000). 
However, all energies remain systematically underestimated, clearly visible when comparing the range of energy values in the output ($\energy=[-512, -350]$) and in the input ($\energy=[-512, 0]$) visible in \figref{fig:dataset-contours}. 
The highest energies are most underestimated. 
These are only present in the validation data. 

The predominance of the bias at high energies can be elucidated by the evolution of coarse-grained local observables, introduced in \sectref{sect:learning-of-observables} (see also \figref{fig:io-energy}): 
The blob to the left of the center at step 10000 in \figref{fig:mag-vs-energy} most likely corresponds to the highest energies (see \figref{fig:dataset-contours}).
Small-scale features are required to represent high energies; however, these are only resolved much later, provided sufficient representation capacity is available. 
The overlap of the crescent-shaped distribution with the ground truth in \figref{fig:dataset-contours} necessitates small-scale features to be resolved. 
Especially samples outside the training distribution, with low magnetization, experience a shift towards higher energies (the orange blob disappearing at Step 100000 in \figref{fig:mag-vs-energy}). 
We conjecture that this shift happens when learning of the energies becomes correlated with the optimization loss. 

The aforementioned trade-off (local minima in \figref{fig:spin-pooling}) occurs at around step 10000 for this particular run.
At this stage, the energy losses (in \figref{fig:energy-pooling}) flatten off and become consistently reduced across scales, reflecting the concentration of outputs towards higher energies that we observed in the distributions (in the vanishing orange blob). 
The consistently monotonic loss evolution across scales appears to coincide with their generalization beyond the training data, at least for the physically motivated observables studied here. 

For other model configurations, we made the following observations (contained in the supplementary material \cite{DARUS-6128_2026}): 
A small bottleneck $\bottleneck=1$ results in a one-dimensional curve that becomes increasingly complex with depth. 
For bottleneck $\bottleneck=8,64$, the behavior was similar to that in \figref{fig:mag-vs-energy}, while depth $\depth=4$ models remained longer in the approximately linear regime. 
The distribution for $\depth=16$ models starts from a single point and either collapses to a single point, shows 2 or 3 distinct vertical lines, or forms a complex 1D curve without a clear hyperparameter dependence. 

\begin{figure*}[ht]
    \centering
    \includegraphics[width=1\linewidth]{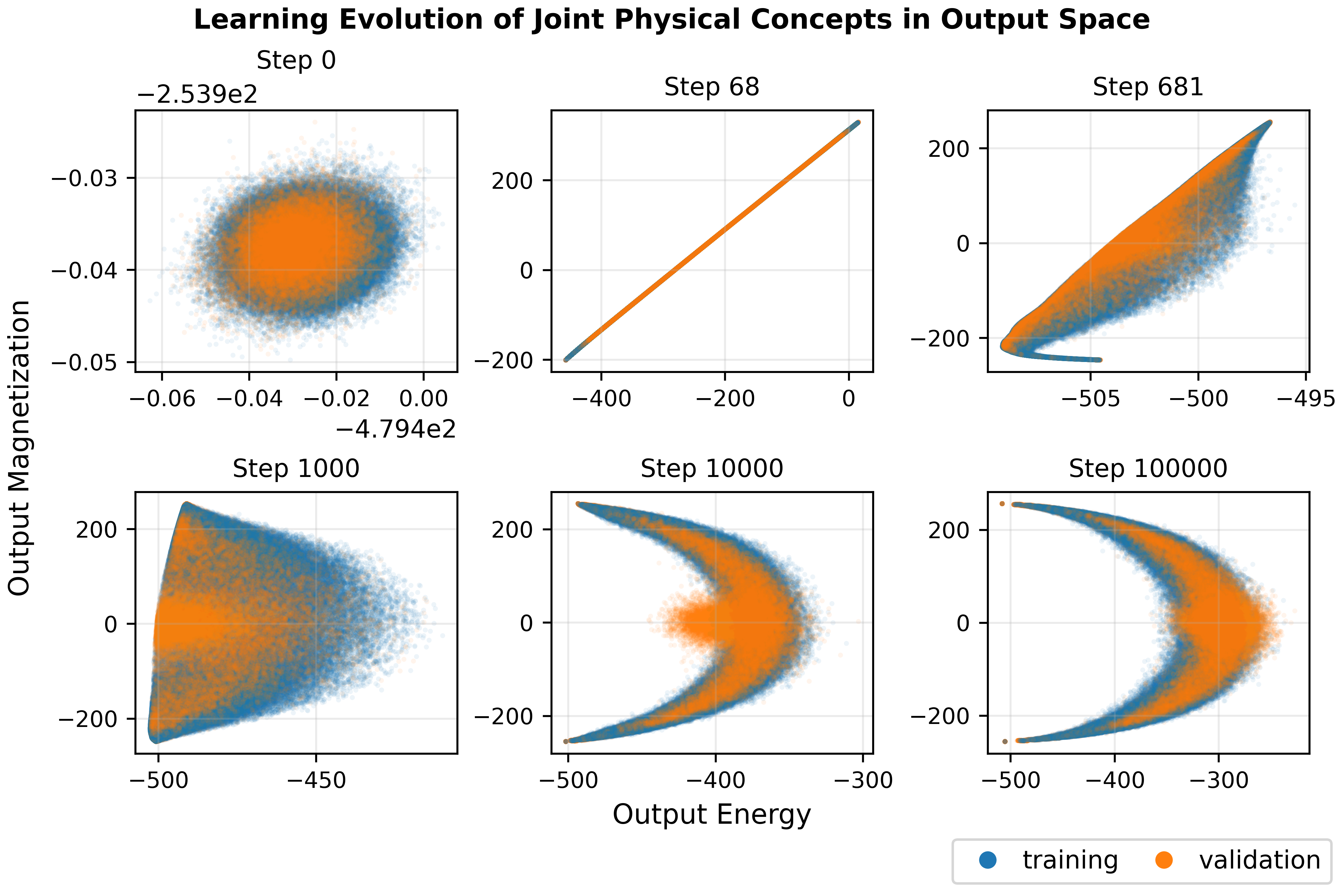}
    \caption{
    Joint distribution of magnetization and energy in the output space at six different checkpoints of \textbf{run MB64} (see \tabref{tab:run-names}) (validation: orange, training: blue). 
    The distribution initially resembles a Gaussian. 
    At step 68, it has deformed into a one-dimensional line. 
    From step 681 on, the distribution extends in two dimensions.  
    At step 10 000, the distribution becomes similar in shape to the expected distribution in the input space for the training data (see \figref{fig:dataset-contours}). 
    For the validation data, the high-energy region with vanishing magnetization is particularly misrepresented. 
    In general, the entire distribution is shifted towards lower energies compared to the input distribution. 
    However, the shift is smaller compared to runs using models with a smaller bottleneck, as shown in \figref{fig:mag-vs-energy}. 
    This shows that increasing the bottleneck size plays a major role in pushing the generalization ability towards higher energies. 
    }
    \label{fig:mag-vs-energy-64}
\end{figure*}

We conclude that the uniform mode of the input (i.e., the magnetization) explains the initial learning dynamics. 
Late-stage dynamics are characterized by learning progressively smaller scales, given sufficient representation capacity. 
However, the smallest scales never reach low loss values compared to large scales, resulting in systematic underestimation that is especially pronounced for high-energy inputs. 
The smaller energy error in \figref{fig:mag-vs-energy-64}, where only the bottleneck size $\bottleneck=64$ was increased, supports the hypothesis of a bottleneck-induced resolution limit. 
The bottleneck limits the reconstruction and representation of high energies most significantly. 

In the following section, we address how the phenomena on the output distribution are related to the dynamics of the internal representations. 

\subsection{Latent Representation Dynamics (Twofold)}\label{sect:representation-dynamics}

Understanding the dynamics of learning cannot be reduced to input-output behavior measured solely by sample losses or distributional differences, as discussed in \sectref{sect:learning-of-observables}. 
An analysis of the evolving sensitivity of internal representations to reconstruction errors provides an additional systemic characterization of the dynamical system. 
In the following sections, we will focus on the representation dynamics of the most constrained part of the model: the bottleneck. 
This will help us understand how the compressed representation of the input space is generated.

\subsubsection{Recursive (Self-Application) Dynamics in Latent Space}\label{sect:recursive-flow}

We explore a novel setting for understanding autoencoder errors as a dynamical system: The recursive application of the autoencoder to its own output generates a dynamic. 
Specifically, the output is used as the input of the next recursion (time) step. 
We call this `recursion dynamics.' 
As it is not dependent on any external degrees of freedom -- such as the evaluation and feedback of the loss function -- self-recursion defines an intrinsic dynamic. 
However, the first step depends on the prediction error used during training, i.e, the difference between the input and predicted output representation. 
If we had no prediction error, there would be no dynamics. 
The dynamics inform us about how the model transforms concepts between input and output. 
Importantly, these dynamics in the input/output space can also be mapped to the output representation of any layer in between. 
Therefore, the first layers up to the corresponding latent representation are applied to obtain the latent dynamics (e.g., using only the encoder to obtain the dynamics inside the bottleneck). 

We aim to compare representational changes during training with the evolving recursion dynamics. 
These are explored at fixed checkpoints. 
Sampling the recursion dynamics for different inputs reveals an underlying flow field (generating the trajectory motion), which changes during training. 
Both the information representation (topology) and its transformation (recursive dynamics) can therefore be probed over training time. 
In the following, we sample this flow field on validation data to reveal the topological evolutions associated with concept formation.
For example, fixed points and attractors relate to coarse-graining and compression occurring inside the model. 

\begin{figure*}
    \centering
    \includegraphics[width=1\linewidth]{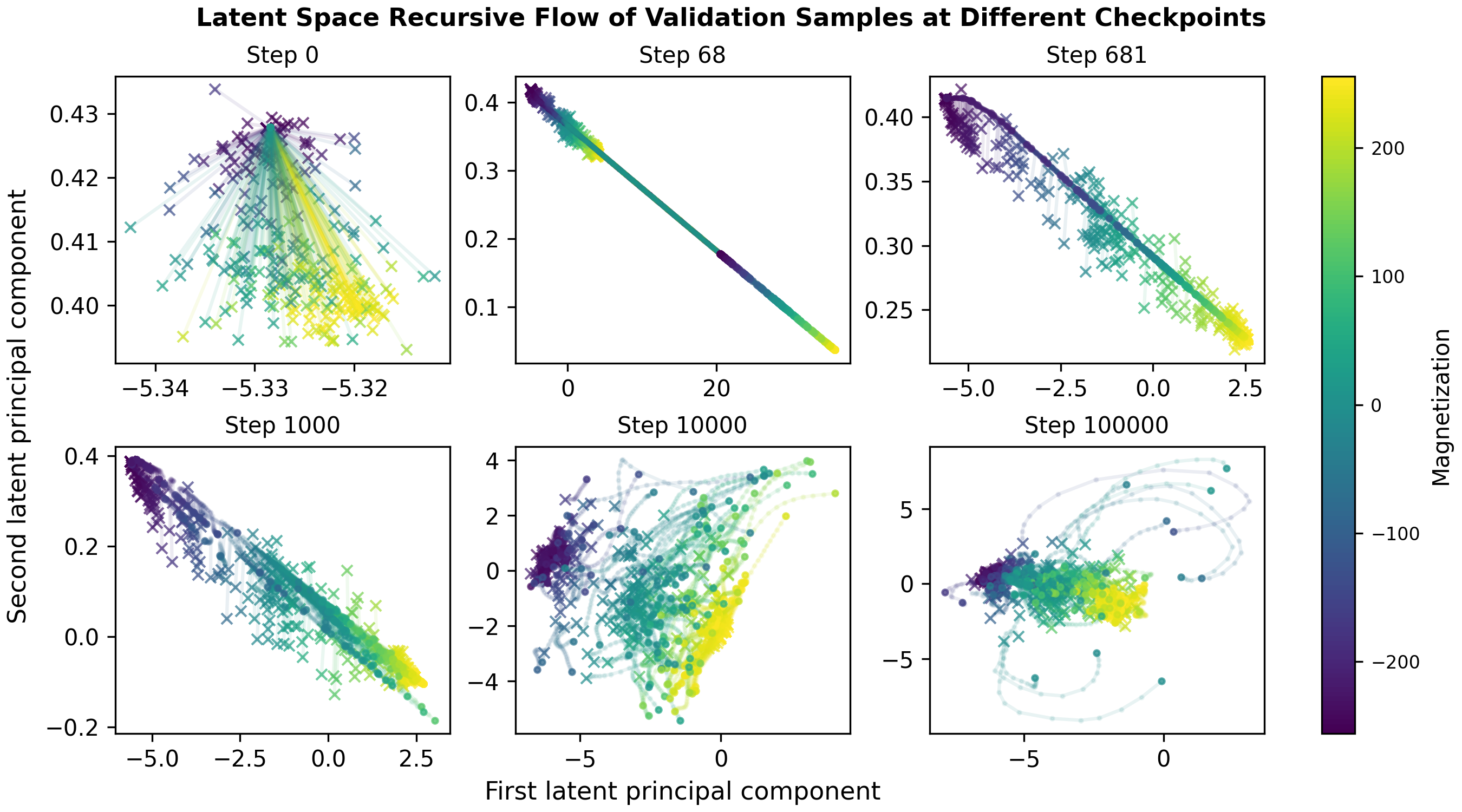} 
    \caption{
    The first two principal components of the trajectory of the 8-dimensional latent representation of validation samples. 
    The trajectory is obtained by recursively applying the autoencoder model to initial samples, at different training checkpoints of \textbf{run MB8} (see \tabref{tab:run-names}). 
    The principal components are computed once across all checkpoints. 
    The latent representation of the initial input (cross) moves with each recursion step to the next position in latent space (small circles), until the last step (big circle). 
    The trajectories are colored according to each sample's magnetization. 
    Initially, samples move towards a single point, then a line, 2d curve, and finally they move through all visible dimensions. 
    After step 68, points with more similar magnetization seem to remain closer together, indicating the emergence of a stable concept representation. 
    }
    \label{fig:latent-flow-magnetization-b8}
\end{figure*}

\figref{fig:latent-flow-magnetization-b8} presents trajectories for multiple validation samples obtained via 20 recursion steps.
To visualize, we connect the latent representations of the initial input samples (via crosses) to those obtained after multiple recursion steps (via small circles) towards the final representation (via big circles). 
This provides an idea of the flow field's structure that changes during training. 
Fixed points in the recursion dynamics reveal samples with perfect reconstruction by the autoencoder. 

Theoretically, all recursion dynamics in a finite space must relax/converge towards a limit cycle, where they oscillate forever, or become stationary at one of possibly multiple limiting fixed points. 
At each moment in training, every input is thus either on a relaxation trajectory towards such a limit, part of a limit cycle, or itself the corresponding fixed point. 
The latent space inherits these dynamical regimes from the input space: 
Under a deterministic model, different inputs may result in the same latent representation, but the same latent representation cannot result in different outputs. 
This is because any difference in the output is inherited from the preceding layers. 
Therefore, when scanning across representation layers, fixed points, cycles, and relaxing trajectories remain what they are under recursion. 
In the following, we will refer to these without specifying the `input' or `latent' space, as layers are equivalent in this regard. 
We analyze the lowest-dimensional latent space, i.e., dynamics at the bottleneck, because the relevant dynamics are compressed into fewer dimensions. 

\begin{figure*}[ht]
    \centering
    \includegraphics[width=1\linewidth]{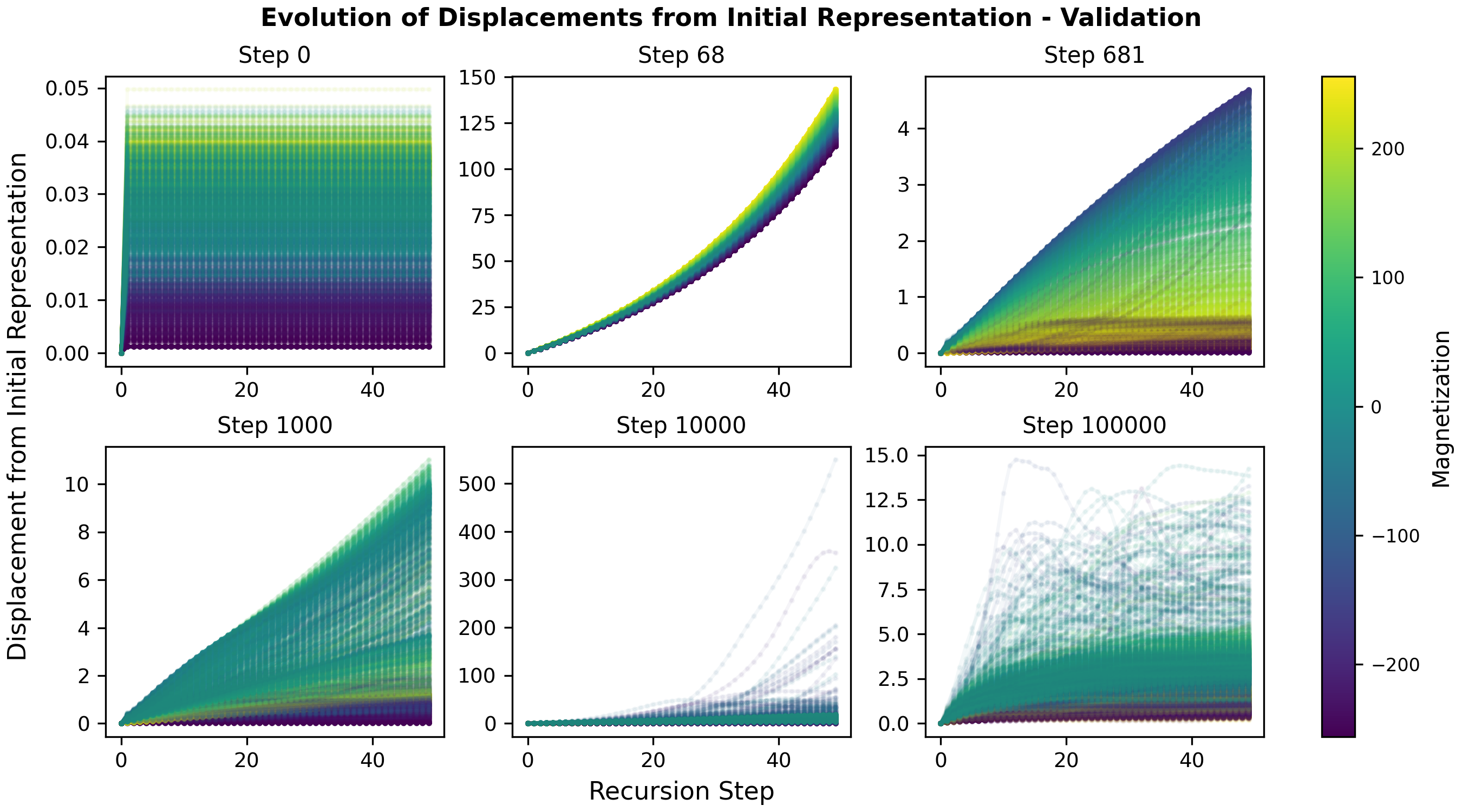}
    \caption{
    Evolution of mean-square displacement (MSD) between the starting point and points along the recursion trajectories inside the 8-dimensional latent representation for \textbf{run MB8} (see \tabref{tab:run-names}) shown in \figref{fig:latent-flow-magnetization-b8}. 
    Each curve is colored according to the magnetization of the corresponding sample. 
    At the start of training, the MSD remains constant after the first step, suggesting that the recursion immediately converges to a fixed point. 
    At step 68, the MSD increases with recursion for every magnetization and the distribution broadens. 
    This demonstrates that the representational shift in \figref{fig:latent-flow-magnetization-b8} also increases the sensitivity to magnetization.
    For later checkpoints, the MSD mainly increases for samples with vanishing magnetization, corresponding to less probable, high-energy samples (see \figref{fig:dataset-contours}). 
    In contrast, samples with high absolute magnetization (low energy and high probability) remain close to the initial values, suggesting that a stable attractor is formed with respect to the recursion dynamics. 
    }
    \label{fig:flow-msd-dist}
\end{figure*}

The initial recursion dynamics, at step 0 in \figref{fig:latent-flow-magnetization-b8}, show an immediate convergence of all inputs towards a single fixed point (within the two principal components). 
The analysis of the MSD between the initial and later points in \figref{fig:flow-msd-dist} suggests that this fixed point exists in all dimensions. 
The immediate convergence signifies the loss of all information that differentiates the encoders' representations of the original input (crosses). 
Theoretically, trajectories converging to the same fixed point indicate a loss of information, suggesting coarse-graining. 
In contrast, the convergence to distinct fixed points signifies the stability or preservation of this information. 
Therefore, the emergence of new fixed points is expected during learning. 

In step 68, the representation drifts along one dimension, concurrent with the first drop in the loss for the raw data (orange line in \figref{fig:raw-msd}) and the coarse-grained spin averages (\figref{fig:spin-pooling}).
The recursive dynamics move points far away from their initial representations. 
However, the points remain ordered with respect to the input magnetization. 
While this topological ordering is maintained, there is a global drift toward higher magnetization. 
Thus, the value of magnetization is not stable; it is overestimated in the output. 
However, one could say that magnetization is a stable `topological' concept at step 68, as the corresponding neighborhood structure is preserved across recursion (from the crosses to big circles). 

In step 681, the points converge onto a one-dimensional line upon one recursion:  
The magnetization value becomes more stable. 
A late drift appears towards points with higher magnetization along this line. 
For the samples with the highest magnetization (yellow points), movement is minimal, indicating the emergence of a stable fixed point. 
After step 1000, the recursion expands in the second principal component dimension, just after the trade-off becomes visible in \figref{fig:spin-pooling}, i.e., mainly smaller scales improve. 
After large-scale losses increase, the recursion weakens the magnetization's previous topological ordering in latent space. 
The flow field becomes less confining, such that the full latent space is traversed under recursion (in these two dimensions). 
A bit of representation drift remains. 
The loss of confinement to a line coincides with an expansion of the joint concept distribution between magnetization and energy in output space in \figref{fig:mag-vs-energy}. 
This highlights the connections between different representation spaces. 

In step 10 000, the drift disappears, and the trajectories expand along the second dimension. 
This happens as additional symmetries are broken, which resolve increasingly smaller spatial domains in the output representation (see \figref{fig:high-loss-outputs}). 
At the same time, outputs are shifted to higher energies, requiring these smaller-scale fluctuations (see \figref{fig:io-energy}). 
In step 100 000, loops begin to emerge in the trajectories. 
We assume that the underlying flow field becomes confining, thereby limiting the spread of latent representations (in the visible dimensions). 
The MSD between initial points and later recursion points (\figref{fig:flow-msd-dist}) first increases but appears to slow down or decrease afterwards. 
Samples with larger absolute magnetization appear to be more confined. 

Generally, the decoder preserves an increasing amount of input information over learning, i.e., more latent representations can be differentiated. 
As an increasing number of dimensions is retained (avoiding collapse), representations become more consistent and stable under recursions. 
This implies transient stability of the concepts represented. 
Additionally, we expect a slowing of recursion dynamics viewed in the output space as the loss vanishes. 
The slowdown in output space could lead to smaller relative steps in the latent space (only relative, due to the scale invariance of ReLU networks). 

The principal components of the latent space provide a partial visualization of the representation dynamics. 
However, concepts can form in all dimensions, which necessitates a quantitative analysis of the topology in the following section.

\subsubsection{Stability of Ising Concept Representations under Recursive Flow}\label{sect:spearman-correlation}

From \figref{fig:latent-flow-magnetization-b8}, we anticipate that, between steps 68-1000, points that are closer together in latent space will correspond to samples with more similar magnetization. 
This relationship should weaken after step 1000, as the energies are learned across scales and the trajectories in \figref{fig:latent-flow-magnetization-b8} expand into the second displayed dimension. 
A Spearman rank correlation can quantify the observed `topological' concept structure across all dimensions. 
It measures the monotonicity of the relationship between the representation distance in latent space and the magnetization difference of the input samples. 
Whenever the topology in latent space exhibits an ordering by a concept, the correlation is high; this means that samples with the same value of the observable are closer than those with different values. 
We refer to the existence of such an order as `topological ordering'. 
We can also measure the topological ordering of other concepts, such as energy. 
Thus, the Spearman rank correlation quantitatively measures how representations are structured according to a concept. 

Evaluating the Spearman correlation across recursive steps probes the topological changes to the representation (at a fixed checkpoint). 
One can disentangle the effects of the encoder and the decoder on the representation topology: 
At recursion index 0, the encoder produces the representation, while the decoder has no influence. 
With additional recursion, the decoder increasingly influences the latent representation structure (along with the encoder) and, therefore, the `topological ordering' with respect to a concept (i.e., magnetization). 
Examining how the Spearman correlation varies across training checkpoints enables us to understand how training modifies the recursion dynamics and topology of latent space. 

\begin{figure*}
    \centering
    \includegraphics[width=0.75\linewidth]{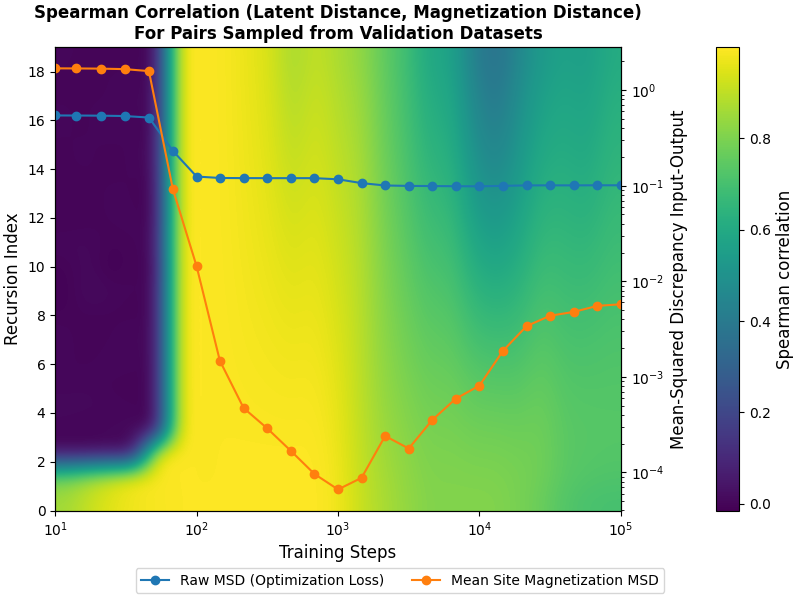}
    \caption{
    Spearman rank correlation between the input magnetization difference and latent space representation distance (L2-norm) of sampled validation data pairs. 
    The correlation is shown for different training checkpoints of \textbf{run MB8} (see \tabref{tab:run-names}) and using the latent representation for different numbers of recursions (feeding the output back into the input). 
    At the beginning of training, the correlation drops rapidly under recursion (lower left) as the encoder maps all points to the same output. 
    Around step 100, the correlation remains stable under recursion, concurrent with the drop in the optimized loss (blue) and the corresponding macroscopic magnetization loss (orange). 
    Around step 1000, the correlation decreases independently of the number of recursions, concurrent with another drop in the optimized loss and an increase in the magnetization loss. 
    It then remains low but gradually decreases with recursion until the end of training, with an exception between around $10^4-6\cdot10^4$. 
    The training-time periods with the higher correlation stable over recursion (around $10^2-10^3$) correspond to a transitory dynamical regime where this correlation is maximal and essentially stable under the closed-loop dynamics; it is visible as pronounced drops in run variance, see \figref{fig:variance-run-spins}. 
    }
    \label{fig:spearman-latent-vs-magnetization-b8}
\end{figure*}

\figref{fig:spearman-latent-vs-magnetization-b8} visualizes the Spearman correlation between representation pairs and their magnetization. 
Initially, the Spearman rank correlation is unstable under recursion, visible in an abrupt drop in the bottom-left corner of \figref{fig:spearman-latent-vs-magnetization-b8}, before becoming more stable under the recursive dynamics. 
This happens concurrently with a decrease in the optimized loss and the related magnetization loss. 
It remains more stable during the training period when the optimized loss plateaus, indicating a transition state with stable topological ordering (at least with respect to the intrinsic dynamics of recursion). 
As suggested by the flow trajectories confined to the one-dimensional region in Step 68 of \figref{fig:latent-flow-magnetization-b8}, the magnetization is only learned (see the orange curve dropping in \figref{fig:spearman-latent-vs-magnetization-b8}) after its topological ordering fully develops in latent space, signified by a high Spearman correlation. 
Additionally, successful learning of the magnetization concept comes after stabilizing its latent space topology (in the intrinsic sense, under recursion). 
At the same time, the trade-off between averages of spins on different scales becomes visible in \figref{fig:spin-pooling}; the correlation suddenly decreases just as the magnetization loss starts to increase. 

After Step 68, the optimization loss decreases, as reconstruction across scales continues to improve. 
At step 1000, the magnetization loss becomes minimal just before energies are learned across all scales (see \figref{fig:energy-pooling}). 
Thereafter, the topological ordering according to magnetization destabilizes, all the while additional degrees of freedom are required to resolve small-scale domains (see \figref{fig:high-loss-outputs}). 
At step 10 000, the latent space flow shows a general spreading and final drift into the upper right corner in \figref{fig:latent-flow-magnetization-b8}, and the Spearman correlation shows a transitory recursion instability in \figref{fig:spearman-latent-vs-magnetization-b8}.

The Spearman correlation with respect to the energy (Appendix \figref{fig:spearman-latent-vs-energy-b8}) is significantly lower compared to the magnetization, in general. 
It is only high in the regime where the latent space is almost perfectly ordered with respect to magnetization. 
Therefore, magnetization now also serves as an initial proxy for the energy in latent space, extending our finding in \sectref{sect:output-distribution-dynamics}. 
When the approximate invariance of energy is broken (\figref{fig:mag-vs-energy}), magnetization is no longer a good proxy, and the Spearman correlation of energy drops rapidly, even turning weakly negative. 
This suggests that energy is not represented by a distinct latent space dimension, but rather by a collective characteristic of averages at different scales. 

In conclusion, the topological ordering of latent space according to a concept and its stability under recursion coincide with learning the concept. 
The early appearance of a stable concept representation suggests that stability is a prerequisite for consistently improving their reconstruction.

\section{Discussion and Conclusion}\label{sect:discussion-conclusion}

We studied how unsupervised autoencoders trained on microscopic spin configurations from the Ising model learn macroscopic, theory-relevant variables like magnetization and energy. 
These observables not only describe aspects of the data-generating process but also provide effective descriptions of the learning dynamics. 
Learning is a dynamical system, which couples different subsystems: 
\begin{enumerate}
    \item the data-generating and -feeding processes, which determine the fluctuations in the structural features of the data and include the data shuffling and sampling;
    \item the model state, which defines the output's dependence on inputs and parameters; 
    \item the optimizer, which causes memory-dependent susceptibility depending on accumulated parameter gradients. 
\end{enumerate}
The fluctuations and correlations in the input data propagate through the model, driving its evolution mediated by the optimizer. 
Our analyses link the formation of physical representations to the model's learning dynamics, such as transition states and arrest across physical scales.

To mimic a typical discovery scenario, we embedded no prior knowledge into the model parametrization or the optimization algorithm, such as the symmetries of the Ising model. 
This approach enables discovery in settings where knowledge about the data-generating process is incomplete. 
Here, knowledge about the data was used to analyze the learning of physically meaningful observables (concepts) and identify connections to the underlying physics.

To attribute effects to specific hyperparameters, we conducted an ablation study. 
We identified different dynamical regimes in the optimization loss (as seen in plateaus and transitions in \figref{fig:raw-msd}) that depend on both model architecture (depth, width, bottleneck size) and optimization hyperparameters, i.e., learning rate. 
There, we observed that deeper models were less effective at learning good representations (low validation loss) and often failed to resolve input differences when trained at higher learning rates. 
Additionally, we found that larger widths and bottleneck sizes can adversely affect performance in this deep model regime. 
This behavior could result from increased mixing of input signals with depth and width, making their differentiation and, thus, reconstruction more difficult. 
The learning dynamics of shallower models are similar in shape but are stretched across time as base learning rates decrease. 
In the future, one could rescale the training time based on the momentary learning rate, which corresponds to the time step used to integrate a dynamical system.

\begin{figure*}[ht]
    \centering
    \includegraphics[width=0.75\linewidth]{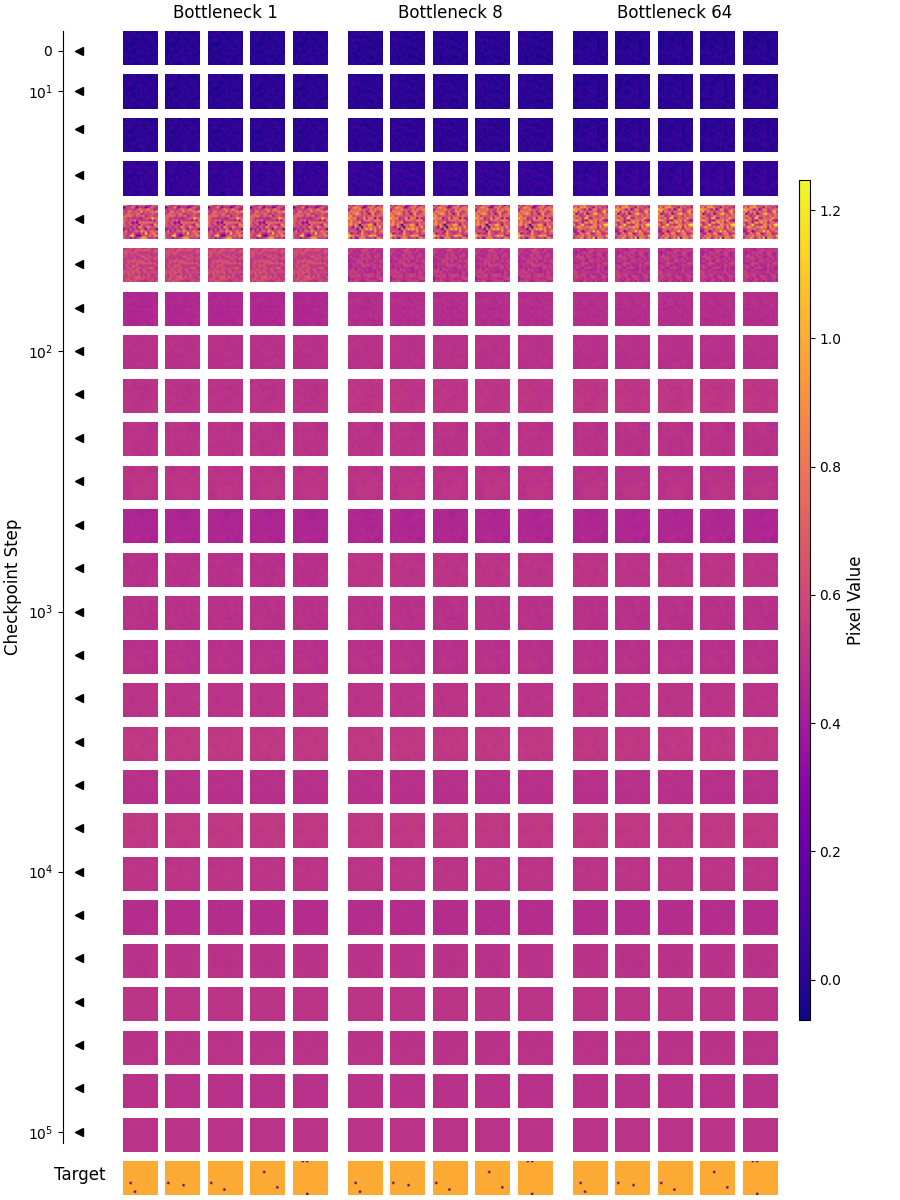}
    \caption{
    Model output evolution for samples from the critical temperature with the lowest final loss for different bottlenecks. 
    Each column of 5 samples shows the temporal evolution of one model, from bottom to top. 
    The last row shows the input, which is the target of the optimization. 
    The model checkpoints are taken from runs \textbf{FB1D16, FB8D16, FB64D16} (see \tabref{tab:run-names}), all training the deepest model $\depth=16$ with the most typical width $\width=512$ using the fastest learning rate $\learningrate=10^{-3}$. 
    The output for all samples starts around zero, where it remains until a rapid transition, with fluctuations between pixels, converges to uniform values slightly above $0.5$. 
    The bias towards high magnetization manifests in the dominance of samples with high magnetization for the lowest loss. 
    The bottleneck has no influence as the models are not able to resolve any input dependence and instead only learn a single global mean for the entire dataset.
    }
    \label{fig:low-loss-outputs-d16}
\end{figure*}

We analyzed the evolution of reconstruction loss for individual samples with the largest loss to visualize generalization errors; we observed overfitting for small bottleneck sizes (lower-left traces in \figref{fig:high-loss-outputs}) and good generalization for larger bottlenecks (via coarse-and-blurred structures). 
Individual samples suggest that deep models get trapped in an arrested regime: 
They reconstruct the same output across different samples, representing a dataset-average independent of the given input (\figref{fig:low-loss-outputs-d16}). 
We argue that this arrest is caused by the ruggedness of loss landscapes and their adverse effect on relaxation dynamics.  
This enriches the existing theoretical literature, which merely marks this regime as `chaotic', without specifying the details that could induce such behavior. 

In successfully trained models, learning begins by reconstructing globally uniform values matching the pixel mode or mean. 
Thereafter, progressively smaller-scale features are reconstructed. 
We conclude that averages on larger scales are learned before those on smaller scales, and that these larger-scale averages are partially unlearned at later stages. 
The fully connected models we studied have no prior for locality. 
Therefore, the reconstruction of local averages must result from the local correlations present in the training data. 
This suggests a more general relationship between the correlations in the data and the coarse-grained reconstructions.

Very deep and wide models rarely resolve input differences, and when they do, they are mostly restricted to the global average. 
This suggests that the models become less sensitive to small-scale features as depth and width increase. 
Possibly, each layer results in increasingly similar weighting of inputs until outputs converge to the same average. 
Therefore, the diversity of model updates is reduced. 
Very deep models only fit the average and do not adapt to finer reconstruction errors, 
i.e., they have limited susceptibility. 
Furthermore, we expect large-scale features expressed by sums to remain more distinguishable after averaging or coarse-graining, relative to small-scale features. 

We quantified learning on multiple spatial (coarse-graining) scales by using a microscopic definition of local magnetization (the spin) and energy (the average spin gradient), which we averaged across progressively larger square areas. 
The resulting scale-dependent loss evolution revealed two distinct dynamical regimes in which either the magnetization or the energy losses decrease monotonically across all scales. 
The transition between these regimes is controlled by the model depth, width, and learning rate. 
These regimes were also identified with learning the respective concepts: first, magnetization, an average scalar value, and then energy, an average gradient, which relies on small-scale features. 
Interestingly, the scale-invariant dynamical regimes coincide with the generalization of the related physical concepts. 

For shallower models ($\depth=1, 4$), this transition coincides with a representational trade-off, marked by a global minimum of the large-scale losses. 
As this trough is lifted with increasing bottleneck size, we conclude that the bottleneck constraint is the cause of the trade-off. 
The trade-off is between representing large scales to approximate many samples using collective features, and small scales to differentiate individual samples. 

Deep models trained at intermediate and fast rates become arrested across all scales and fail to reach these trade-off regimes, which are ultimately essential for learning the energy representation. 
They exhibit highly chaotic dynamics, reflected in their path-dependence and greater sensitivity to combinations in initialization, training data ordering, and batch size (see run variance in \figref{fig:variance-run-spins}). 
In contrast, the sensitivity of shallower models was two orders of magnitude lower. 
It reaches high values during transitions in the loss and low values when crossing plateaus, where trajectories converge. 
Very deep models exhibit largely the same run variance across all scales, indicating a mode collapse in susceptibility. 
For models that fail to learn, the final run variance vanishes; all models predict the dataset average. 

To provide a detailed picture of magnetization and energy in the output space, we studied the evolution of their joint input-output distribution. 
The output energy distribution is initially determined by the magnetization. 
It broadens as smaller scales are resolved, which only become visible in \figref{fig:high-loss-outputs} after the magnetization is learned (see \figref{fig:spin-pooling}). 
As finer-grained fluctuations are resolved, the output energy distribution shifts towards higher energies. 
However, these remain below an upper limit: the energy cutoff is influenced by the bottleneck, which induces a scale-resolution trade-off. 
The latter is particularly evident from the heightened loss at the largest energies, where small-scale fluctuations dominate. 
The energy cutoff shifted to higher energies when the bottleneck was expanded (\figref{fig:io-energy} to \ref{fig:io-energy-b64}). 
The learning rate and model depth appear to have the strongest effect on the characteristics of the learning dynamics (see also supplementary materials \cite{DARUS-6128_2026}).

We introduced a novel analysis using recursive trajectories to refine the dynamical perspective on learning. 
This approach demonstrates how prediction errors induce flow fields that generate representation dynamics. 
A transient `topological' stability of the latent space is maintained when learning the magnetization. 
At the same time, a slight drift under recursion was observed (\figref{fig:latent-flow-magnetization-b8}), vanishing after the concept is learned. 
Thereafter, fluctuations are learned across all scales, which underlie the generalization of energy (\figref{fig:io-energy}); trajectories expand into additional latent-space dimensions, thereby deteriorating the magnetization concept and its generalization (see \figref{fig:io-magnetization}). 
The utilized Spearman correlation highlights the intrinsic stability of concepts, which are otherwise hidden. 
Though self-recursion is categorically distinct from training dynamics, it nonetheless shows transient and metastable states; possibly related to those encountered during training: 
The formation of the magnetization concept is pinpointed by transitions from non- or metastable to more stable structures in latent space. 
Notably, such transitions are characteristic of complex non-equilibrium phase transformations. 
For other concepts, like energy, different measures of stability are needed.

In this study, we built on the intuition that learning operates as a process driven far from equilibrium by fluctuations in the training data. 
The model's evolution is controlled by memory-dependent susceptibilities induced by the optimizer. 
These dynamical correspondences between machine learning and physical processes form an analogy that grounds artificial intelligence in physics. 
Representation dynamics can be effectively described in terms of coarse-grained observables underlying the physical data-generating process. 
Observing scale-bridging dynamics as in this study underlines the collective nature of learning \cite{Hopfield1982}. 
Our insights should transfer to computing that directly exploits physical mechanisms \cite{jaegerFormalTheoryComputing2023a}. 
The emergence of physical concepts in complex systems -- from artificial to natural neural networks -- highlights profound connections between human understanding and other forms of `intelligence'.

\begin{acknowledgements}
Funded by the WIN Program of the Heidelberg Academy of Sciences and Humanities as well as by ``Künstliche Intelligenz \& Gesellschaft: Reflecting Intelligent Systems for Diversity, Demography and Democracy'' (IRIS3D), both
financed by the Ministry of Science, Research and the Arts Baden-Württemberg (Az. 33-7533-9-19/54/5), 
and by Deutsche Forschungsgemeinschaft (DFG, German Research Foundation) under Germany's Excellence Strategy -- EXC 2075 -- 390740016.
We acknowledge further support of the Stuttgart Center for Simulation Science (SimTech), the Interchange Forum for Reflecting on Intelligent Systems (IRIS) at the University of Stuttgart, and the International Max Planck Research School for Intelligent Systems (IMPRS-IS). We acknowledge ideation exchanges with Maria Wirzberger and Eric Raidl, supported by Joachim Stein, and thank Yannick Mühlhäuser for a preliminary exploration.
\end{acknowledgements}

\section*{Data Availability}
The code to generate the data and the model checkpoints used in this analysis, the generated data and model checkpoints to reproduce the experiments described in the article, and additional plots for other hyperparameters and datasets are publicly available on DaRUS \cite{DARUS-6128_2026}.

\bibliography{references}

@book{Zinn-Justin-book-2007,
  title={Phase Transitions and Renormalization Group},
  author={Zinn-Justin, Jean},
  year={2007},
  publisher={Oxford University Press},
  location={Oxfrod},
  series={Oxford Graduate Texts}
}

@book{Kardar-book-2007,
  title={Statistical Physics of Fields},
  author={Kardar, Mehran},
  year={2007},
  publisher={Cambridge University Press}
}

@article{Balian1986,
  title = {Dissipation in many-body systems: A geometric approach based on information theory},
  volume = {131},
  ISSN = {0370-1573},
  url = {http://dx.doi.org/10.1016/0370-1573(86)90005-0},
  DOI = {10.1016/0370-1573(86)90005-0},
  number = {1-2},
  journal = {Physics Reports},
  publisher = {Elsevier BV},
  author = {Balian,  Roger and Alhassid,  Yoram and Reinhardt,  Hugo},
  year = {1986},
  month = Jan,
  pages = {1–146}
}

@article{Hohenberg1977,
  title = {Theory of dynamic critical phenomena},
  volume = {49},
  ISSN = {0034-6861},
  url = {http://dx.doi.org/10.1103/RevModPhys.49.435},
  DOI = {10.1103/revmodphys.49.435},
  number = {3},
  journal = {Reviews of Modern Physics},
  publisher = {American Physical Society (APS)},
  author = {Hohenberg,  P. C. and Halperin,  B. I.},
  year = {1977},
  month = July,
  pages = {435–479}
}

@book{Dhont-book-1996,
  title={An Introduction to Dynamics of Colloids},
  author={Dhont, Jan KG},
  volume={2},
  year={1996},
  publisher={Elsevier}
}

@article{Das2004,
  title = {Mode-coupling theory and the glass transition in supercooled liquids},
  volume = {76},
  ISSN = {1539-0756},
  url = {http://dx.doi.org/10.1103/RevModPhys.76.785},
  DOI = {10.1103/revmodphys.76.785},
  number = {3},
  journal = {Reviews of Modern Physics},
  publisher = {American Physical Society (APS)},
  author = {Das,  Shankar P.},
  year = {2004},
  month = Oct,
  pages = {785–851}
}

@article{Hopfield1982,
  title = {Neural networks and physical systems with emergent collective computational abilities.},
  volume = {79},
  ISSN = {1091-6490},
  url = {http://dx.doi.org/10.1073/pnas.79.8.2554},
  DOI = {10.1073/pnas.79.8.2554},
  number = {8},
  journal = {Proceedings of the National Academy of Sciences},
  publisher = {National Academy of Sciences},
  author = {Hopfield,  J J},
  year = {1982},
  month = Apr,
  pages = {2554–2558}
}

@article{alexandrouCriticalTemperature2DIsing2020,
  title = {The Critical Temperature of the {{2D-Ising}} Model through {{Deep Learning Autoencoders}}},
  author = {Alexandrou, Constantia and Athenodorou, Andreas and Chrysostomou, Charalambos and Paul, Srijit},
  year = 2020,
  month = dec,
  journal = {Eur. Phys. J. B},
  volume = {93},
  number = {12},
  eprint = {1903.03506},
  primaryclass = {cond-mat},
  pages = {226},
  issn = {1434-6028, 1434-6036},
  doi = {10.1140/epjb/e2020-100506-5},
  url = {http://arxiv.org/abs/1903.03506},
  urldate = {2026-01-09},
  archiveprefix = {arXiv}
}

@book{alpaydinIntroductionMachineLearning2010,
  title = {Introduction to Machine Learning},
  author = {Alpaydin, Ethem},
  year = 2010,
  series = {Adaptive Computation and Machine Learning},
  edition = {2nd ed},
  publisher = {MIT Press},
  address = {Cambridge, Mass},
  isbn = {978-0-262-01243-0},
  lccn = {Q325.5 .A46 2010}
}

@inproceedings{anselPyTorch2Faster2024,
  title = {{{PyTorch}} 2: {{Faster Machine Learning Through Dynamic Python Bytecode Transformation}} and {{Graph Compilation}}},
  shorttitle = {{{PyTorch}} 2},
  booktitle = {Proceedings of the 29th {{ACM International Conference}} on {{Architectural Support}} for {{Programming Languages}} and {{Operating Systems}}, {{Volume}} 2},
  author = {Ansel, Jason and Yang, Edward and He, Horace and Gimelshein, Natalia and Jain, Animesh and Voznesensky, Michael and Bao, Bin and Bell, Peter and Berard, David and Burovski, Evgeni and Chauhan, Geeta and Chourdia, Anjali and Constable, Will and Desmaison, Alban and DeVito, Zachary and Ellison, Elias and Feng, Will and Gong, Jiong and Gschwind, Michael and Hirsh, Brian and Huang, Sherlock and Kalambarkar, Kshiteej and Kirsch, Laurent and Lazos, Michael and Lezcano, Mario and Liang, Yanbo and Liang, Jason and Lu, Yinghai and Luk, C. K. and Maher, Bert and Pan, Yunjie and Puhrsch, Christian and Reso, Matthias and Saroufim, Mark and Siraichi, Marcos Yukio and Suk, Helen and Zhang, Shunting and Suo, Michael and Tillet, Phil and Zhao, Xu and Wang, Eikan and Zhou, Keren and Zou, Richard and Wang, Xiaodong and Mathews, Ajit and Wen, William and Chanan, Gregory and Wu, Peng and Chintala, Soumith},
  year = 2024,
  month = apr,
  pages = {929--947},
  publisher = {ACM},
  address = {La Jolla CA USA},
  doi = {10.1145/3620665.3640366},
  url = {https://dl.acm.org/doi/10.1145/3620665.3640366},
  urldate = {2026-05-27},
  isbn = {979-8-4007-0385-0},
  langid = {english}
}

@article{baldiNeuralNetworksPrincipal1989,
  title = {Neural Networks and Principal Component Analysis: {{Learning}} from Examples without Local Minima},
  shorttitle = {Neural Networks and Principal Component Analysis},
  author = {Baldi, Pierre and Hornik, Kurt},
  year = 1989,
  month = jan,
  journal = {Neural Networks},
  volume = {2},
  number = {1},
  pages = {53--58},
  issn = {0893-6080},
  doi = {10.1016/0893-6080(89)90014-2},
  url = {https://www.sciencedirect.com/science/article/pii/0893608089900142},
  urldate = {2026-01-12},
  langid = {american}
}

@book{bishopPatternRecognitionMachine2006,
  title = {Pattern Recognition and Machine Learning},
  author = {Bishop, Christopher M.},
  year = 2006,
  series = {Information Science and Statistics},
  publisher = {Springer},
  address = {New York},
  isbn = {978-0-387-31073-2},
  lccn = {Q327 .B52 2006}
}

@article{carrasquillaMachineLearningPhases2017,
  title = {Machine Learning Phases of Matter},
  author = {Carrasquilla, Juan and Melko, Roger G.},
  year = 2017,
  month = may,
  journal = {Nature Phys},
  volume = {13},
  number = {5},
  pages = {431--434},
  publisher = {Nature Publishing Group},
  issn = {1745-2481},
  doi = {10.1038/nphys4035},
  url = {https://www.nature.com/articles/nphys4035},
  urldate = {2026-01-09},
  copyright = {2017 Springer Nature Limited},
  langid = {english}
}

@incollection{cauchyFormulesQuiResultent2009,
  title = {Sur Les Formules Qui R\'esultent de l'emploi Du Signe {$>$} Ou{$<$}, et Sur Les Moyennes Entre Plusieurs Quantit\'es},
  booktitle = {Cours d'analyse de l'{{\'Ecole Royale Polytechnique}}},
  editor = {Cauchy, Augustin-Louis},
  year = 2009,
  series = {Cambridge {{Library Collection}} - {{Mathematics}}},
  pages = {438--459},
  publisher = {Cambridge University Press},
  address = {Cambridge},
  doi = {10.1017/CBO9780511693328.017},
  url = {https://www.cambridge.org/core/books/cours-danalyse-de-lecole-royale-polytechnique/sur-les-formules-qui-resultent-de-lemploi-du-signe-ou-et-sur-les-moyennes-entre-plusieurs-quantites/017A77D0CB14B2F29DC20D9C495B260B},
  urldate = {2026-06-22},
  isbn = {978-1-108-00208-0}
}

@book{cohenAppliedMultipleRegression2013,
  title = {Applied {{Multiple Regression}}/{{Correlation Analysis}} for the {{Behavioral Sciences}}},
  author = {Cohen, Jacob and Cohen, Patricia and West, Stephen G. and Aiken, Leona S.},
  year = 2013,
  month = jun,
  edition = {3},
  publisher = {Routledge},
  address = {New York},
  doi = {10.4324/9780203774441},
  isbn = {978-0-203-77444-1}
}

@article{dangeloLearningIsingModel2020,
  title = {Learning the {{Ising}} Model with Generative Neural Networks},
  author = {D'Angelo, Francesco and B{\"o}ttcher, Lucas},
  year = 2020,
  month = jun,
  journal = {Phys. Rev. Res.},
  volume = {2},
  number = {2},
  pages = {023266},
  publisher = {American Physical Society},
  doi = {10.1103/PhysRevResearch.2.023266},
  url = {https://link.aps.org/doi/10.1103/PhysRevResearch.2.023266},
  urldate = {2026-01-09},
  langid = {american}
}

@data{DARUS-6128_2026,
  title = {Replication Data for: {{Nonequilibrium}} Dynamics in Autoencoder Architectures: {{Learning}} Concepts of the {{2D}} Ising Model},
  author = {Weinmann, Max and Klopotek, Miriam},
  year = 2026,
  publisher = {DaRUS},
  doi = {10.18419/DARUS-6128},
  url = {https://doi.org/10.18419/DARUS-6128}
}

@incollection{gidelImplicitRegularizationDiscrete2019,
  title = {Implicit Regularization of Discrete Gradient Dynamics in Linear Neural Networks},
  booktitle = {Proceedings of the 33rd {{International Conference}} on {{Neural Information Processing Systems}}},
  author = {Gidel, Gauthier and Bach, Francis and {Lacoste-Julien}, Simon},
  year = 2019,
  month = dec,
  number = {288},
  pages = {3202--3211},
  publisher = {Curran Associates Inc.},
  address = {Red Hook, NY, USA},
  url = {https://dl.acm.org/doi/10.5555/3454287.3454575},
  urldate = {2026-01-12},
  langid = {american}
}

@article{glauberTimeDependentStatisticsIsing1963,
  title = {Time-{{Dependent Statistics}} of the {{Ising Model}}},
  author = {Glauber, Roy J.},
  year = 1963,
  month = feb,
  journal = {Journal of Mathematical Physics},
  volume = {4},
  number = {2},
  pages = {294--307},
  issn = {0022-2488, 1089-7658},
  doi = {10.1063/1.1703954},
  url = {https://pubs.aip.org/jmp/article/4/2/294/230204/Time-Dependent-Statistics-of-the-Ising-Model},
  urldate = {2026-05-27},
  langid = {english}
}

@article{goldLanguageIdentificationLimit1967,
  title = {Language Identification in the Limit},
  author = {Gold, E Mark},
  year = 1967,
  month = may,
  journal = {Information and Control},
  volume = {10},
  number = {5},
  pages = {447--474},
  issn = {0019-9958},
  doi = {10.1016/S0019-9958(67)91165-5},
  url = {https://www.sciencedirect.com/science/article/pii/S0019995867911655},
  urldate = {2026-07-07}
}

@article{goldtModelingInfluenceData2020,
  title = {Modeling the {{Influence}} of {{Data Structure}} on {{Learning}} in {{Neural Networks}}: {{The Hidden Manifold Model}}},
  shorttitle = {Modeling the {{Influence}} of {{Data Structure}} on {{Learning}} in {{Neural Networks}}},
  author = {Goldt, Sebastian and M{\'e}zard, Marc and Krzakala, Florent and Zdeborov{\'a}, Lenka},
  year = 2020,
  month = dec,
  journal = {Phys. Rev. X},
  volume = {10},
  number = {4},
  pages = {041044},
  publisher = {American Physical Society},
  doi = {10.1103/PhysRevX.10.041044},
  url = {https://link.aps.org/doi/10.1103/PhysRevX.10.041044},
  urldate = {2026-06-29}
}

@article{golubSingularValueDecomposition1970,
  title = {Singular Value Decomposition and Least Squares Solutions},
  author = {Golub, G. H. and Reinsch, C.},
  year = 1970,
  month = apr,
  journal = {Numer. Math.},
  volume = {14},
  number = {5},
  pages = {403--420},
  issn = {0945-3245},
  doi = {10.1007/BF02163027},
  url = {https://doi.org/10.1007/BF02163027},
  urldate = {2026-07-07},
  langid = {english}
}

@misc{heDelvingDeepRectifiers2015,
  title = {Delving {{Deep}} into {{Rectifiers}}: {{Surpassing Human-Level Performance}} on {{ImageNet Classification}}},
  shorttitle = {Delving {{Deep}} into {{Rectifiers}}},
  author = {He, Kaiming and Zhang, Xiangyu and Ren, Shaoqing and Sun, Jian},
  year = 2015,
  month = feb,
  number = {arXiv:1502.01852},
  eprint = {1502.01852},
  primaryclass = {cs.CV},
  publisher = {arXiv},
  doi = {10.48550/arXiv.1502.01852},
  url = {http://arxiv.org/abs/1502.01852},
  urldate = {2026-07-07},
  archiveprefix = {arXiv}
}

@article{householderTheorySteadystateActivity1941,
  title = {A Theory of Steady-State Activity in Nerve-Fiber Networks: {{I}}. {{Definitions}} and Preliminary Lemmas},
  shorttitle = {A Theory of Steady-State Activity in Nerve-Fiber Networks},
  author = {Householder, Alston S.},
  year = 1941,
  month = jun,
  journal = {Bulletin of Mathematical Biophysics},
  volume = {3},
  number = {2},
  pages = {63--69},
  issn = {1522-9602},
  doi = {10.1007/BF02478220},
  url = {https://doi.org/10.1007/BF02478220},
  urldate = {2026-07-08},
  langid = {english}
}

@article{huDiscoveringPhasesPhase2017,
  title = {Discovering Phases, Phase Transitions, and Crossovers through Unsupervised Machine Learning: {{A}} Critical Examination},
  shorttitle = {Discovering Phases, Phase Transitions, and Crossovers through Unsupervised Machine Learning},
  author = {Hu, Wenjian and Singh, Rajiv R. P. and Scalettar, Richard T.},
  year = 2017,
  month = jun,
  journal = {Phys. Rev. E},
  volume = {95},
  number = {6},
  pages = {062122},
  publisher = {American Physical Society},
  doi = {10.1103/PhysRevE.95.062122},
  url = {https://link.aps.org/doi/10.1103/PhysRevE.95.062122},
  urldate = {2024-12-02},
  langid = {american}
}

@article{isingBeitragZurTheorie1925,
  title = {{Beitrag zur Theorie des Ferromagnetismus}},
  author = {Ising, Ernst},
  year = 1925,
  month = feb,
  journal = {Z. Physik},
  volume = {31},
  number = {1},
  pages = {253--258},
  issn = {0044-3328},
  doi = {10.1007/BF02980577},
  url = {https://doi.org/10.1007/BF02980577},
  urldate = {2025-12-08},
  langid = {ngerman}
}

@article{kieferStochasticEstimationMaximum1952,
  title = {Stochastic {{Estimation}} of the {{Maximum}} of a {{Regression Function}}},
  author = {Kiefer, J. and Wolfowitz, J.},
  year = 1952,
  month = sep,
  journal = {The Annals of Mathematical Statistics},
  volume = {23},
  number = {3},
  pages = {462--466},
  publisher = {Institute of Mathematical Statistics},
  issn = {0003-4851, 2168-8990},
  doi = {10.1214/aoms/1177729392},
  url = {https://projecteuclid.org/journals/annals-of-mathematical-statistics/volume-23/issue-3/Stochastic-Estimation-of-the-Maximum-of-a-Regression-Function/10.1214/aoms/1177729392.full},
  urldate = {2026-07-07},
  langid = {english}
}

@misc{kingmaAdamMethodStochastic2014,
  title = {Adam: {{A Method}} for {{Stochastic Optimization}}},
  shorttitle = {Adam},
  author = {Kingma, Diederik P. and Ba, Jimmy},
  year = 2014,
  publisher = {arXiv},
  doi = {10.48550/ARXIV.1412.6980},
  url = {https://arxiv.org/abs/1412.6980},
  urldate = {2026-05-27},
  copyright = {arXiv.org perpetual, non-exclusive license}
}

@article{kramerNonlinearPrincipalComponent1991,
  title = {Nonlinear Principal Component Analysis Using Autoassociative Neural Networks},
  author = {Kramer, Mark A.},
  year = 1991,
  journal = {AIChE Journal},
  volume = {37},
  number = {2},
  pages = {233--243},
  issn = {1547-5905},
  doi = {10.1002/aic.690370209},
  url = {https://onlinelibrary.wiley.com/doi/abs/10.1002/aic.690370209},
  urldate = {2026-01-12},
  copyright = {Copyright \copyright{} 1991 American Institute of Chemical Engineers},
  langid = {english}
}

@inproceedings{kuninLossLandscapesRegularized2019,
  title = {Loss {{Landscapes}} of {{Regularized Linear Autoencoders}}},
  booktitle = {Proceedings of the 36th {{International Conference}} on {{Machine Learning}}},
  author = {Kunin, Daniel and Bloom, Jonathan and Goeva, Aleksandrina and Seed, Cotton},
  year = 2019,
  month = may,
  pages = {3560--3569},
  publisher = {PMLR},
  issn = {2640-3498},
  url = {https://proceedings.mlr.press/v97/kunin19a.html},
  urldate = {2026-01-12},
  langid = {english}
}

@misc{loshchilovSGDRStochasticGradient2017,
  title = {{{SGDR}}: {{Stochastic Gradient Descent}} with {{Warm Restarts}}},
  shorttitle = {{{SGDR}}},
  author = {Loshchilov, Ilya and Hutter, Frank},
  year = 2017,
  month = may,
  number = {arXiv:1608.03983},
  eprint = {1608.03983},
  primaryclass = {cs.LG},
  publisher = {arXiv},
  doi = {10.48550/arXiv.1608.03983},
  url = {http://arxiv.org/abs/1608.03983},
  urldate = {2026-07-07},
  archiveprefix = {arXiv}
}

@book{mohriFoundationsMachineLearning2012,
  title = {Foundations of Machine Learning},
  author = {Mohri, Mehryar and Rostamizadeh, Afshin and Talwalkar, Ameet},
  year = 2012,
  series = {Adaptive Computation and Machine Learning Series},
  publisher = {MIT Press},
  address = {Cambridge, MA},
  isbn = {978-0-262-01825-8},
  lccn = {Q325.5 .M64 2012}
}

@inproceedings{NIPS1993_9e3cfc48,
  title = {Autoencoders, Minimum Description Length and Helmholtz Free Energy},
  booktitle = {Advances in Neural Information Processing Systems},
  author = {Hinton, Geoffrey E and Zemel, Richard},
  editor = {Cowan, J. and Tesauro, G. and Alspector, J.},
  year = 1993,
  volume = {6},
  publisher = {Morgan-Kaufmann},
  url = {https://proceedings.neurips.cc/paper_files/paper/1993/file/9e3cfc48eccf81a0d57663e129aef3cb-Paper.pdf}
}

@article{pearsonLIIILinesPlanes1901,
  title = {{{LIII}}. {{On}} Lines and Planes of Closest Fit to Systems of Points in Space},
  author = {Pearson, Karl},
  year = 1901,
  month = nov,
  journal = {The London, Edinburgh, and Dublin Philosophical Magazine and Journal of Science},
  volume = {2},
  number = {11},
  pages = {559--572},
  publisher = {Taylor \& Francis},
  issn = {1941-5982},
  doi = {10.1080/14786440109462720},
  url = {https://doi.org/10.1080/14786440109462720},
  urldate = {2026-07-07}
}

@misc{raghuExpressivePowerDeep2017,
  title = {On the {{Expressive Power}} of {{Deep Neural Networks}}},
  author = {Raghu, Maithra and Poole, Ben and Kleinberg, Jon and Ganguli, Surya and {Sohl-Dickstein}, Jascha},
  year = 2017,
  month = jun,
  number = {arXiv:1606.05336},
  eprint = {1606.05336},
  primaryclass = {stat.ML},
  publisher = {arXiv},
  doi = {10.48550/arXiv.1606.05336},
  url = {http://arxiv.org/abs/1606.05336},
  urldate = {2026-06-30},
  archiveprefix = {arXiv}
}

@article{refinettiDynamicsRepresentationLearning2023,
  title = {The Dynamics of Representation Learning in Shallow, Non-Linear Autoencoders*},
  author = {Refinetti, Maria and Goldt, Sebastian},
  year = 2023,
  month = nov,
  journal = {J. Stat. Mech.},
  volume = {2023},
  number = {11},
  pages = {114010},
  publisher = {IOP Publishing},
  issn = {1742-5468},
  doi = {10.1088/1742-5468/ad01af},
  url = {https://doi.org/10.1088/1742-5468/ad01af},
  urldate = {2026-01-12},
  langid = {english}
}

@article{robbinsStochasticApproximationMethod1951,
  title = {A {{Stochastic Approximation Method}}},
  author = {Robbins, Herbert and Monro, Sutton},
  year = 1951,
  month = sep,
  journal = {The Annals of Mathematical Statistics},
  volume = {22},
  number = {3},
  pages = {400--407},
  publisher = {Institute of Mathematical Statistics},
  issn = {0003-4851, 2168-8990},
  doi = {10.1214/aoms/1177729586},
  url = {https://projecteuclid.org/journals/annals-of-mathematical-statistics/volume-22/issue-3/A-Stochastic-Approximation-Method/10.1214/aoms/1177729586.full},
  urldate = {2026-07-07},
  langid = {english}
}

@book{robertsPrinciplesDeepLearning2022a,
  title = {The {{Principles}} of {{Deep Learning Theory}}},
  author = {Roberts, Daniel A. and Yaida, Sho and Hanin, Boris},
  year = 2022,
  month = may,
  eprint = {2106.10165},
  primaryclass = {cs.LG},
  doi = {10.1017/9781009023405},
  url = {http://arxiv.org/abs/2106.10165},
  urldate = {2026-05-15},
  archiveprefix = {arXiv},
  langid = {american}
}

@article{rosenblattPerceptronProbabilisticModel1958,
  title = {The Perceptron: {{A}} Probabilistic Model for Information Storage and Organization in the Brain},
  shorttitle = {The Perceptron},
  author = {Rosenblatt, F.},
  year = 1958,
  journal = {Psychological Review},
  volume = {65},
  number = {6},
  pages = {386--408},
  publisher = {American Psychological Association},
  address = {US},
  issn = {1939-1471},
  doi = {10.1037/h0042519}
}

@article{saxeMathematicalTheorySemantic2019,
  title = {A Mathematical Theory of Semantic Development in Deep Neural Networks},
  author = {Saxe, Andrew M. and McClelland, James L. and Ganguli, Surya},
  year = 2019,
  month = jun,
  journal = {Proceedings of the National Academy of Sciences},
  volume = {116},
  number = {23},
  pages = {11537--11546},
  publisher = {Proceedings of the National Academy of Sciences},
  doi = {10.1073/pnas.1820226116},
  url = {https://www.pnas.org/doi/full/10.1073/pnas.1820226116},
  urldate = {2026-01-12},
  langid = {american}
}

@article{solomonoffFormalTheoryInductive1964,
  title = {A Formal Theory of Inductive Inference. {{Part I}}},
  author = {Solomonoff, R. J.},
  year = 1964,
  month = mar,
  journal = {Information and Control},
  volume = {7},
  number = {1},
  pages = {1--22},
  issn = {0019-9958},
  doi = {10.1016/S0019-9958(64)90223-2},
  url = {https://www.sciencedirect.com/science/article/pii/S0019995864902232},
  urldate = {2026-07-07}
}

@misc{tishbyInformationBottleneckMethod2000,
  title = {The Information Bottleneck Method},
  author = {Tishby, Naftali and Pereira, Fernando C. and Bialek, William},
  year = 2000,
  month = apr,
  number = {arXiv:physics/0004057},
  eprint = {physics/0004057},
  publisher = {arXiv},
  url = {http://arxiv.org/abs/physics/0004057},
  urldate = {2024-01-16},
  archiveprefix = {arXiv},
  langid = {english}
}

@article{valiantTheoryLearnable1984,
  title = {A Theory of the Learnable},
  author = {Valiant, L. G.},
  year = 1984,
  month = nov,
  journal = {Commun. ACM},
  volume = {27},
  number = {11},
  pages = {1134--1142},
  issn = {0001-0782},
  doi = {10.1145/1968.1972},
  url = {https://dl.acm.org/doi/10.1145/1968.1972},
  urldate = {2026-07-07}
}

@book{vapnikNatureStatisticalLearning1995,
  title = {The {{Nature}} of {{Statistical Learning Theory}}},
  author = {Vapnik, Vladimir Naumovich},
  year = 1995,
  edition = {1st ed},
  publisher = {Springer New York},
  address = {New York},
  isbn = {978-1-4757-2440-0},
  langid = {english}
}

@article{vapnikUniformConvergenceRelative1971,
  title = {On the {{Uniform Convergence}} of {{Relative Frequencies}} of {{Events}} to {{Their Probabilities}}},
  author = {Vapnik, V. N. and Chervonenkis, A. {\relax Ya}.},
  year = 1971,
  month = jan,
  journal = {Theory Probab. Appl.},
  volume = {16},
  number = {2},
  pages = {264--280},
  publisher = {{Society for Industrial and Applied Mathematics}},
  issn = {0040-585X},
  doi = {10.1137/1116025},
  url = {https://epubs.siam.org/doi/10.1137/1116025},
  urldate = {2026-07-07}
}

@article{wangDiscoveringPhaseTransitions2016,
  title = {Discovering Phase Transitions with Unsupervised Learning},
  author = {Wang, Lei},
  year = 2016,
  month = nov,
  journal = {Phys. Rev. B},
  volume = {94},
  number = {19},
  pages = {195105},
  publisher = {American Physical Society},
  doi = {10.1103/PhysRevB.94.195105},
  url = {https://link.aps.org/doi/10.1103/PhysRevB.94.195105},
  urldate = {2026-01-09},
  langid = {american}
}

@misc{wetzelInterpretableMachineLearning2025,
  title = {Interpretable {{Machine Learning}} in {{Physics}}: {{A Review}}},
  shorttitle = {Interpretable {{Machine Learning}} in {{Physics}}},
  author = {Wetzel, Sebastian Johann and Ha, Seungwoong and Iten, Raban and Klopotek, Miriam and Liu, Ziming},
  year = 2025,
  month = mar,
  number = {arXiv:2503.23616},
  eprint = {2503.23616},
  primaryclass = {physics.comp-ph},
  publisher = {arXiv},
  doi = {10.48550/arXiv.2503.23616},
  url = {http://arxiv.org/abs/2503.23616},
  urldate = {2026-07-07},
  archiveprefix = {arXiv}
}

@article{wetzelMachineLearningExplicit2017a,
  title = {Machine Learning of Explicit Order Parameters: {{From}} the {{Ising}} Model to {{SU}}(2) Lattice Gauge Theory},
  shorttitle = {Machine Learning of Explicit Order Parameters},
  author = {Wetzel, Sebastian J. and Scherzer, Manuel},
  year = 2017,
  month = nov,
  journal = {Phys. Rev. B},
  volume = {96},
  number = {18},
  pages = {184410},
  publisher = {American Physical Society},
  doi = {10.1103/PhysRevB.96.184410},
  url = {https://link.aps.org/doi/10.1103/PhysRevB.96.184410},
  urldate = {2026-01-09},
  langid = {american}
}

@article{wetzelUnsupervisedLearningPhase2017a,
  title = {Unsupervised Learning of Phase Transitions: {{From}} Principal Component Analysis to Variational Autoencoders},
  shorttitle = {Unsupervised Learning of Phase Transitions},
  author = {Wetzel, Sebastian J.},
  year = 2017,
  month = aug,
  journal = {Phys. Rev. E},
  volume = {96},
  number = {2},
  pages = {022140},
  publisher = {American Physical Society},
  doi = {10.1103/PhysRevE.96.022140},
  url = {https://link.aps.org/doi/10.1103/PhysRevE.96.022140},
  urldate = {2026-01-09}
}

@article{yevickVariationalAutoencoderAnalysis2022,
  title = {Variational Autoencoder Analysis of {{Ising}} Model Statistical Distributions and Phase Transitions},
  author = {Yevick, David},
  year = 2022,
  month = mar,
  journal = {Eur. Phys. J. B},
  volume = {95},
  number = {3},
  pages = {56},
  issn = {1434-6036},
  doi = {10.1140/epjb/s10051-022-00296-y},
  url = {https://doi.org/10.1140/epjb/s10051-022-00296-y},
  urldate = {2026-01-09},
  langid = {english}
}

@article{yueIncrementalLearningPhase2022,
  title = {Incremental Learning of Phase Transition in {{Ising}} Model: {{Preprocessing}}, Finite-Size Scaling and Critical Exponents},
  shorttitle = {Incremental Learning of Phase Transition in {{Ising}} Model},
  author = {Yue, Zhenyi and Wang, Yuqi and Lyu, Pin},
  year = 2022,
  month = aug,
  journal = {Physica A: Statistical Mechanics and its Applications},
  volume = {600},
  pages = {127538},
  issn = {0378-4371},
  doi = {10.1016/j.physa.2022.127538},
  url = {https://www.sciencedirect.com/science/article/pii/S0378437122003776},
  urldate = {2026-01-09},
  langid = {american}
}

@article{jaegerFormalTheoryComputing2023a,
  title = {Toward a Formal Theory for Computing Machines Made out of Whatever Physics Offers},
  author = {Jaeger, Herbert and Noheda, Beatriz and {van der Wiel}, Wilfred G.},
  year = 2023,
  month = aug,
  journal = {Nat. Commun.},
  volume = {14},
  number = {1},
  pages = {4911},
  publisher = {Nature Publishing Group},
  issn = {2041-1723},
  doi = {10.1038/s41467-023-40533-1},
  url = {https://www.nature.com/articles/s41467-023-40533-1},
  urldate = {2025-02-20},
  copyright = {2023 The Author(s)},
  langid = {english}
}

\clearpage
\onecolumngrid
\section{Appendix}\label{sect:appendix}

\subsection{Data Distribution -- Data-Generating Process}\label{sect:appendix-data-distribution}

We provide all plots that describe the data distribution or the more general properties of the underlying data-generating process.

\begin{figure*}[ht]
    \centering
    \includegraphics[width=0.75\linewidth]{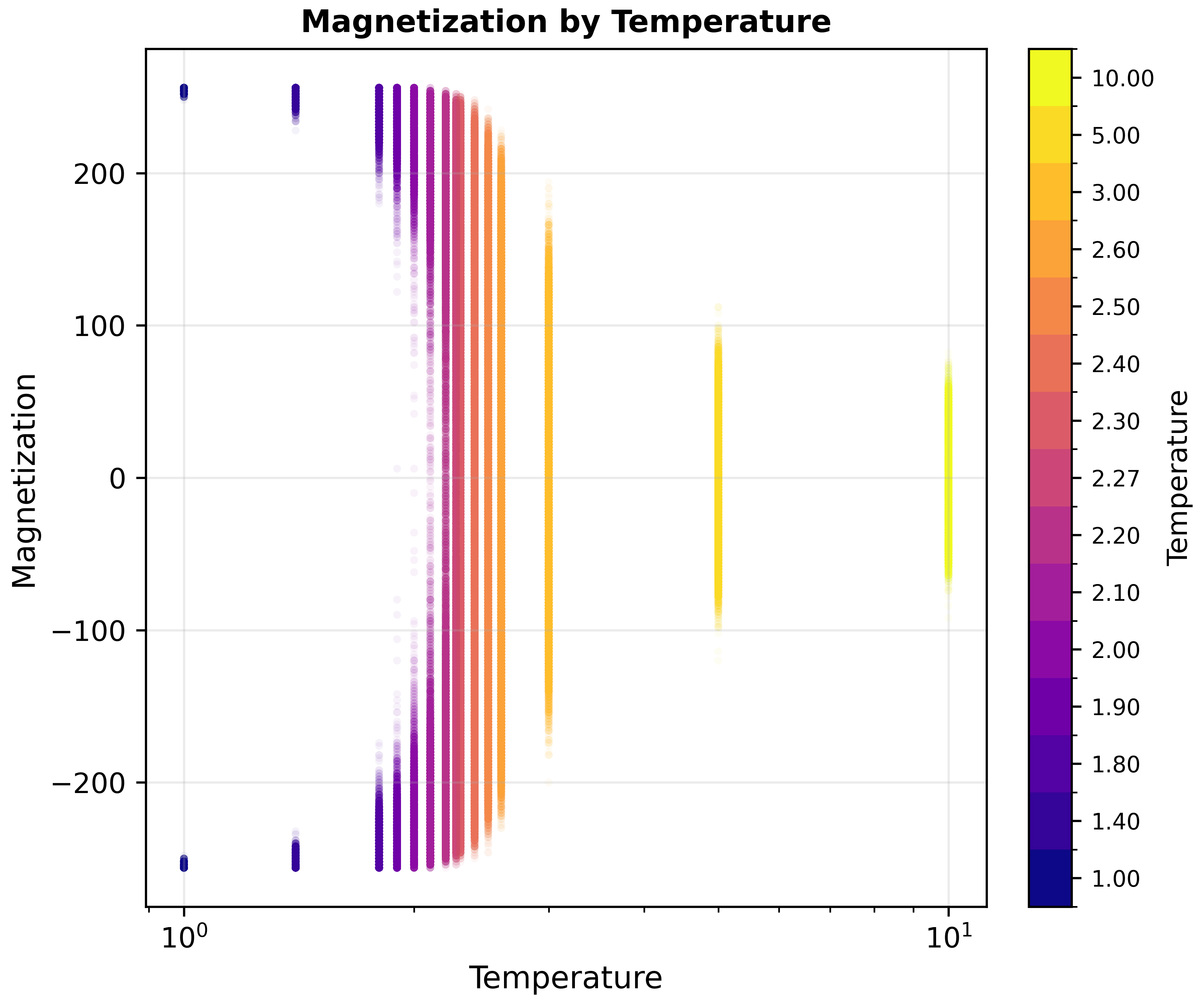}
    \caption{
    Temperature-dependent magnetization distribution of all datasets. 
    At low temperatures, two high absolute magnetization clusters dominate close to the extreme values ($\magnetization_{max}=\pm 256$). 
    The clusters become wider until they merge around the critical temperature ($\temperature=2.27$) and then become concentrated around zero magnetization at higher temperatures.
    }
    \label{fig:magnetization-over-temp}
\end{figure*}

\begin{figure*}[ht]
    \centering
    \includegraphics[width=0.75\linewidth]{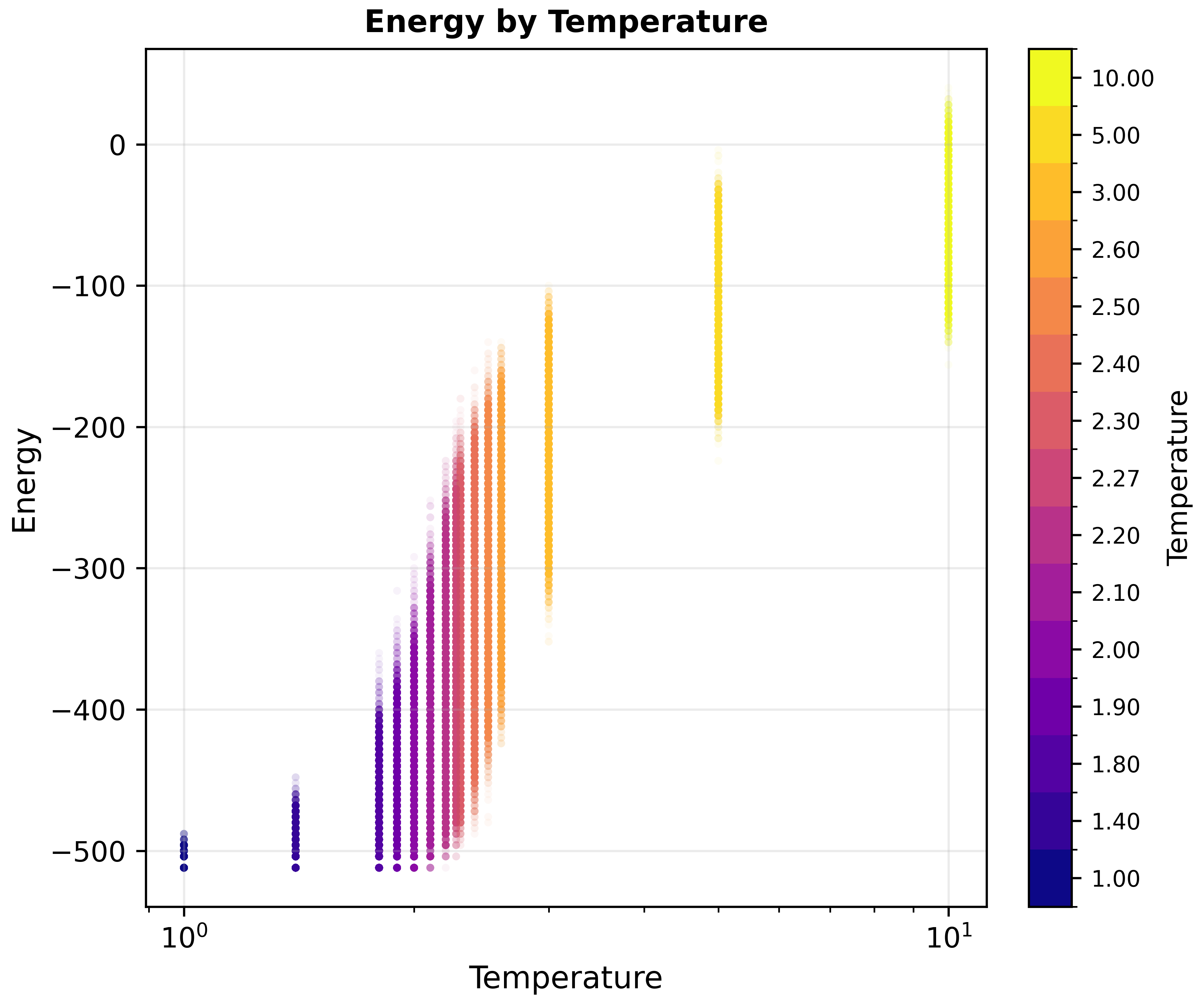}
    \caption{
    Temperature-dependent energy distribution of all datasets. 
    At low temperatures, the low energies dominate. 
    As the temperature rises, the distribution is shifted to energies around zero in the limit. 
    Until the transition temperature, the distribution widens and then narrows again.
    }
    \label{fig:energy-over-temp}
\end{figure*}

\begin{figure*}[ht]
    \centering
    \includegraphics[width=0.75\linewidth]{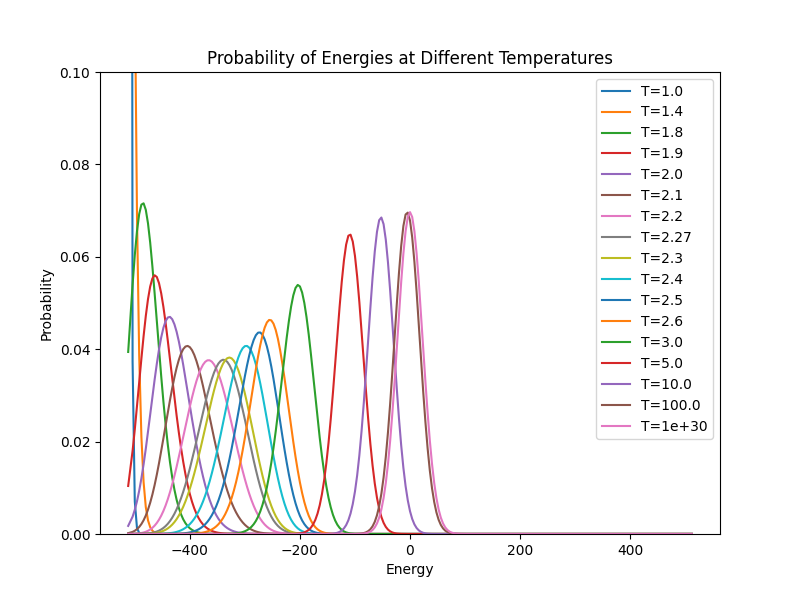}
    \caption{
    Theoretical distribution of energies for different temperatures. 
    The values are calculated using the analytical expression of the Boltzmann factor at the respective temperature and an estimate of the density of states across energies for a 16x16 lattice. 
    The estimate was obtained from a 3rd degree polynomial fitted to a histogram with 205 energy bins generated with the Wang-Landau algorithm.
    Before fitting, the histogram was normalized and manually augmented with correct theoretical values at the energy extremes, where statistics are poor due to the few samples. 
    At low temperatures, the distribution is dominated by low energies. 
    As the temperature rises, the distribution is shifted to energies around zero in the limit. 
    Until the transition temperature, the distribution widens and then narrows again towards the limiting width. 
    The distribution for $\temperature=100$ is almost identical to the infinite limit, where the distribution is fully defined by the density of states.
    }
    \label{fig:energy-dos}
\end{figure*}

\clearpage
\subsection{Training Configuration}\label{sect:appendix-training-configuration}

\subsubsection{Bottleneck Size}\label{sect:bottleneck-size}

The size of the lowest-dimensional latent space, the \textbf{bottleneck size} $\bottleneck$, can be viewed as a task choice, rather than an optimization hyperparameter. 
As the bottleneck size decreases, more nonlinearities are required to achieve better reconstruction performance. 
The stronger nonlinearities make gradients more parameter- and input-dependent, potentially increasing fluctuations in the learning dynamics. 
If the confinement reduces the number of distinguishable states beyond the number of training inputs, the reconstructions must be imperfect. 
In this scenario, the optimal trade-off is coarse-grained averages that represent the most probable collective modes, thereby minimizing the expected MSD.

In our study, we examined latent spaces with bottleneck sizes $\bottleneck=1$, $8$, and $64$ dimensions, yielding compression factors of 256, 32, and 4 relative to the input dimensions.
For a linear model, the latent space dimension limits the matrix rank $\matrixrank \leq \bottleneck$ of the linear transformation (linear rank) and thus the number of independent linear input features that can be reconstructed.
A linear transformation limited to the corresponding rank $\matrixrank=\bottleneck$ can explain 11\% (1), 18\% (8), and 45\% (64) of the variance in both the training and validation datasets (with the first $\bottleneck$ principal components).
The ratio of captured variance is directly connected to the complexity of the physical fluctuations inside the dataset.

The variance that can be explained by the linear rank corresponding to the bottleneck size ($\matrixrank=\bottleneck$) provides an upper bound for a linear model of the same dimension. 
To overcome this linear bound, additional nonlinearities are necessary, but not sufficient. 
Learning the resulting generally nonlinear input-dependent functions requires diverse data and sufficient nonlinear representation capacity, which increases with model depth and width.
Without sufficient nonlinear representation capacity, the bottleneck must couple or remove linear input features and can only capture a subset of all fluctuations.
Therefore, the different bottleneck sizes allow us to study the types of physical fluctuations that are synchronized and ultimately represented by the autoencoder over training.

\subsubsection{Model Width}\label{sect:model-width}

We keep the architecture simple by using the same width for all intermediate layers except for the bottleneck. 
This results in a uniform block structure for both the encoder and decoder. 
Similar to the bottleneck, the \textbf{model width} $\width$ influences the degree of linear compression or expansion within any layer. 
Theoretically, a greater width provides greater parallel processing capabilities, e.g., affine transformation channels, and the ability to retain more information, e.g., linear transformations of higher rank. 
Retaining information between layers about a given subspace, i.e., the data distribution, involves applying a transformation that is reversible in this subspace. 
This is fundamental to the overall learning behavior, especially in deep neural networks with many layers.
In each layer, independent input features that convey important information might be irreversibly mixed with others, limiting the reconstruction quality of autoencoders. 
Linear layers are reversible in their entire input space whenever the weight matrix has full rank (i.e., scaling, rotation, and mirroring transformations that retain the dimensionality).
The ReLU nonlinearity is only reversible for positive inputs, as all negative inputs are mapped to zero, removing fluctuations inside this region. 
This can cause models with width $\width=k$ to be on par with the PCA error using the top $k/2$ components \cite{refinettiDynamicsRepresentationLearning2023}. 
Theoretically, a sequence of layers (including nonlinearities) remains reversible for datasets where all activations are positive, and all weight matrices have full rank.
In the more general case with negative activations, reversibility is a more difficult question, depending on the independence of activations over the input domain. 

Given the finite range of input activations (0,1), all ReLU functions with a sufficiently large negative bias are theoretically reversible. However, weights and biases are initialized from a zero-centered distribution, likely leading to negative activations that are masked by the ReLU layers. 
By adding two ReLU layer outputs whose inputs are mirrored around the origin, a linear (identity) function can be obtained, which is reversible for any input. 

Therefore, we focus our choice of model configuration around the model class with a width of $\width=512$ dimensions (2 x input dimension), which inherently includes the purely linear model by design. 
Additionally, we selected a thinner model with $\width=256$ dimensions that is invertible only for half of the input domain, as well as a wider model with $\width=1024$ dimensions. 
The wider model can lead to greater variance in the projections, potentially increasing the risk of overfitting to the training data. 
The widest $\width=1024$ models have more than a million parameters per layer, resulting in high memory requirements for training checkpoints. 
However, saving these checkpoints is essential for analyzing the learning dynamics effectively.
Because memory requirements grow quadratically with the width, we don't evaluate any models larger than this.

\subsubsection{Model Depth}\label{sect:model-depth}

The depth of a model limits its capacity to perform a sequence of operations to generate an output. Theoretically, the maximum number of piecewise linear segments in a ReLU model's function grows exponentially with the depth \cite{raghuExpressivePowerDeep2017}. In contrast, this number increases only polynomially with the width of the layers \cite{raghuExpressivePowerDeep2017}. Therefore, depth is expected to be the key factor influencing the model's ability to fit non-linear functions. But as we will see, this may not imply better generalization by the model; this capacity may remain unused when the model is stuck in the arrested state we have identified. 
According to \refref{\cite{robertsPrinciplesDeepLearning2022a}}, the depth alone is less descriptive of a model's behavior both during initialization and during training. 
Instead, the authors claim that the depth-to-width ratio (representing a "relative depth") is the main factor controlling the non-Gaussianity of activations inside multi-layer perceptrons. 
This is used to distinguish different dynamical regimes. 
When the model is shallow compared to its width (not really deep), the learning dynamics are well described by a Gaussian model that assumes independent activations. 
In an intermediate regime, the activations become dependent but can be explained by a first-order perturbation with respect to inverse layer width, making the depth-to-width ratio a relevant order parameter \cite{robertsPrinciplesDeepLearning2022a}. 
When the model is overly deep compared to its width, fine-tuned initialization is required, and the model training becomes `chaotic', resulting in signals either blowing up or decaying. 
As the theory is not tailored to models with small bottlenecks, we intend to provide empirical evidence for this setting. 
Therefore, we cover a broad range of depths across all these regimes. 

Encoders and decoders are mirror-symmetric sub-models, and the `depth' refers to each part alone. 
In our setting, the \textbf{model depth} $\depth$ indicates the number of hidden layers and the number of nonlinearities in both the encoder and decoder, as neither has a non-linear output activation function.
A depth of 0 corresponds to an encoder and a decoder that consist of a single linear layer without activation functions.
A depth of 1 means that both sub-models have a linear layer followed by a non-linearity, and then another linear layer (see \figref{fig:auto-encoder}). 

To encompass all reasonable depths for a non-linear model, we consider depths of 1, 4, and 16. Models with depths greater than 16 (34 linear layers) would lead to lengthy gradient computations and evaluation times, which are not feasible for this investigation.
We will see that this range seems sufficient to reveal a regime in which, for some hyperparameters, all input signals decay regularly. 
Additionally, the performance of deeper models appears to be equal to or significantly worse than that of shallow models across a wide range of hyperparameters.

\subsubsection{Batch Size}\label{batch-size}

The \textbf{batch size} $\batchsize$ refers to the number of samples processed simultaneously to calculate model parameter gradients during each optimization step. 
A larger batch size reduces fluctuations in the driving forces across epochs because each batch is more likely to contain similar data points. 
This data sharing increases the number of shared features across batches within the same epoch. 
Together, these factors can gradually alter the optimization landscape during training. 
In the case of full batch gradient descent, the randomness (stochasticity) is eliminated, resulting in a static optimization landscape (in parameter space). 

For our investigation, we use batches of $\batchsize=30$, $300$, and $3000$ data points that are sampled from a training dataset of $300,000$ data points. 
These batch sizes ensure that memory requirements remain low enough for parallel computation on our GPU, regardless of model size. 
Consequently, the computational time primarily depends on the number of optimization steps. 

We are limiting the total number of steps to 100,000 across all batch sizes to ensure a fair comparison of computational time. 
Additionally, this setup enables us to examine how repeating the same data points 10, 100, and 1000 times might affect training. 
However, this effect may be confounded with the learning rate scheduler, which does not follow a repeating pattern. 

We decided against cycling the scheduler because it could significantly change the learning dynamics across different batch sizes, making them less comparable. 
Additionally, to ensure the integrated learning rate remains consistent, especially during the initial linear learning rate `warm-up' phase after model initialization, we would need to introduce an unusual learning rate schedule. 
This approach could limit the applicability of our findings to common practice scenarios.

\subsubsection{Learning Rate}\label{sect:learning-rate}

The \textbf{learning rate} determines the size of the steps taken during the optimization process. This factor affects how effectively the model can avoid settling into narrow, local minima: 
When the parameter fluctuations are adjusted, naturally, they alter the course of the learning dynamics -- such as the visitation of transitory and metastable states -- which directly affects how and which physical concepts are learned by the autoencoder (as we will see in \sectref{sect:results}). 
The learning rate can also be interpreted as the model's coupling strength with the external (driving) environment, as it controls how much the parameters change in response to the training data. 

There are three main ways the effective learning rate can be influenced:

\begin{itemize}
\item
  A global choice of the base learning rate at the initialization of the optimizer.
\item
  Time-dependent adjustments to the current learning rate through a learning rate scheduler.
\item
  Data-dependent modifications to the effective learning rate via gradient accumulation and scaling in the \emph{Adam} optimizer, based on exponential averages of the first and second moments.
\end{itemize}

In our implementation, the learning rate is scheduled to increase linearly from 1\% to 100\% of the base learning rate over 100 steps. This is followed by cosine annealing, which reduces the learning rate from 100\% at step 101 to 0\% at step 100,000.

A common approach to selecting an appropriate base \textbf{learning rate} $\learningrate$ is to use a learning rate finder. 
This method begins training with very low learning rates and then gradually increases them exponentially, all the while monitoring the loss response. 
The loss response can theoretically be influenced by various hyperparameters, such as model architecture, initialization, data batch size, and the order of data presentation to the architecture. 

\begin{figure*}[ht]
    \centering
    \includegraphics[width=0.75\linewidth]{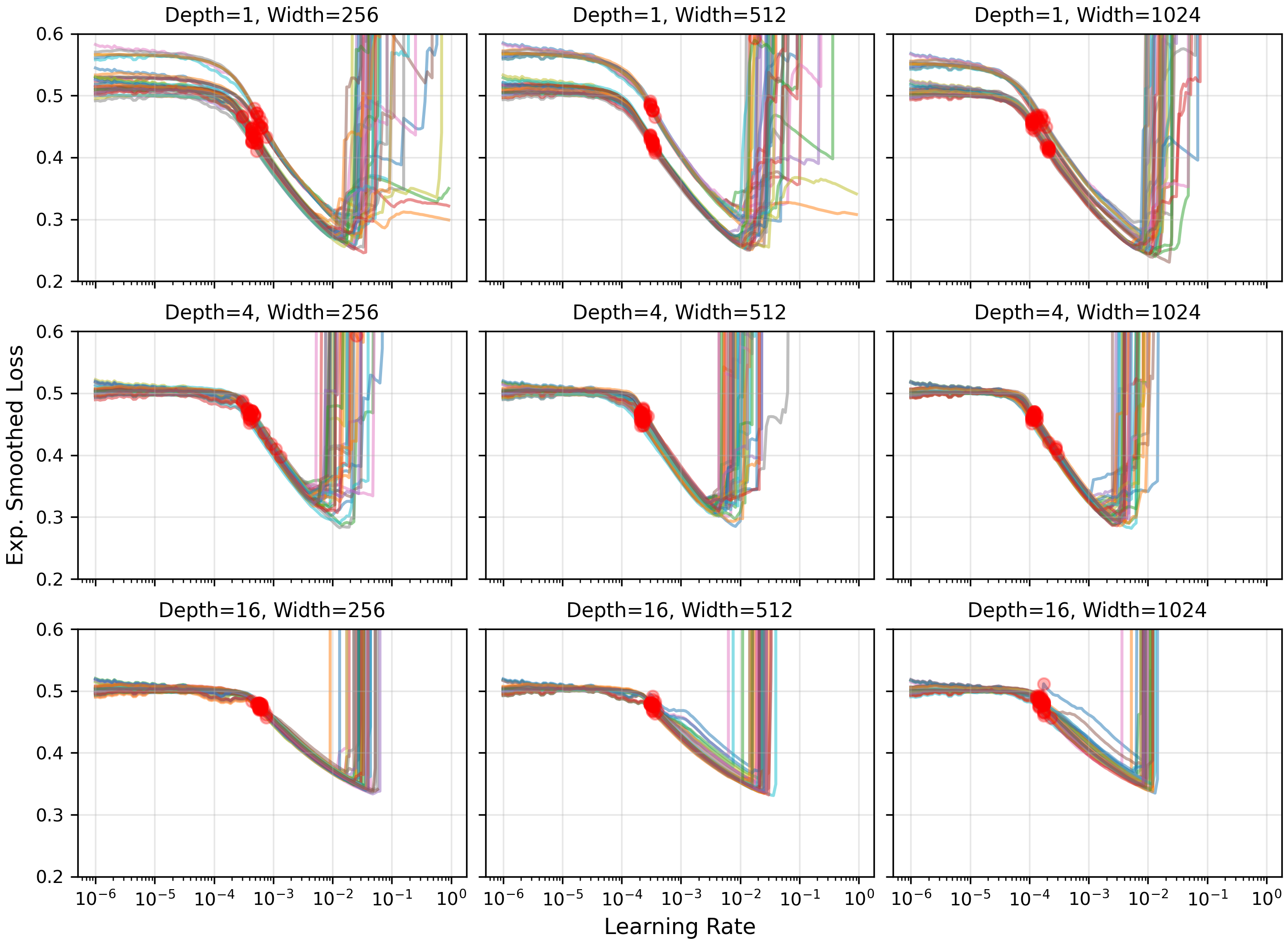}
    \caption{
    Exponentially smoothed batch loss for training with an exponentially increasing learning rate. 
    Each subplot shows the evolution of individual runs with a specific depth and width. 
    The steepest descent is marked with a red dot. 
    For low learning rates, the losses change little until they begin to decrease rapidly between $\learningrate=10^{-4}$ and $\learningrate=10^{-3}$. 
    Almost all runs have the steepest descent in this region. 
    With higher learning rates, the descent slows down before exploding rapidly around $\learningrate=10^{-2}$. 
    The loss stability and its fast descent provide an indicator to choose a good learning rate for the particular setup. 
    }
    \label{fig:lr-finding}
\end{figure*}

However, based on empirical observations, we find that the response curves are generally quite similar across most hyperparameter settings, differing mainly at higher learning rates where instability is anticipated (see \figref{fig:lr-finding}).
The learning rate is typically selected at the point where the gradient is lowest to avoid rates that are either too slow (naive) or too fast (unstable). 
In our hyperparameter testing, we found that the suggested learning rate is typically around $\learningrate=10^{-4}$ (see \figref{fig:lr-finding}). 
This timescale tends to align well with the varying feature sizes in the optimization landscape, e.g., peak and valley widths, around the initialization.

To explore different learning regimes, we selected three learning rates: 
one that may be too slow ($\learningrate=10^{-5}$), 
one that is close to optimal ($\learningrate=10^{-4}$), 
and one that may be too fast ($\learningrate=10^{-3}$), according to the heuristic argument.

\subsubsection{Loss Function}\label{sect:loss-function}

The learning process is guided by an evaluation of the mean-squared error loss, denoted $\msdloss \left( \batchinput,\batchoutput \right)$, which is a feature defined externally to the neural network state itself. The function measures the difference between the training batch inputs, $\batchinput$, and the outputs, $\batchoutput$. This loss quantifies how effectively the model reconstructs the absolute values of the inputs and represents only one of many learnable properties of the data. 

The evolution of a loss function depicts learning dynamics in a basic way. A decrease in the loss indicates that the model is learning the specific property better, whereas an increase suggests that the model may be `unlearning' or `forgetting' that property (with respect to the samples over which the loss is averaged).

In \sectref{sect:results}, we will explore various losses related to different observables defined on the input space, which typically correspond to physical concepts of the Ising model.

\subsubsection{Hyper-parameter Combinations}\label{sect:appendix-hyper-parameters}

\begin{table}[ht]
    \centering
    \begin{tabular}{llc}
        \hline
        \textbf{Category} & \textbf{Hyper-parameters} & \textbf{Combi.} \\
        \hline
        Constants
        & \begin{tabular}[t]{@{}l@{}}Number of steps: 100,000\\Weight decay: 0.0\\Dataset size: 300,000\end{tabular}
        & $1$\\
        \hline
        Model classes
        & \begin{tabular}[t]{@{}l@{}}Model depths: 1, 4, 16\\Model widths: 256, 512, 1024\\Bottleneck sizes: 1, 8, 64\end{tabular}
        & $27$\\
        \hline
        Statistics
        & \begin{tabular}[t]{@{}l@{}}Initialization seeds: 0, 1\\Training dataloader seeds: 0, 1\end{tabular}
        & $4$\\
        \hline
        Training
        & \begin{tabular}[t]{@{}l@{}}Batch size: 30, 300, 3000\\Data split: critical\\Learning rates: $10^{-5}$, $10^{-4}$, $10^{-3}$\end{tabular}
        & $9$\\
        \hline
    \end{tabular}
    \caption{
    Hyperparameter search space summary of all run variations. The product space of all possible hyperparameter combinations $27\times4\times9 = 972$ defines the total number of run variations that were used in this work.
    }
    \label{tab:hparam-summary}
\end{table}

\begin{table*}[ht]
    \centering
    \begin{tabular}{l|ccccc}
    \hline
         Shorthand & width & depth & bottleneck & learning rate & batch size $\batchsize$, $\modelseed$, $\dataloaderseed$\\
    \hline
         T & 512 & 4 & 1 & $10^{-4}$ & $\batchsize=300$, $\modelseed=0$, $\dataloaderseed=0$\\
    \hline
         MB1 & 512 & 4 & 1 & $10^{-4}$ & see run T\\
         MB8 & 512 & 4 & 8 & $10^{-4}$ & see run T\\
         MB64 & 512 & 4 & 64 & $10^{-4}$ & see run T\\
    \hline
         FB1D16 & 512 & 16 & 1 & $10^{-4}$ & see run T\\
         FB8D16 & 512 & 16 & 8 & $10^{-4}$ & see run T\\
         FB64D16 & 512 & 16 & 64 & $10^{-4}$ & see run T\\
    \hline
         SB8DXWX & \{256,512,1024\} & \{1,4,16\} & 8 & $10^{-5}$ & contains all variations\\
         MB8DXWX & \{256,512,1024\} & \{1,4,16\} & 8 & $10^{-4}$ & contains all variations\\
         SB64DXWX & \{256,512,1024\} & \{1,4,16\} & 64 & $10^{-3}$ & contains all variations\\
         MB64DXWX & \{256,512,1024\} & \{1,4,16\} & 64 & $10^{-4}$ & contains all variations\\
         FB64DXWX & \{256,512,1024\} & \{1,4,16\} & 64 & $10^{-5}$ & contains all variations\\
    \end{tabular}
    \caption{Shorthand names and corresponding hyper-parameters for all runs that were shown individually in one of the figures. These differentiate between slow (S), medium (M) and fast (F) learning rates, and provide the bottleneck size (B), depth (D) and width (W) whenever it differentiates from the typical run. }
    \label{tab:run-names}
\end{table*}

\clearpage
\subsection{Optimization Loss Evolution}\label{sect:appendix-raw-data-loss}

Plots depicting the evolution of the mean squared deviation between the raw inputs and outputs of the model. 
The same function was used when computing the gradients during optimization.

\begin{figure*}[ht]
    \centering
    \includegraphics[width=0.9\linewidth]{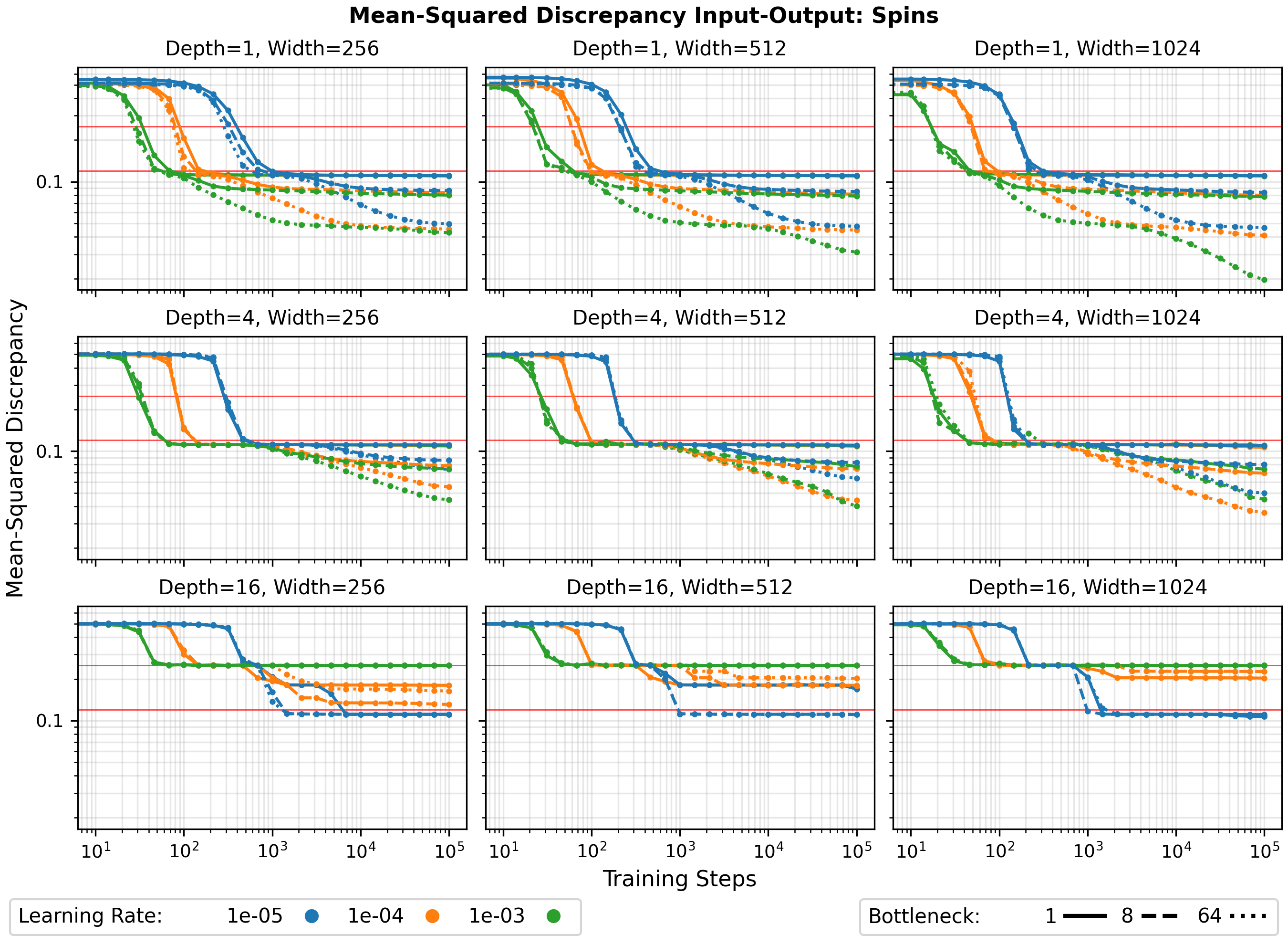}
    \caption{
    Training loss evolution of raw data over training checkpoints averaged over all runs with the same model (depth, width, bottleneck) and learning rate. 
    Each subplot shows the average across different seeds ($\modelseed \in \{0,1\}, \dataloaderseed \in \{0,1\}$) and batch sizes $\batchsize \in \{30,300,3000\}$ for a given model width and depth. 
    The line styles correspond to different bottleneck sizes, while the line colors indicate the base learning rate used for training. 
    All loss curves show a rapid initial decrease, followed by none or another rapid or slow decrease depending on the hyperparameters. 
    The MSD plateau values are equivalent to the loss obtained by predicting the training data average ($\msdloss \approx 0.25$) and corresponding sample averages ($\msdloss \approx 0.1$), respectively, indicated by the red horizontal lines. 
    The average training loss dynamics look almost identical compared to the average validation loss shown in \figref{fig:raw-msd}. 
    The individual loss curves used to compute the averages are shown in the appendix (\figref{fig:raw-msd-per-run-training}).
    }
    \label{fig:raw-msd-training}
\end{figure*}

\begin{figure*}[ht]
    \centering
    \includegraphics[width=\linewidth]{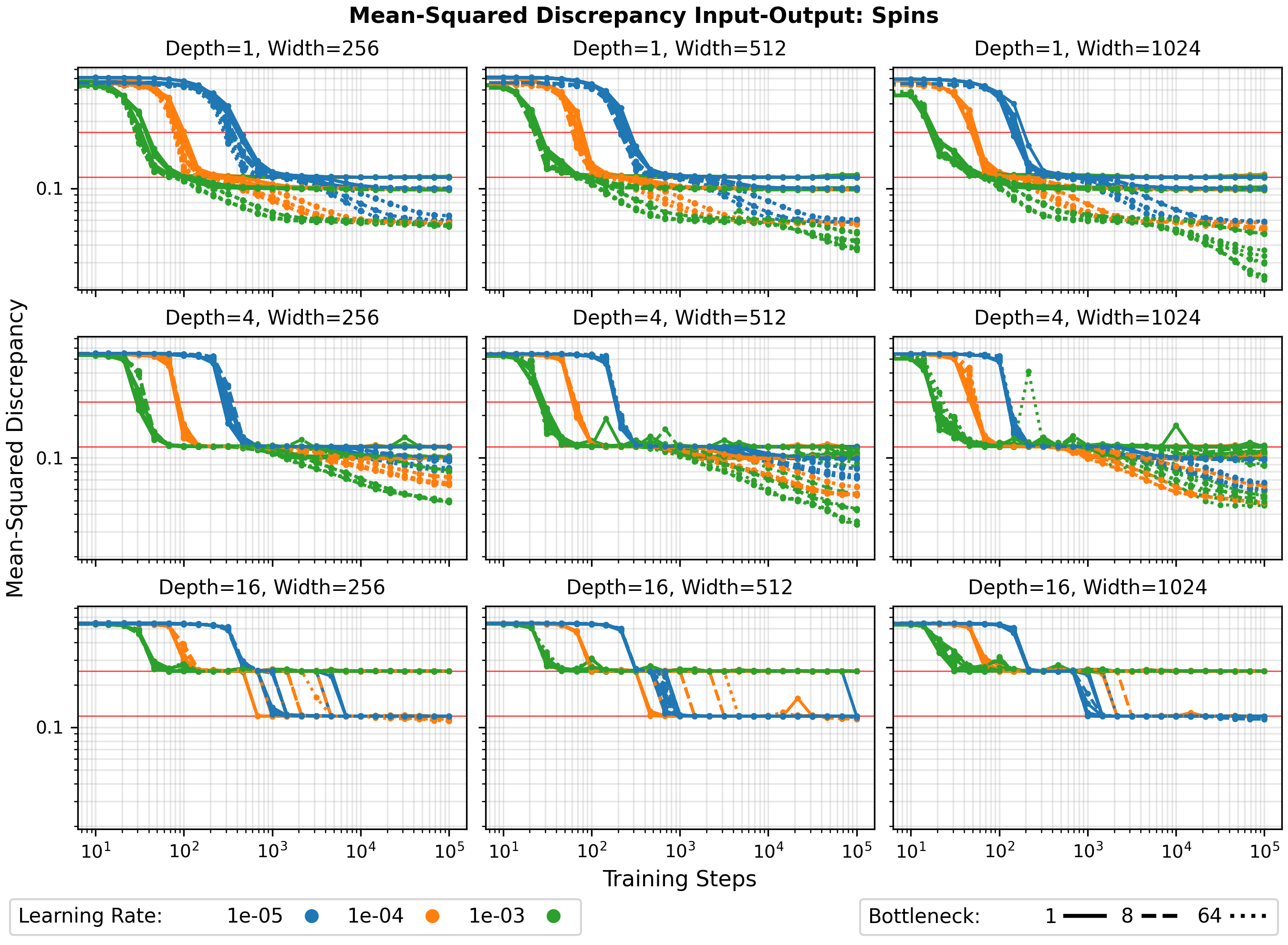}
    \caption{
    Validation loss evolution over training checkpoints for every run (see \tabref{tab:hparam-summary}). 
    Each subplot shows runs for a specific model width and depth, while the color of lines indicates the learning rate, and their style indicates the bottleneck size. 
    Lines with the same color and style correspond to different batch sizes or seeds used for data ordering and model initialization. 
    All loss curves show a rapid initial decrease, followed by none or another rapid or slow decrease depending on the hyperparameters. 
    The MSD plateau values are equivalent to the loss obtained by predicting the training data average ($\msdloss \approx 0.25$) and corresponding sample averages ($\msdloss \approx 0.1$), respectively, indicated by the red horizontal lines. 
    In particular, at fast rates, training does not always result in stable learning dynamics, which can produce outliers that shift the average significantly. 
    One example is the peak around step 2000 for $\depth=4, \width=1024$ (green dotted line), which shows up in the average across all scales in \figref{fig:spin-pooling-lr3-bn64}. 
    The slow and medium rates are mostly stable except for the deepest $\depth=16$ models. 
    In this deep regime, the transient time at an intermediate plateau shows a lot of variation between runs with mostly identical hyperparameters. 
    The average across different seeds ($\modelseed \in \{0,1\}, \dataloaderseed \in \{0,1\}$) and batch sizes $\batchsize \in \{30,300,3000\}$ for each subplot (depth, width) are shown in \figref{fig:raw-msd}.
    }
    \label{fig:raw-msd-per-run}
\end{figure*}

\begin{figure*}[ht]
    \centering
    \includegraphics[width=\linewidth]{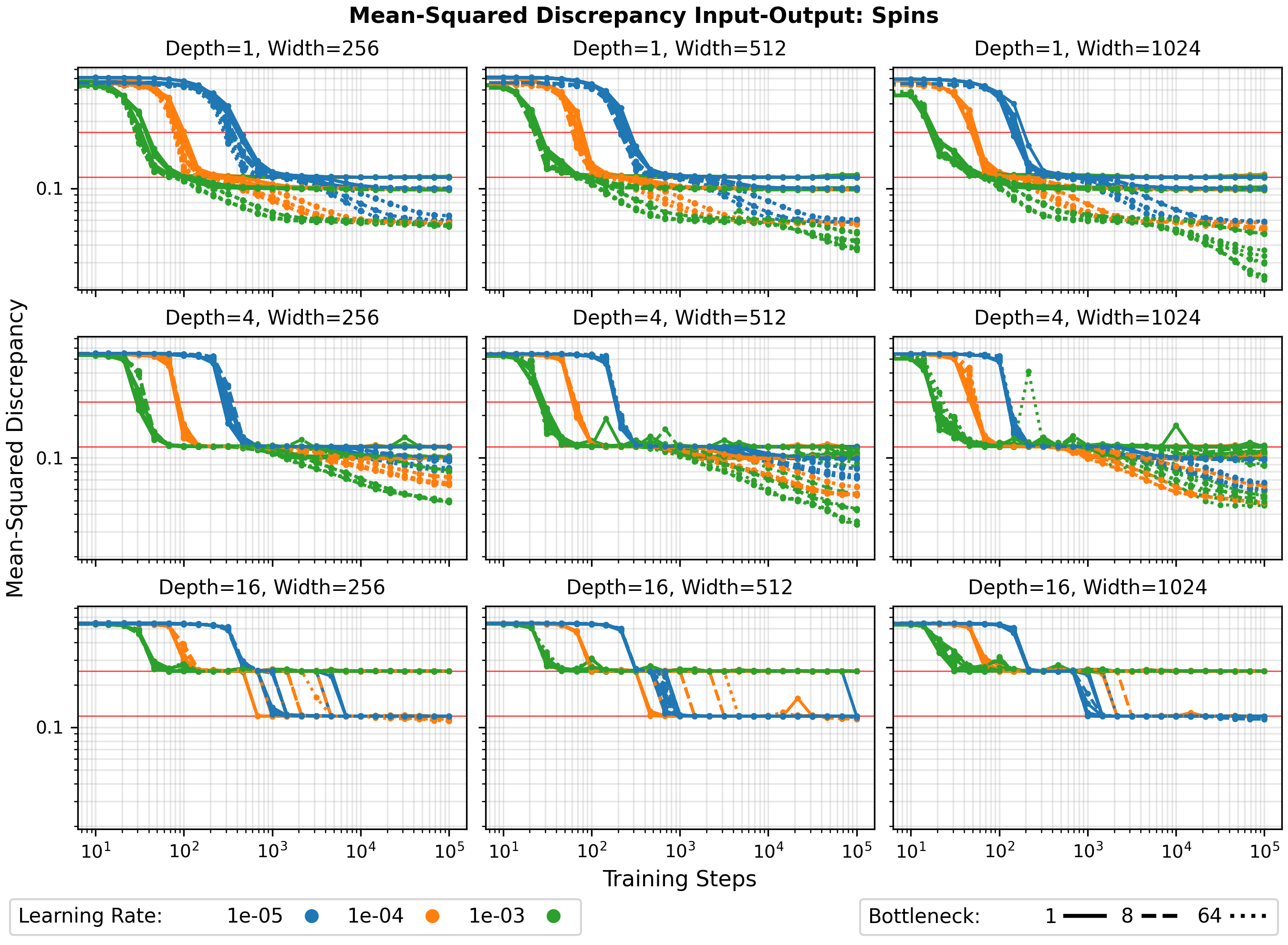}
    \caption{
    Training loss evolution over training checkpoints for every run (see \tabref{tab:hparam-summary}). 
    Each subplot shows runs for a specific model width and depth, while the color of lines indicates the learning rate, and their style indicates the bottleneck size. 
    Lines with the same color and style correspond to different batch sizes or seeds used for data ordering and model initialization. 
    All loss curves show a rapid initial decrease, followed by none or another rapid or slow decrease depending on the hyperparameters. 
    The MSD plateau values are equivalent to the loss obtained by predicting the training data average ($\msdloss \approx 0.25$) and corresponding sample averages ($\msdloss \approx 0.1$), respectively, indicated by the red horizontal lines. 
    In particular, for fast rates, training does not always result in stable learning dynamics, which can produce outliers shifting the average significantly. 
    One example is the peak around step 2000 for $\depth=4, \width=1024$ (green dotted line), which shows up in the average across all scales in \figref{fig:spin-pooling-lr3-bn64}. 
    The slow and medium rates are mostly stable except for the deepest $\depth=16$ models. 
    In this deep regime, the transient time at an intermediate plateau shows a lot of variation between runs with mostly identical hyperparameters. 
    The training loss dynamics look almost identical compared to the validation loss in \figref{fig:raw-msd-per-run} with the exception of the fastest learning rate often having lower training losses. 
    The average across different seeds ($\modelseed \in \{0,1\}, \dataloaderseed \in \{0,1\}$) and batch sizes $\batchsize \in \{30,300,3000\}$ for each subplot (depth, width) are shown in \figref{fig:raw-msd-training}.
    }
    \label{fig:raw-msd-per-run-training}
\end{figure*}

The averages across runs in \figref{fig:raw-msd} and \figref{fig:raw-msd-training} don't reveal the fluctuations between individual runs. 
Therefore, \figref{fig:raw-msd-per-run} (validation) and \figref{fig:raw-msd-per-run-training} (training) show the MSD of individual runs.

\clearpage
\subsection{Sample Reconstruction Evolution}\label{sect:appendix-reconstruction-evolution}

\clearpage
\subsection{Observable Loss Evolution}\label{sect:appendix-obs-losses-evo}

MSD evolutions of different observables split by the temperatures used to generate the dataset or the hyperparameters of the runs.

\subsubsection{Example of non-monotonic loss scaling}\label{sect:appendix-msd-scaling-example}

In the following, we provide a counterexample to prove that losses are not necessarily decreasing for larger averaging scales.

Let's assume a periodic error signal with period 3, resulting in $\kernelloss{1}=\tfrac{1}{2}$:
\begin{align*}
    &\overline{e} = e = [1,\,-\tfrac{1}{2},\,-\tfrac{1}{2},\,1,\,-\tfrac{1}{2},\,-\tfrac{1}{2},\dots]\\
    &\rightarrow \kernelloss{1}=\tfrac{1}{3}(1^2+(-\tfrac{1}{2})^2+(-\tfrac{1}{2})^2)=\tfrac{1}{2}
\end{align*}
Every kernel window of size 3 would average the signal to zero $
\overline{e}=[0, 0, 0, \dots] \rightarrow \kernelloss{3}=0
$, while a kernel window of size 4, averaging over a larger area, results in $
\overline{e}=[\tfrac{1}{4},-\tfrac{1}{8},-\tfrac{1}{8}, \dots] \rightarrow \kernelloss{4}=\frac{1}{32}
$. 
In this example the loss ($\kernelloss{1,3,4}=\tfrac{1}{2},0,\frac{1}{32}$) is a non-monotonic function over the kernel sizes ($\kernelsize=1,3,4$). 
To summarize, while $\kernelloss{a} \leq \kernelloss{1}$ is guaranteed for $a>1$ (i.e. $0 \leq \tfrac{1}{2}$ and $\frac{1}{32} \leq \tfrac{1}{2}$), we can't generally expect  $\kernelloss{a} \leq \kernelloss{b}$ for any $a>b$ (i.e. $\tfrac{1}{2} \nleq \frac{1}{32}$). 

\subsubsection{Magnetization}\label{sect:appendix-magnetization-loss}

\begin{figure*}[ht]
    \centering
    \includegraphics[width=1\linewidth]{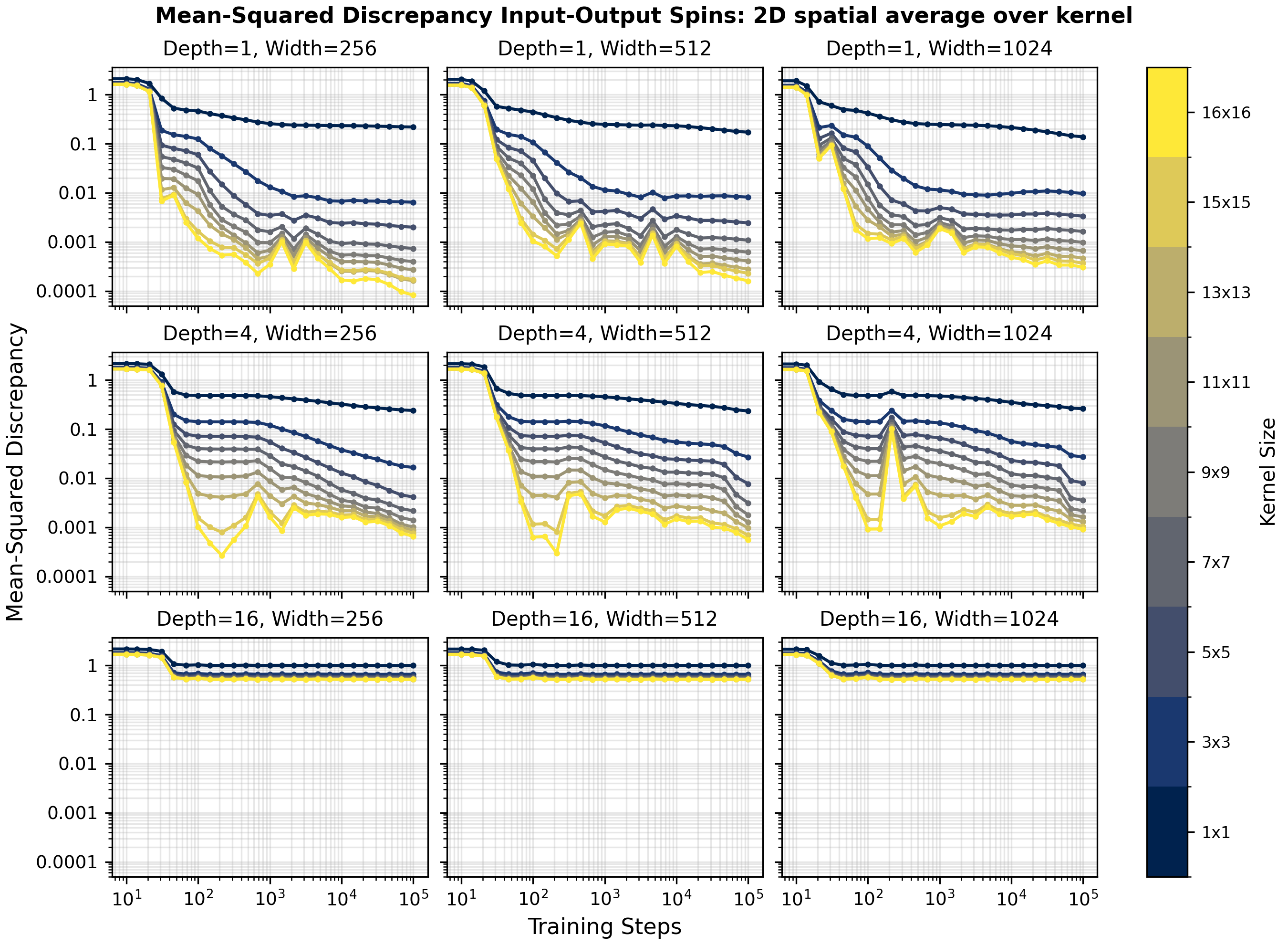}
    \caption{
    Validation loss evolution of coarse-grained averages of local lattice site spins using different kernel sizes, plotted across training checkpoints. 
    The averages include all \textbf{FB64DXWX runs} (see \tabref{tab:run-names}) with a learning rate of $\learningrate=10^{-3}$ and a bottleneck size of $\bottleneck=64$. 
    Each subplot shows averages across runs with a specific model width and depth, with line color indicating the kernel size used for averaging. 
    The validation loss decreases as the kernel size increases, despite the lack of a guarantee that the different averages have a particular order. 
    Large-scale losses show increasing fluctuations in a regime where these losses are slightly higher for depth $\depth=1,4$ models. 
    The average for $\depth=4,\width=1025$ models shows a peak at which all scales become worse by orders of magnitude. 
    This behavior can be traced back to an outlier run (dotted green line in \figref{fig:raw-msd-per-run} showing a temporarily high loss. 
    The deepest models $\depth=16$ show little changes after a small initial transition between steps 10-50.
    }
    \label{fig:spin-pooling-lr3-bn64}
\end{figure*}

\begin{figure*}[ht]
    \centering
    \includegraphics[width=1\linewidth]{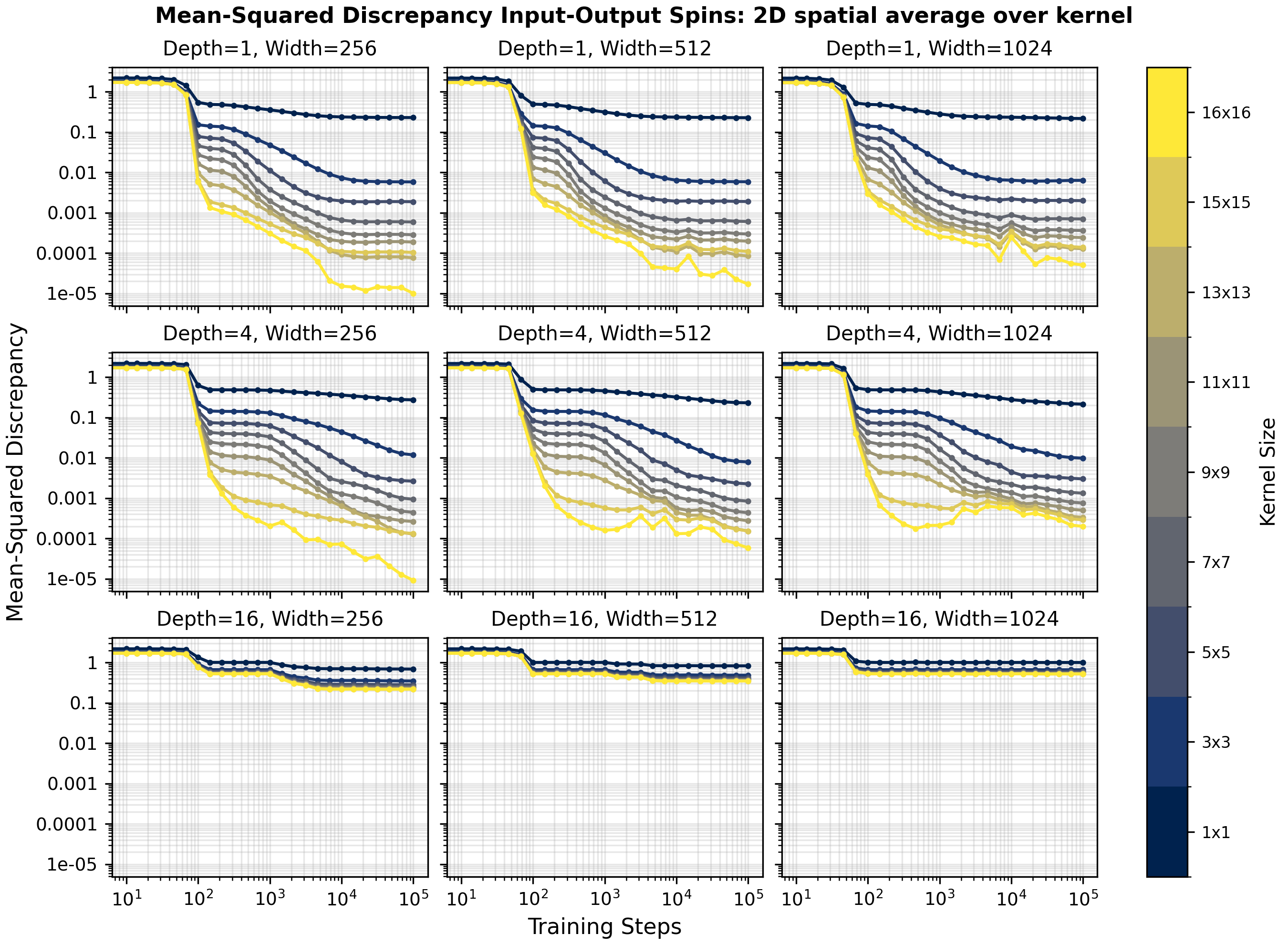}
    \caption{
    Validation loss evolution of coarse-grained averages of local lattice site spins using different kernel sizes, plotted across training checkpoints. 
    The averages include all \textbf{MB64DXWX runs} (see \tabref{tab:run-names}) with a learning rate of $\learningrate=10^{-4}$ and a bottleneck size of $\bottleneck=64$. 
    Each subplot shows averages across runs with a specific model width and depth, with line color indicating the kernel size used for averaging. 
    The validation loss decreases as the kernel size increases, despite the lack of a guarantee that the different averages have a particular order. 
    Large-scale losses only have a temporary minimum for some depth $\depth=4$ models (in contrast to \figref{fig:spin-pooling}).
    }
    \label{fig:spin-pooling-lr4-bn64}
\end{figure*}

\begin{figure*}[ht]
    \centering
    \includegraphics[width=1\linewidth]{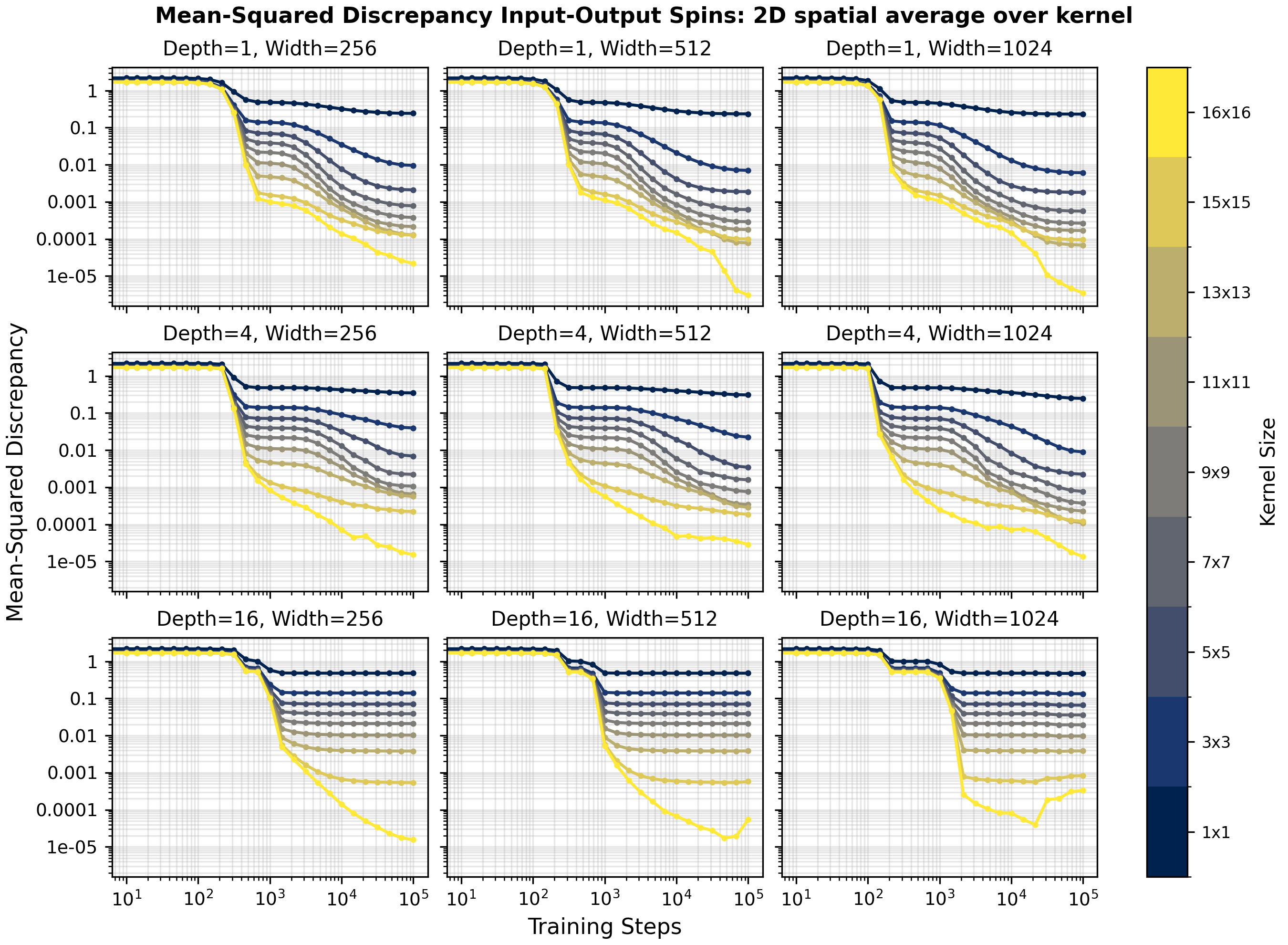}
    \caption{
    Validation loss evolution of coarse-grained averages of local lattice site spins using different kernel sizes, plotted across training checkpoints. 
    The averages include all \textbf{SB64DXWX runs} (see \tabref{tab:run-names}) with a learning rate of $\learningrate=10^{-5}$ and a bottleneck size of $\bottleneck=8$. 
    Each subplot shows averages across runs with a specific model width and depth, with line color indicating the kernel size used for averaging. 
    For most configurations, the validation loss decreases as the kernel size increases, despite the lack of a guarantee that the different averages have a particular order. 
    The shallower and wider models represent an exception, where the 13x13 kernel losses cross with neighboring scales.
    }
    \label{fig:spin-pooling-lr5-bn64}
\end{figure*}

Exemplary loss curves showing the MSD between input and output magnetization for various learning rates are shown in \figref{fig:spin-pooling-lr3-bn64}, \ref{fig:spin-pooling-lr4-bn64}, and \ref{fig:spin-pooling-lr5-bn64}. 

\subsubsection{Energies}\label{sect:energy-pooling}

In contrast to magnetization, which is an order parameter and a mean variable or ``field" without an inherent scale, the coarse-graining scale is a relevant variable for the energy function/operator, which is defined through local gradients in the configurations, i.e., spin-pair operations. 

Given the local spins defined as $\sitespin{i}=2\rawpixel{i}-1$ for the raw data $\rawinput,\rawoutput$, the energy per site $i$ can be calculated using the formula:
$$
\energy_{i} = \sum_{j \in \siteneighborhood{i}} |\sitespin{i} - \sitespin{j}| - 1,
$$
where $j$ represents neighboring sites in the neighborhood $\siteneighborhood{i}$ of site $i$. 

We utilize a distance-based energy function, which is equivalent to the theoretical product-based energy for valid spin states $\sitespin{i} \in \{-1, 1\}$. 
This translation-invariant energy facilitates a clear separation of the reconstruction analysis into absolute errors measured with scale-dependent magnetizations $\scalemagnetization$ (covered in the previous section \sectref{sect:spin-pooling}), and relative errors in the form of isotropic gradients measured by local energies $\siteenergy{i}$, discussed in the following.
With this perspective in mind, we can identify the local magnetization, or its absolute value, with the zeroth-order term, and the energy, or average local gradient, with one of the four first-order components of a Taylor expansion around each lattice site. 
While the first-order expansion can also be described by individual gradients in each direction, the collective modes provide a better basis due to the energy-based coupling.
This background provides additional motivation for using physics-inspired observables to analyze learning dynamics.

\begin{figure*}
    \centering
    \includegraphics[width=1\linewidth]{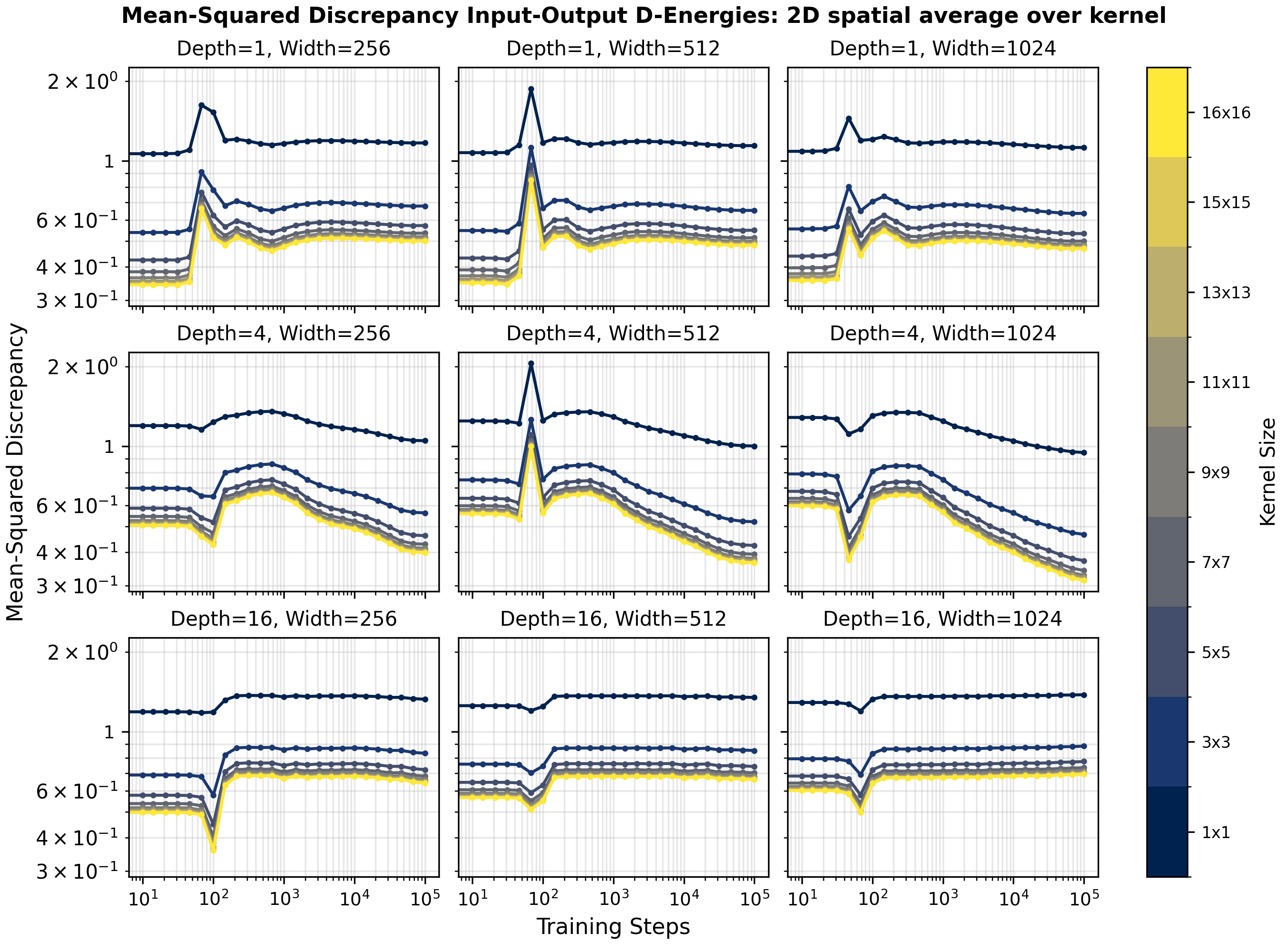}
    \caption{
    Validation loss evolution of coarse-grained averages of local lattice site energies using different kernel sizes over training checkpoints. 
    The averages include all \textbf{MB8DXWX runs} (see \tabref{tab:run-names}) with a learning rate of $\learningrate=10^{-4}$ and a bottleneck size of $\bottleneck=8$. 
    Each subplot shows run averages across specific model width and depth, with line color indicating the kernel size used for averaging. 
    The losses for different kernel sizes decrease with scale, as observed in \figref{fig:spin-pooling}. 
    The final average loss of energies across different scales is larger than the initial loss for most models, except at intermediate depth ($\depth=4$). 
    A large deviation (peak or trough) is visible for all model widths and depths between steps 40-100, concurrent with the first drop in the loss for the raw data (orange line in \figref{fig:raw-msd}) and the coarse-grained averages (\figref{fig:spin-pooling}).
    Note that in the deep learning regime, the curves remain nearly constant. 
    This indicates that the models do not learn to reconstruct the energy. 
    }
    \label{fig:energy-pooling}
\end{figure*}

In \figref{fig:energy-pooling}, the average validation loss for coarse-grained energies decreases as the kernel size increases across all averaged runs, similar to what is observed for spin averages in \figref{fig:spin-pooling}. 
This indicates that the energy error field has fluctuations at multiple scales, reflecting the resolution of gradients at those scales. 
However, the resulting differences between larger scales are minimal, suggesting that the error field is primarily determined by the inability to reconstruct small-scale fluctuations.
While low-frequency modes may be properly reconstructed, high-frequency details may not, such as small-scale domain boundaries, as observed for individual samples in \figref{fig:high-loss-outputs}. 
Therefore, we expect the energy to be decoupled from the optimization objective, while the global average is learned (inconsistent peaks or troughs). 
Only when the smaller scales become more relevant is the energy increasingly coupled (decreasing across scales).

In contrast to the small-scale absolute losses in \figref{fig:spin-pooling}, not a single loss on any of the scales in \figref{fig:energy-pooling} consistently decreases over training. 
Notably, the losses even increase compared to the randomly initialized model in most cases. 
Additionally, the loss curves of all model widths and depths show large deviations (peaks or troughs) between steps 40-100, concurrent with the first drop in the loss for the raw data (orange line in \figref{fig:raw-msd}) and the coarse-grained averages (\figref{fig:spin-pooling}).

The output evolution for individual samples, as shown in \figref{fig:high-loss-outputs}, indicates that the initial decrease of the loss is accompanied by strong fluctuations across pixels, especially in smaller models. 
As training progresses, the energy loss reduces across all scales as the trade-off between global and local averages emerges, marked by temporary minima in large-scale losses, as illustrated in \figref{fig:spin-pooling}. 
From then on, the energy loss decreases consistently as the model begins fitting local averages, helping resolve spin-domain boundaries and, in turn, the relative differences that contribute to the energy. 
A detailed examination of the joint output distribution of magnetization and energy, presented in \sectref{sect:output-distribution-dynamics}, shows that low-energy configurations, which contain larger domains or clusters, are learned before high-energy configurations that require smaller-scale feature resolution. 
We conclude that the energy is not linked to the primary optimization objective during periods of high loss, where errors in the global averages are dominant. 
The scale-bridging energy learning regime only develops when the spin-error averages become scale-dependent, with only small scales improving. 
These small-scale losses become low enough to necessitate resolving spin-domain boundaries at the dominant scales. 

\clearpage
\subsubsection{Principal Components}\label{sect:appendix-pca-loss}

All loss curves showing the MSD between input and output principal components.

\begin{figure*}[ht]
    \centering
    \includegraphics[width=1\linewidth]{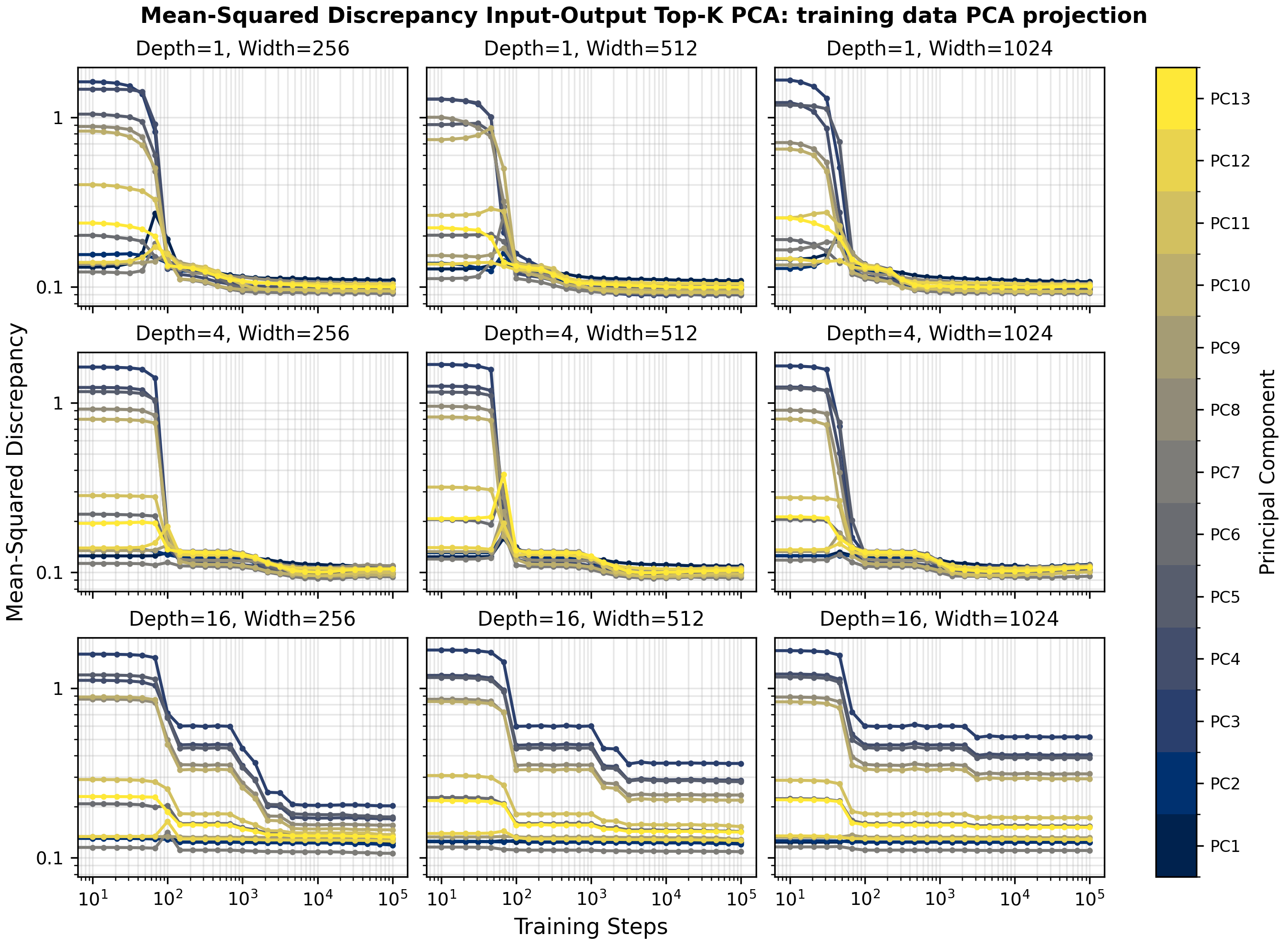}
    \caption{
    Average validation loss evolution over training checkpoints for all \textbf{MB8DXWX runs} (see \tabref{tab:run-names}) with a learning rate of $\learningrate=10^{-4}$ and a bottleneck size of $\bottleneck=8$. 
    The validation error is decomposed into different contributions by projecting the reconstruction error onto the top 13 principal components of the training data. 
    The higher the principal component number (PC1 vs PC13), the less variance inside the training data is explained by the corresponding direction. 
    The principal components are not learned in order of decreasing explained variance but rather appear to be learned mostly simultaneously. 
    Some components show a peak around step 100, concurrent with the peak observable in the energy loss (\figref{fig:energy-pooling}). 
    However, the peak is only noticeable for some principal components, while the peak is visible across scales of the energy loss. 
    The principal components' explained variance is mainly defined by increasingly smaller-scale variations. 
    Components with similar variance appear in groups of e.g. four, with the group size corresponding to the number of independent 2D modes (with a different spatial phase distribution). 
    The similar dynamics across scales initially visible in the magnetization (\figref{fig:spin-pooling}) and generally for energy (\figref{fig:energy-pooling}) might explain the synchronous learning of different PCA components and the peak observed for some components. 
    However, it appears that the PCA components are more selectively showing the peak, suggesting that additional investigations are required to fully understand the behavior. 
    }
    \label{fig:topkpca-lr4-bn8}
\end{figure*}

\clearpage
\subsection{Sample Observables Evolution}\label{sect:appendix-output-observable-evo}

The training and validation data samples are taken as inputs. For each input and output, the corresponding observable is computed to obtain the following plots.

\subsubsection{Magnetization}

\begin{figure*}[ht]
    \centering
    \includegraphics[width=\linewidth]{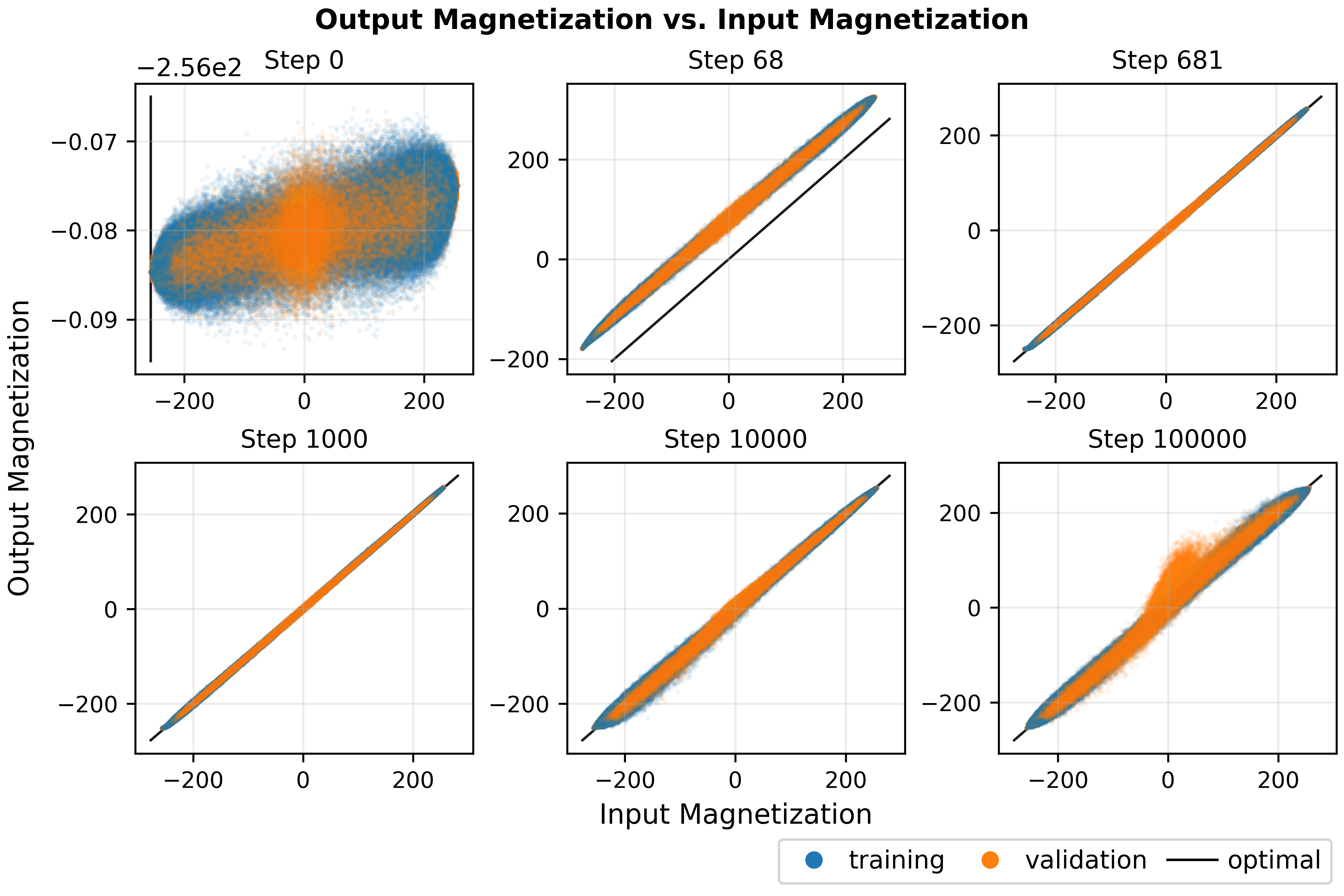}
    \caption{
    Input vs. output magnetization of all training (blue) and validation (orange) samples for different model checkpoints of \textbf{run MB8} (see \tabref{tab:run-names}) during training. 
    Initially, almost all outputs have the same magnetization value $\magnetization=-256$. 
    After the first 68 steps, the distribution resembles a one-dimensional line that is shifted to higher output magnetization compared to the optimal relationship. 
    From step 681-1000, the magnetization is almost perfectly reconstructed in the output. 
    At step $10^4$, marking the end of the topological (folding) transition in \figref{fig:io-energy}, the variance is larger than before. 
    This period of magnetization `unlearning' is also visible in the scale-dependent losses in \figref{fig:spin-pooling}. 
    At the end of training (step 100 000), the magnetization for some validation samples with almost vanishing magnetization is overestimated in the output. 
    These samples must have high energies and thus spatial fluctuations on small length scales. 
    As a result, these samples don't benefit as much from global averages compared to small-scale averages. This can explain why the magnetization reconstruction becomes worse in the scale-bridging energy learning regime. 
    }
    \label{fig:io-magnetization}
\end{figure*}

\subsubsection{Energy}

\begin{figure*}
    \centering
    \includegraphics[width=\linewidth]{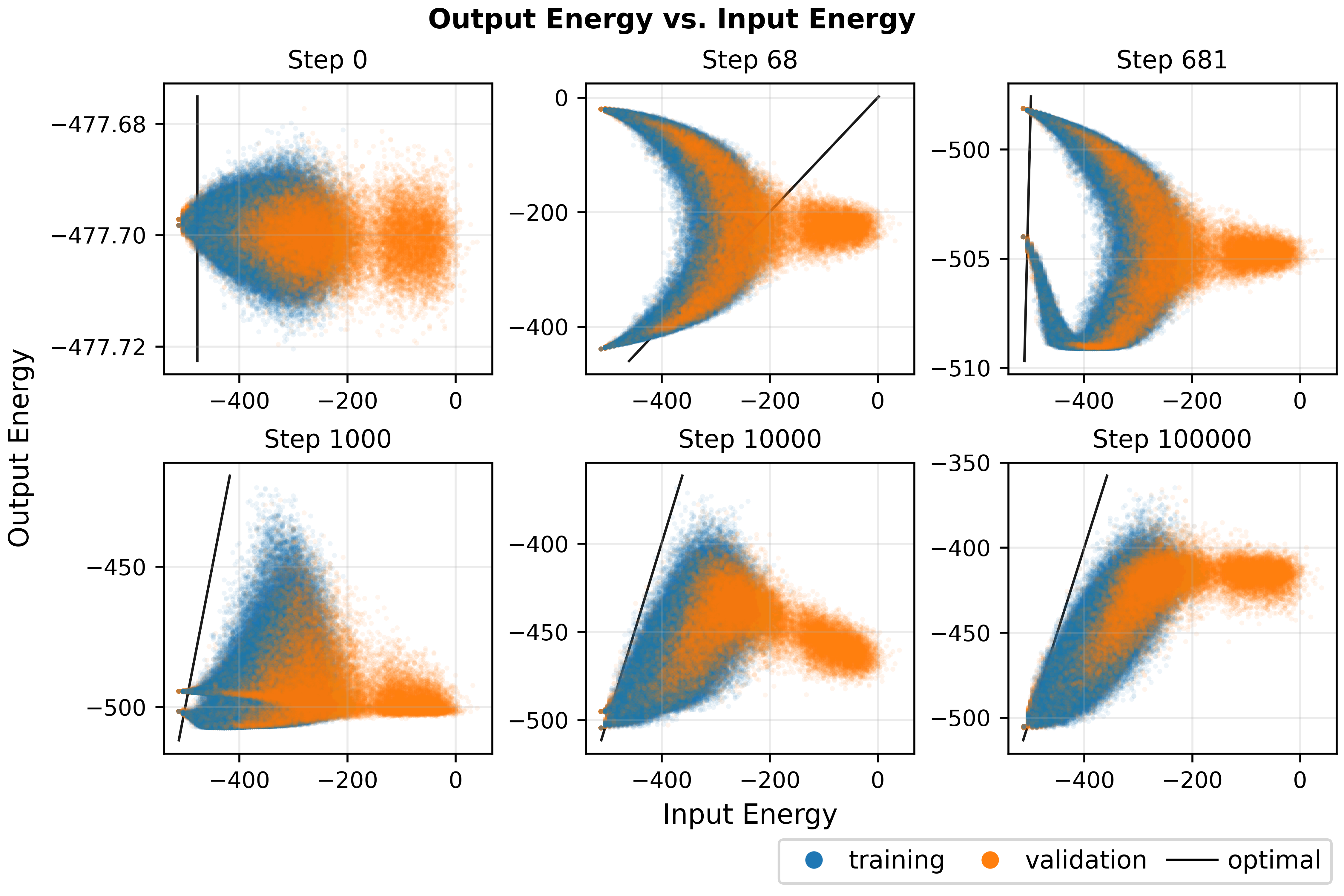}
    \caption{
    Input vs. output energy of all training (blue) and validation (orange) samples for different model checkpoints of \textbf{run MB8} (see \tabref{tab:run-names}) during training. 
    Initially, almost all outputs have the same energy value. 
    After the first 68 steps, the distribution resembles the joint distribution of energy and magnetization, as the output energy is correlated with magnetization (see \figref{fig:mag-vs-energy}). 
    At step 681, the ends of low energies are close to the optimum line, while the rest of the distribution appears to fold. 
    At step $10^4$, the distribution resembles a folded version of this joint distribution, indicating the end of a topological transition. 
    At later stages, the distribution of the training data converges towards a straight line with decreasing variance. 
    Yet the slope is underestimated, which suggests a systemic factor limiting high-energy resolution. 
    We note that the slope of the training distribution approaches the optimal value as the bottleneck size increases (\figref{fig:io-energy-b64}). 
    For both bottleneck sizes, the validation data distribution at high energies doesn't converge to a straight line, underscoring that the output energies are underestimated above the critical temperature of the Ising model. 
    }
    \label{fig:io-energy}
\end{figure*}

In \figref{fig:io-energy}, we explore how the autoencoder forms an approximation for the true energy concept--by looking at the co-evolving output and input distributions of the global energy (\eqqref{eq:global-energy}). 
Notably, for the bottleneck-8 system specified, around step 68, the co-distribution closely resembles the joint distribution of ground-truth magnetization and energy that characterizes the underlying dataset shown in \figref{fig:dataset-contours}. 
At this stage, the output energy becomes almost perfectly correlated with the output magnetization (\figref{fig:mag-vs-energy}). 
Simultaneously, the magnetization concept is nearly perfectly learned (see \figref{fig:io-magnetization}), acting as an effective first-order proxy for the energy concept.
In later stages, the input vs. output energy distribution becomes folded and converges towards the optimal value indicated (solid black line in \figref{fig:mag-vs-energy}). 
For low input energies, the output energies converge to the ground truth very quickly with minimal remaining variance. 
In contrast, this is approached more slowly for high-energy configurations, which exhibit greater variance. 
The output representation rarely reaches energies above $\energy=-400$, while inputs represent energies up to $\energy=0$. This high-energy-cutoff shifts to $\energy=-250$ for the larger $\bottleneck=64$, suggesting that the resolution of small scales, necessary to represent high-energy configurations, is limited by the bottleneck. 
Notice that this behavior is expected; we observed that increasingly small details are resolved on the sample level (\figref{fig:high-loss-outputs}) while losses governing large scales are much lower across runs (on average, see \figref{fig:spin-pooling}). 

In general, models with a small bottleneck or large depth tend to exhibit less smooth changes in distribution (see the edge artifacts in the supplementary materials). 
Very deep models initially produce only a single output energy level, followed by an emerging second. 
Deep-and-thin models ultimately converge to a spread-out distribution in energy, while deep-and-wide models converge only to the single level.

\begin{figure*}[ht]
    \centering
    \includegraphics[width=\linewidth]{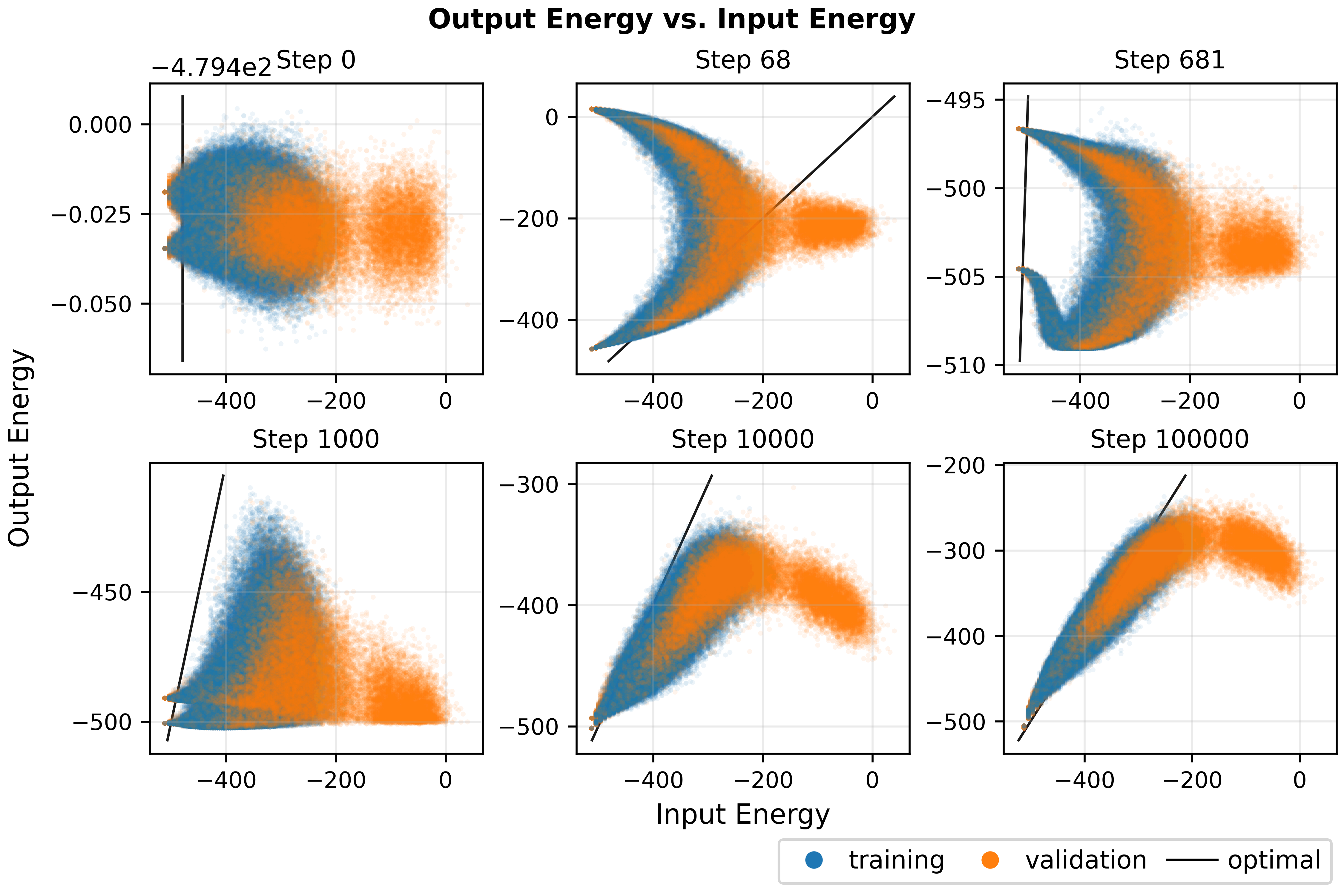}
    \caption{
    Input vs. output energy of all training (blue) and validation (orange) samples for different model checkpoints of \textbf{run MB64} (see \tabref{tab:run-names}) during training. 
    Initially, almost all outputs have the same energy value. 
    After the first 68 steps, the distribution resembles the joint distribution of energy and magnetization, with the output energy correlating with magnetization (see \figref{fig:mag-vs-energy}). 
    At step 681, the ends of low energies are close to the optimum line, while the rest of the distribution appears to fold. 
    At step $10^4$, the distribution resembles a folded version of this joint distribution, indicating the end of a topological transition. 
    At later stages, the distribution of the training data converges towards a straight line with decreasing variance. 
    Yet the slope is underestimated at high energies, suggesting a systemic factor limiting high-energy resolution. 
    The training distribution approaches the optimal value more closely compared to the smaller bottleneck (\figref{fig:io-energy}). 
    The validation data distribution at high energies doesn't converge to a straight line, underscoring that the output energies are underestimated above the Ising critical temperature. 
    With higher energies, the deviation becomes larger and even bends downwards. 
    This suggests that the limiting factor originates from the small domains, which become increasingly prevalent for higher energies.
    }
    \label{fig:io-energy-b64}
\end{figure*}

\clearpage
\subsection{Latent Flow}\label{sect:appendix-latent-flow}

The changes in the latent representation under recursion.

\subsubsection{Magnetization}

\begin{figure*}[ht]
    \centering
    \includegraphics[width=1\linewidth]{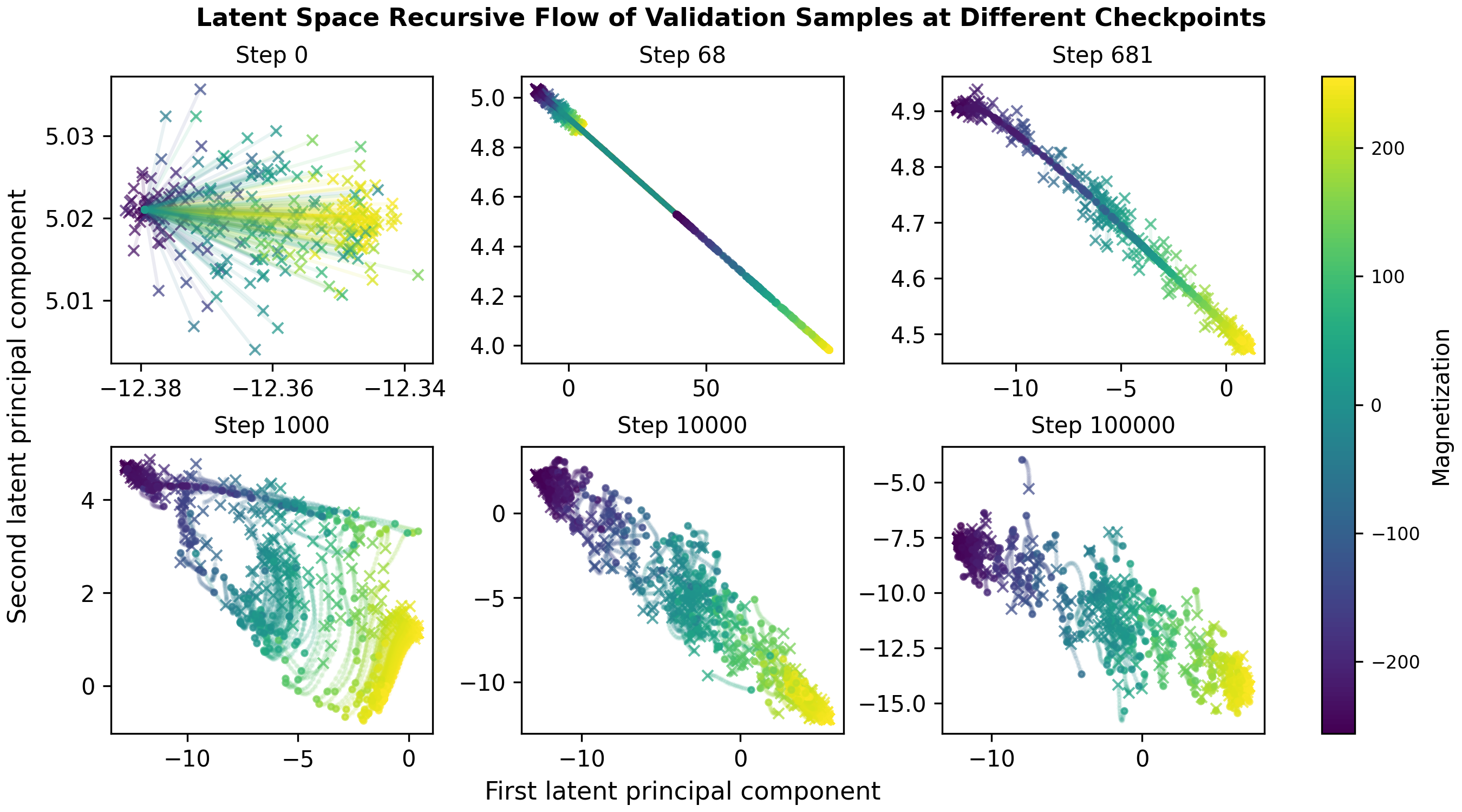}
    \caption{
    The first two principal components of the trajectory of the 8-dimensional latent representation of validation samples. 
    The trajectory is obtained by recursively applying the autoencoder model to initial samples, and is defined by the autoencoder's state at different training checkpoints of \textbf{run MB64} (see \tabref{tab:run-names}). 
    The principal components are computed once across all checkpoints. 
    The latent representation of the initial input (cross) moves with each recursion step to the next position in latent space (small circles), until the last step (big circle). 
    The trajectories are colored according to each sample's magnetization. 
    Initially, samples move towards a single point, then on a line, 2d curve, and finally they move through all visible dimensions. 
    After step 68, points with more similar magnetization seem to remain closer together. 
    At step 1000, the trajectories fan out into the second visible dimension before they appear to become more confined in the following checkpoints (step 10 000 and 100 000). 
    }
    \label{fig:latent-flow-magnetization-b64}
\end{figure*}

\clearpage
\subsubsection{Energy}

\begin{figure*}[ht]
    \centering
    \includegraphics[width=1\linewidth]{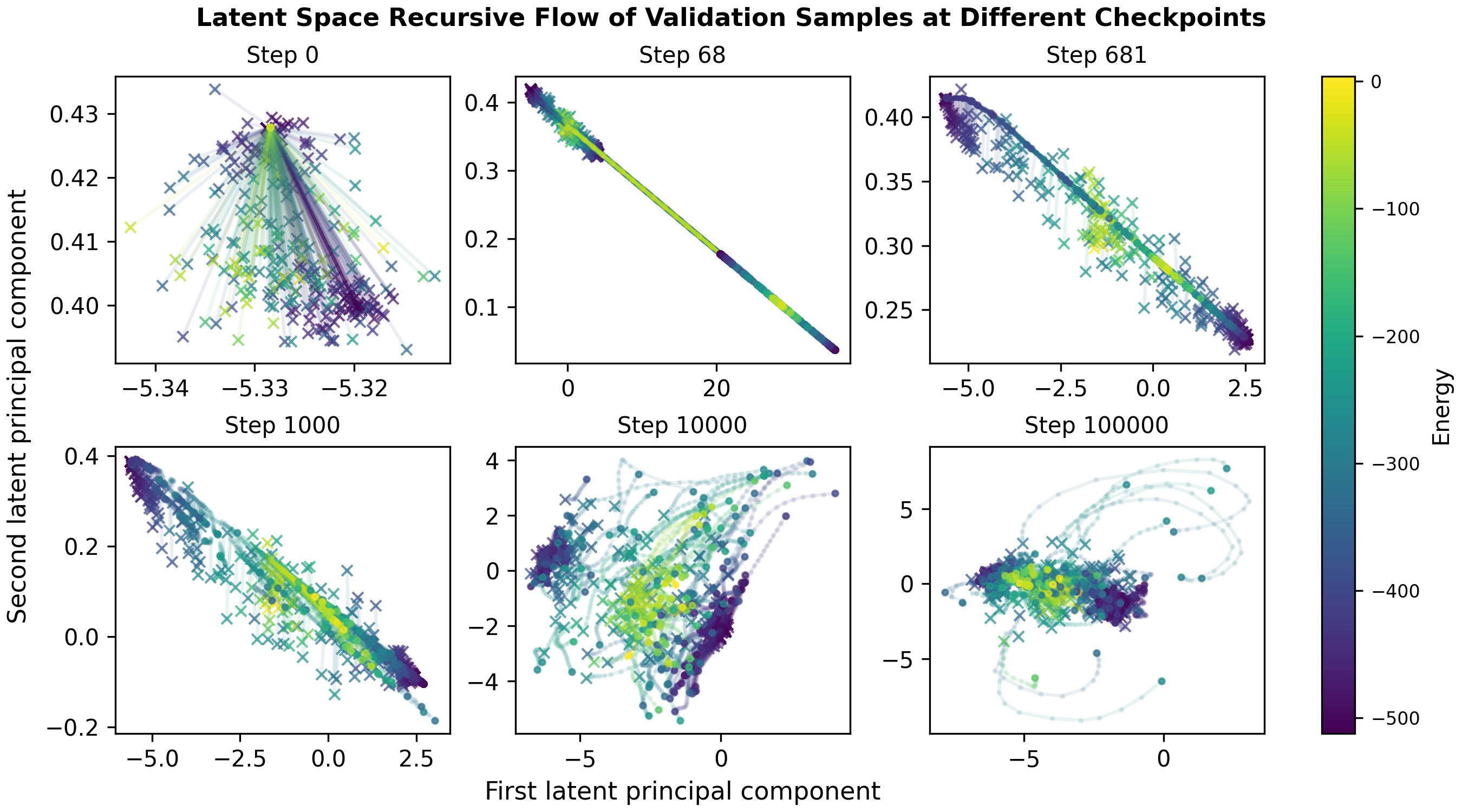}
    \caption{
    The first two principal components of the trajectory of the 8-dimensional latent representation of validation samples. 
    The trajectory is obtained by recursively applying the autoencoder model to initial samples, and is defined by the autoencoder's state at different training checkpoints of \textbf{run MB8} (see \tabref{tab:run-names}). 
    The principal components are computed once across all checkpoints. 
    The latent representation of the initial input (cross) moves with each recursion step to the next position in latent space (small circles), until the last step (big circle). 
    The trajectories are colored according to each sample's energy. 
    Initially, samples move towards a single point, then on a line, 2d curve, and finally they move through all visible dimensions. 
    After step 68, points with more similar energy seem to remain closer together. 
    However, there is no linear topological ordering, as samples with the same low energy are on opposite sites, in contrast to magnetization (\figref{fig:latent-flow-magnetization-b8}). 
    }
    \label{fig:latent-flow-energy-b8}
\end{figure*}

\begin{figure*}[ht]
    \centering
    \includegraphics[width=1\linewidth]{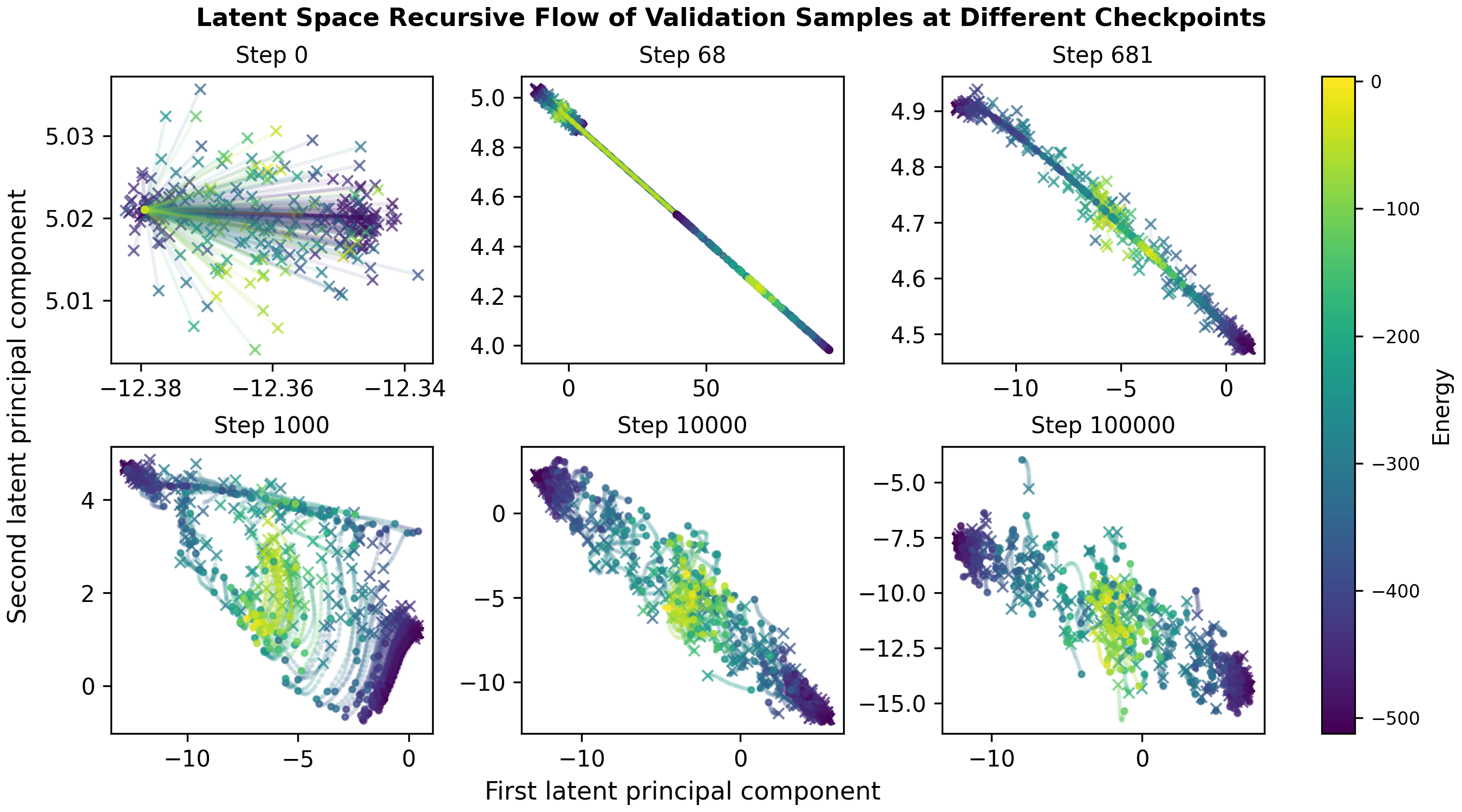}
    \caption{
    The first two principal components of the trajectory of the 8-dimensional latent representation of validation samples. 
    The trajectory is obtained by recursively applying the autoencoder model to initial samples, and is defined by the autoencoder's state at different training checkpoints of \textbf{run MB64} (see \tabref{tab:run-names}). 
    The principal components are computed once across all checkpoints. 
    The latent representation of the initial input (cross) moves with each recursion step to the next position in latent space (small circles), until the last step (big circle). 
    The trajectories are colored according to each sample's energy. 
    Initially, samples move towards a single point, then on a line, 2d curve, and finally they move through all visible dimensions. 
    After step 68, points with more similar energy seem to remain closer together. 
    However, there is no linear topological ordering, as samples with the same low energy are on opposite sites, in contrast to magnetization (\figref{fig:latent-flow-magnetization-b64}). 
    At step 1000, the trajectories fan out into the second visible dimension before they appear to become more confined in the following checkpoints (step 10 000 and 100 000). 
    }
    \label{fig:latent-flow-energy-b64}
\end{figure*}

\clearpage
\subsection{Correlation to Input Observables}\label{sect:appendix-spearman-correlation}

Relationship between observables (magnetization, energy) and the latent space representation.

\subsubsection{Small Bottleneck}

\begin{figure*}[ht]
    \centering
    \includegraphics[width=0.75\linewidth]{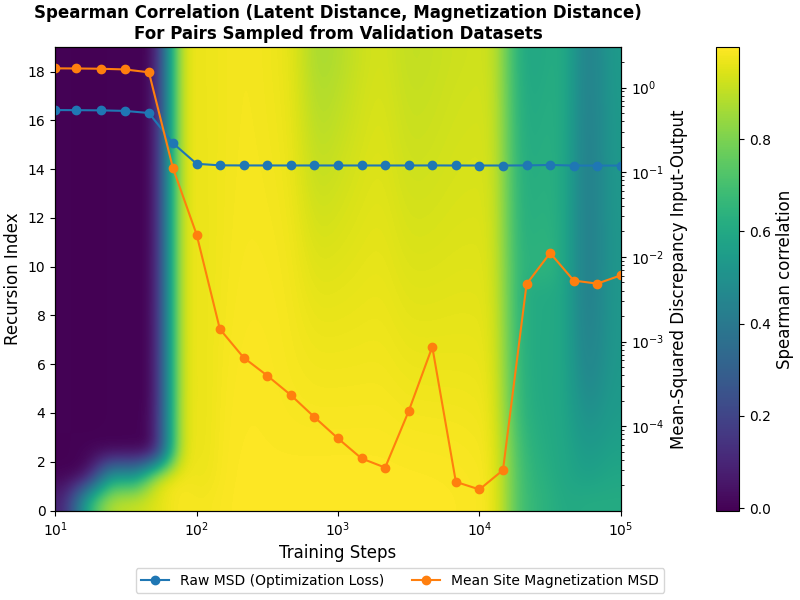}
    \caption{
    Spearman rank correlation between the input magnetization difference and latent space representation distance (L2-norm) of sampled validation data pairs. 
    The correlation is shown for different training checkpoints of \textbf{run MB1} (see \tabref{tab:run-names}) and using the latent representation for different numbers of recursions (feeding the output back into the input). 
    At the beginning of training, the correlation drops rapidly under recursion (lower left) as the encoder maps all points to the same output. 
    Around step 100, the correlation remains stable under recursion, concurrent with the drop in the optimized loss (blue) and the corresponding macroscopic magnetization loss (orange). 
    Around step 20 000, the correlation decreases independent of the number of recursions, concurrent with an increase in the magnetization loss. 
    It then remains low and stable under recursion until the end of training, with an exception between around $6\cdot10^4-8\cdot10^4$ where it decreases with recursion. 
    The topological ordering according to the global magnetization seems to be lost in the final stages, even with the one-dimensional bottleneck. 
    This might be related to overfitting to the training data (see worse generalization in \figref{fig:high-loss-outputs}), as the learning rate decays to zero in this period. 
    }
    \label{fig:spearman-latent-vs-magnetization-b1}
\end{figure*}

\begin{figure*}[ht]
    \centering
    \includegraphics[width=0.75\linewidth]{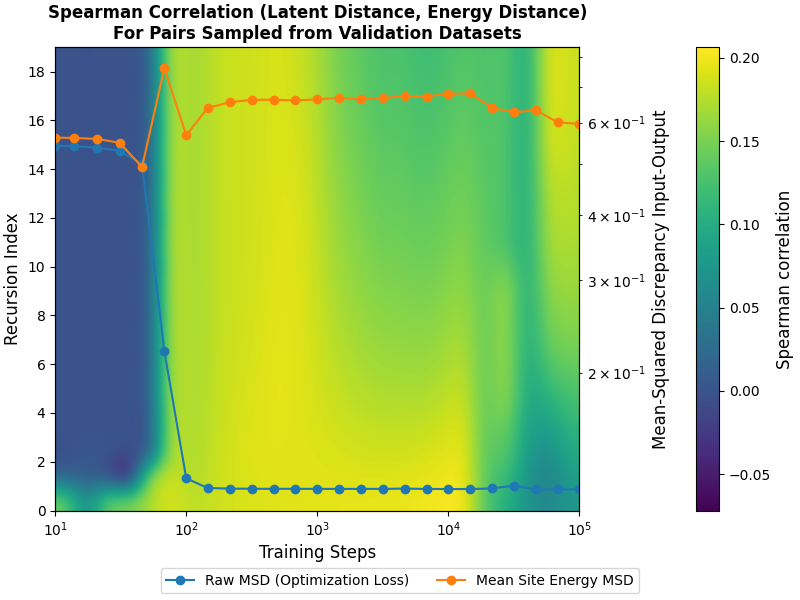}
    \caption{
    Spearman rank correlation between the input energy difference and latent space representation distance (L2-norm) of sampled validation data pairs. 
    The correlation is shown for different training checkpoints of \textbf{run MB1} (see \tabref{tab:run-names}) and using the latent representation for different numbers of recursions (feeding the output back into the input). 
    At the beginning of training, the correlation drops rapidly under recursion (lower left) as the encoder maps all points to the same output. 
    Around step 60, the correlation remains stable under recursion, concurrent with the drop in the optimized loss (blue) and a spike in the corresponding macroscopic energy loss (orange). 
    Around step 1000, the correlation decays noticeably under recursion. 
    Around step 20 000, the correlation decreases further independently, concurrent with a decrease in the energy loss. 
    The correlation and thus the topological ordering with respect to energy is unstable and even improves under recursion. 
    This is in strong contrast to the magnetization-based topological ordering quantified in \figref{fig:spearman-latent-vs-magnetization-b1}, which shows a more stable behavior under recursion. 
    There is no topological ordering present for the energy in the scale-bridging magnetization learning regime, where samples with the same low energy are on opposite sites. 
    This might be related to overfitting to the training data (see worse generalization in \figref{fig:high-loss-outputs}), as the learning rate decays to zero in this period. 
    The instability under recursion in the final stage of training could be explained by overfitting and the advent of a scale-bridging energy learning regime in which the topological magnetization order is not well preserved. 
    }
    \label{fig:spearman-latent-vs-energy-b1}
\end{figure*}

\clearpage
\subsubsection{Medium Bottleneck}

\begin{figure*}[ht]
    \centering
    \includegraphics[width=0.75\linewidth]{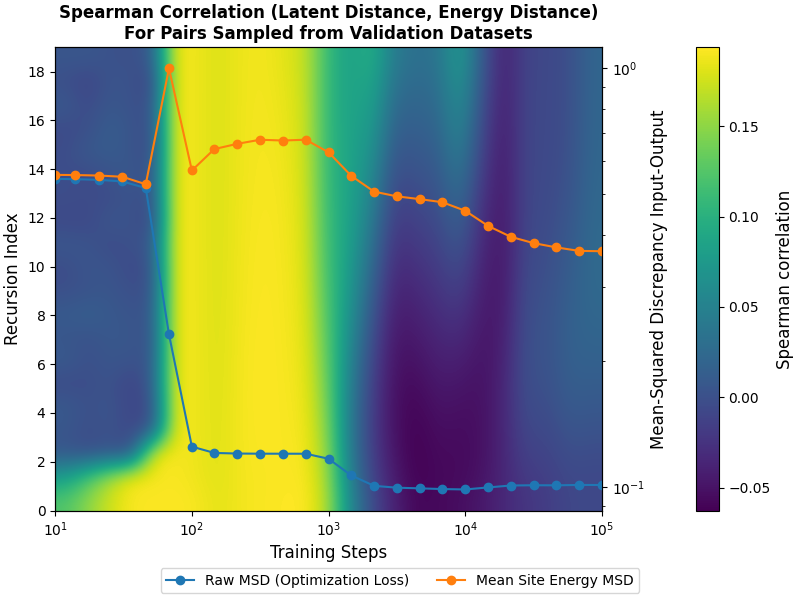}
    \caption{
    Spearman rank correlation between the input energy difference and latent space representation distance (L2-norm) of sampled validation data pairs. 
    The correlation is shown for different training checkpoints of \textbf{run MB8} (see \tabref{tab:run-names}) and using the latent representation for different numbers of recursions (feeding the output back into the input). 
    At the beginning of training, the correlation drops rapidly under recursion (lower left) as the encoder maps all points to the same output. 
    Around step 100, the correlation remains stable under recursion, concurrent with the drop in the optimized loss (blue) and a spike in the corresponding macroscopic energy loss (orange). 
    Around step 1000, the correlation decreases for any number of recursions, concurrent with another drop in the optimized loss and a decrease in the energy loss. 
    It then remains low (or even negative) and changes only slightly under recursion until the end of training. 
    The training-time periods with the higher correlation stable over recursion (around $10^2-10^3$) correspond to a transitory dynamical regime where this correlation is maximal and essentially stable under the closed-loop dynamics; it is visible as pronounced drops in run variance, see \figref{fig:variance-run-spins}. 
    This suggests that the correlation is only high with respect to the energy while energy is approximated with the magnetization. 
    As soon as the energy loss is consistently decreasing, the correlation vanishes together with the magnetization topology in \figref{fig:latent-flow-energy-b8}. 
    }
    \label{fig:spearman-latent-vs-energy-b8}
\end{figure*}

\clearpage
\subsubsection{Large Bottleneck}

\begin{figure*}[ht]
    \centering
    \includegraphics[width=0.75\linewidth]{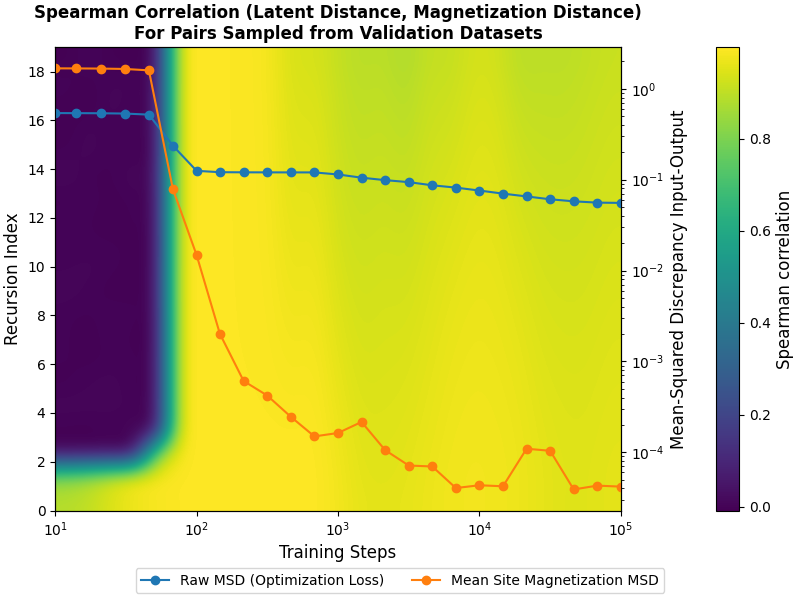}
    \caption{
    Spearman rank correlation between the input magnetization difference and latent space representation distance (L2-norm) of sampled validation data pairs. 
    The correlation is shown for different training checkpoints of \textbf{run MB64} (see \tabref{tab:run-names}) and using the latent representation for different numbers of recursions (feeding the output back into the input). 
    At the beginning of training, the correlation drops rapidly under recursion (lower left) as the encoder maps all points to the same output. 
    Around step 100, the correlation remains stable under recursion, concurrent with the drop in the optimized loss (blue) and a decrease in the corresponding macroscopic magnetization loss (orange). 
    The correlation stays high with a small decrease following the temporary increase of the magnetization loss. 
    We suspect that the increase in magnetization is due to transitions in the representation that deform the topology slightly. 
    This temporary deformation/expansion can be seen in \figref{fig:latent-flow-magnetization-b64}. 
    There is no final decay or instability in the correlation as observed for smaller bottleneck sizes in \figref{fig:spearman-latent-vs-magnetization-b1} and \figref{fig:spearman-latent-vs-magnetization-b8}. 
    This suggests that the topological ordering with respect to the magnetization can be maintained while the sample's energy is being reconstructed better (\figref{fig:high-loss-outputs}). 
    The tradeoff between these two concepts thus seems to vanish with sufficient representation capacity. 
    }
    \label{fig:spearman-latent-vs-magnetization-b64}
\end{figure*}

\begin{figure*}[ht]
    \centering
    \includegraphics[width=0.75\linewidth]{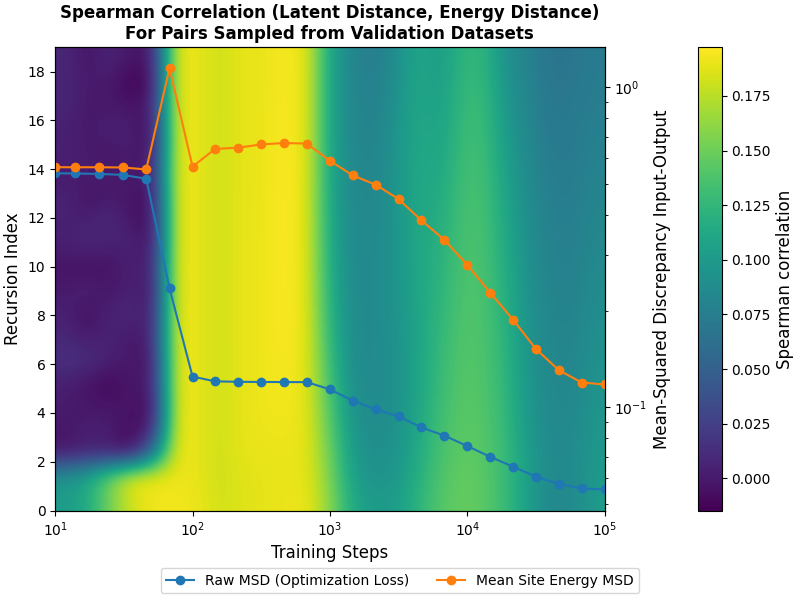}
    \caption{
    Spearman rank correlation between the input energy difference and latent space representation distance (L2-norm) of sampled validation data pairs. 
    The correlation is shown for different training checkpoints of \textbf{run MB64} (see \tabref{tab:run-names}) and using the latent representation for different numbers of recursions (feeding the output back into the input). 
    At the beginning of training, the correlation drops rapidly under recursion (lower left) as the encoder maps all points to the same output. 
    Around step 100, the correlation remains stable under recursion, concurrent with the drop in the optimized loss (blue) and a spike in the corresponding macroscopic energy loss (orange). 
    Around step 1000, the correlation decreases for any number of recursions, concurrent with another drop in the optimized loss and the macroscopic energy loss. 
    After step 1000, the correlation remains low and mostly stable under recursion until the end of training. 
    The areas with lower correlation are aligned with those visible for the magnetization in \figref{fig:spearman-latent-vs-magnetization-b64}. 
    The stability under recursion stands in stark contrast to the runs with smaller bottlenecks in \figref{fig:spearman-latent-vs-energy-b1} and \figref{fig:spearman-latent-vs-energy-b8}. 
    The strong alignment with magnetization suggests that additional latent space dimensions help to retain topological structures with respect to the magnetization across recursion. 
    This stability is then translated to a more stable correlation with respect to energy. 
    }
    \label{fig:spearman-latent-vs-energy-b64}
\end{figure*}

\end{document}